\documentclass[11pt]{article}
\usepackage[sort&compress,square,comma,authoryear]{natbib}
\usepackage[a4paper, total={6.1in, 10in}]{geometry}

\usepackage[utf8]{inputenc} 
\usepackage[T1]{fontenc}    
\usepackage{hyperref}       
\usepackage{url}            
\usepackage{booktabs}       
\usepackage{tabularx}
\newcolumntype{Y}{>{\centering\arraybackslash}X}
\newcolumntype{Z}{>{\raggedleft\arraybackslash}X}
\usepackage[parfill]{parskip}
\usepackage{amsfonts}       
\usepackage{nicefrac}       
\usepackage{microtype}      
\usepackage[dvipsnames]{xcolor}         
\usepackage{amsmath,amssymb,amsthm}
\usepackage{bbold}
\usepackage{colortbl}
\usepackage{graphicx} 
\usepackage{mathrsfs,mathtools}
\usepackage{subfigure}
\usepackage{thm-restate}
\usepackage{wrapfig}
\usepackage{caption}
\usepackage[capitalize]{cleveref}
\usepackage{tikz}
\usetikzlibrary{calc, fadings, decorations, shapes, positioning, arrows}
\usepackage[graphicx]{realboxes}

\definecolor{salmon}{RGB}{234,153,153}
\definecolor{cornflowerblue}{RGB}{100,149,237}
\hypersetup{
    colorlinks,
    linkcolor={cornflowerblue},
    citecolor={cornflowerblue},
    urlcolor={salmon}
}

\theoremstyle{plain}
\newtheorem{theorem}{Theorem}[section]

\newtheorem{lemma}[theorem]{Lemma}
\newtheorem{corollary}[theorem]{Corollary}
\theoremstyle{definition}
\newtheorem{definition}[theorem]{Definition}

\theoremstyle{remark}

\newcommand{\expect}[2]{\mathbb{E}_{#1} \left [ #2 \right ]}
\newcommand{\hsciconly}{\textsc{HSCIC}}
\newcommand{\hscic}[2]{\textsc{HSCIC}( #1 \mid #2 )}
\newcommand{\indep}{\perp \!\!\! \perp}

\newcommand{\pr}{\mathbb{P}}
\DeclareMathOperator{\var}{\mathrm{var}}
\newcommand{\VCF}{\textsc{VCF}}
\newcommand{\cD}{\mathcal{D}}
\newcommand{\cF}{\mathcal{F}}

\newcommand{\midsepremove}{\aboverulesep = 0mm \belowrulesep = 0mm}
\midsepremove


\title{Learning Counterfactually Invariant Predictors}

\date{}

\usepackage{authblk}

\author[1]{Francesco Quinzan\thanks{Part of this work was done while Francesco Quinzan visited the Max Planck Institute for Intelligent Systems, Tübingen, Germany.}${}^{\dagger}$}
\author[2]{Cecilia Casolo${}^{\dagger}$}
\author[5]{Krikamol Muandet}
\author[3]{\\Yucen Luo${}^{\ddagger}$}
\author[4,2]{Niki Kilbertus${}^{\ddagger}$}
\affil[1]{KTH Royal Institute of Technology}
\affil[2]{Helmholtz AI, Munich}
\affil[3]{Max Planck Institute for Intelligent Systems}
\affil[4]{Technical University of Munich}
\affil[5]{CISPA Helmholtz Center for Information Security}
\affil[ ]{}
\affil[ ]{\small \textit{${}^{\dagger}$,${}^{\ddagger}$equal contribution}}

\begin{document}

\maketitle

\begin{abstract}
\noindent
Notions of counterfactual invariance (CI) have proven essential for predictors that are fair, robust, and generalizable in the real world. We propose graphical criteria that yield a sufficient condition for a predictor to be counterfactually invariant in terms of a conditional independence in the observational distribution. In order to learn such predictors, we propose a model-agnostic framework, called Counterfactually Invariant Prediction (CIP), building on the Hilbert-Schmidt Conditional Independence Criterion (HSCIC), a kernel-based conditional dependence measure. Our experimental results demonstrate the effectiveness of CIP in enforcing counterfactual invariance across various simulated and real-world datasets including scalar and multi-variate settings.
\end{abstract}

\section{Introduction}\label{sec:introduction}

Invariance, or equivariance to certain data transformations, has proven essential in numerous applications of machine learning (ML), since it can lead to better generalization capabilities \citep{arjovsky2019invariant, chen2020simple, bloem2020probabilistic}. For instance, in image recognition, predictions ought to remain unchanged under scaling, translation, or rotation of the input image. Data augmentation, an early heuristic to promote such invariances, has become indispensable for successfully training deep neural networks (DNNs)~\citep{shorten2019survey, NEURIPS2020_44feb009}. Well-known examples of ``invariance by design'' include convolutional neural networks (CNNs) for translation invariance~\citep{NIPS2012_c399862d}, group equivariant NNs for general group transformations~\citep{cohen2016group}, recurrent neural networks (RNNs) and transformers for sequential data~\citep{vaswani2017attention}, DeepSet~\citep{NIPS2017_f22e4747} for sets, and graph neural networks (GNNs) for different types of geometric structures~\citep{battaglia2018relational}.

Many applications in modern ML, however, call for arguably stronger notions of invariance based on causality. This case has been made for image classification, algorithmic fairness \citep{NIPS2016_9d268236, mitchell2021algorithmic}, robustness \citep{buhlmann2020invariance}, and out-of-distribution  generalization \citep{lu2021invariant}. The goal is invariance with respect to hypothetical manipulations of the data generating process (DGP). Various works develop methods that assume observational distributions (across environments or between training and test) to be governed by shared causal mechanisms, but differ due to various types of distribution shifts encoded by the causal model
\citep{peters2016causal,heinze2018invariant,rojas2018invariant,arjovsky2019invariant,buhlmann2020invariance,subbaswamy2022unifying,yi2022breaking,makar2022causally}. Typical goals include to train predictors invariant to such shifts, to learn about causal mechanisms and to improve robustness against spurious correlations or out of distribution generalization. The term ``counterfactual invariance'' has also been used in other out of distribution learning contexts unrelated to our task, e.g., to denote invariance to certain symmetry transformations \citep{DBLP:conf/iclr/MouliR22}.

While we share the broader motivation, these works are orthogonal to ours, because even though counterfactual distributions are also generated from the same causal model, they are fundamentally different from such shifts \citep{peters2017elements}. Intuitively, counterfactuals are about events that did not, but could have happened had circumstances been different in a controlled way. A formal discussion of what we mean by counterfactuals is required to properly position our work in the existing literature and describe our contributions.
\section{Problem setting and related work}\label{sec:preliminaries}
\subsection{Preliminaries and terminology}
\begin{definition}[Structural causal model (SCM)]
\label{def:SCM}
A structural causal model is a tuple $\mathcal{S} = (\mathbf{U}, \mathbf{V}, F, \pr_{\mathbf{U}})$ such that $\mathbf{U}$ is a set of background variables that are exogenous to the model; $\mathbf{V}$ is a set of observable (endogenous) variables; $F=\{f_{V}\}_{V\in\mathbf{V}}$ is a set of functions from (the domains of) $\mathsf{pa}(V) \cup U_V$ to (the domain of) $V$, where $U_V \subset \mathbf{U}$ and $\mathsf{pa}(V) \subseteq \mathbf{V} \setminus \{V\}$ such that $V = f_V(\mathsf{pa}(V), U_V)$; $\pr_{\mathbf{U}}$ is a probability distribution over the domain of $\mathbf{U}$.
Further, the subsets $\mathsf{pa}(V) \subseteq \mathbf{V} \setminus \{V\}$ are chosen such that the graph $\mathcal{G}$ over $\mathbf{V}$ where the edge $V' \to V$ is in $\mathcal{G}$ if and only if $V' \in \mathsf{pa}(V)$ is a directed acyclic graph (DAG).
\end{definition}
\paragraph{Observational distribution.} An SCM implies a unique observational distribution over $\mathbf{V}$, which can be thought of as being generated by transforming the distribution over $\pr_{\mathbf{U}}$ via the deterministic functions in $F$ iteratively to obtain a distribution over $\mathbf{V}$.\footnote{Note that all randomness stems from $\pr_{\mathbf{U}}$. The observational distribution is well-defined and unique, essentially because every DAG allows for a topological order. }

\paragraph{Interventions.} Given a variable $A \in \mathbf{V}$, an intervention $A \gets a$ amounts to replacing $f_A$ in $F$ with the constant function setting $A$ to $a$.
This yields a new SCM, which induces the \emph{interventional distribution} under intervention $A \gets a$.\footnote{The observational distribution in an intervened SCM is called interventional distribution of the base SCM.}
Similarly, we can intervene on multiple variables $\mathbf{V} \supseteq \mathbf{A} \gets \mathbf{a}$.
For an outcome (or prediction target) variable $\mathbf{Y} \subset \mathbf{V}$, we then write $\mathbf{Y}_{\mathbf{a}}$ for the outcome in the intervened SCM, also called \emph{potential outcome}.
Note that the interventional distribution $\pr_{\mathbf{Y}_{\mathbf{a}}}(\mathbf{y})$ differs in general from the conditional distribution $\pr_{\mathbf{Y}\mid \mathbf{A}}(\mathbf{y} \mid \mathbf{a})$.\footnote{We use $\pr$ for distributions (common in the kernel literature) and $\mathbf{Y}_{\mathbf{a}}$ instead of the do notation.}
This is typically the case when $Y$ and $A$ have a shared ancestor, i.e., they are confounded.
In interventional distributions, potential outcomes are random variables via the exogenous variables $\mathbf{u}$, i.e., $Y_{\mathbf{a}}(u)$ where $\mathbf{u} \sim \pr_{\mathbf{U}}$.
Hence, interventions capture ``population level'' properties, i.e., the action is performed for all units $\mathbf{u}$.
\paragraph{Counterfactuals.} Counterfactuals capture what happens under interventions for a ``subset'' of possible units $\mathbf{u}$ that are compatible with observations $\mathbf{W} = \mathbf{w}$ for a subset of observed variables $\mathbf{W} \subseteq \mathbf{V}$.
This can be described in a three step procedure.
(i) \emph{Abduction:} We restrict our attention to units compatible with the observations, i.e., consider the new SCM $\mathcal{S}^{\mathbf{w}} = (\mathbf{U}, \mathbf{V}, F, \pr_{\mathbf{U} \mid \mathbf{W} = \mathbf{w}})$.
(ii) \emph{Intervention:} Within $\mathcal{S}^{\mathbf{w}}$, perform an intervention $\mathbf{A} \gets \mathbf{a}$ on some variables $\mathbf{A}$ (which need not be disjoint from $\mathbf{W}$).
(iii) \emph{Prediction:} Finally, we are typically interested in the outcome $\mathbf{Y}$ in an interventional distribution of $\mathcal{S}^{\mathbf{w}}$, which we denote by $\pr_{\mathbf{Y}^*_{\mathbf{a}} \mid \mathbf{W} = \mathbf{w}} (\mathbf{y})$ and call a \emph{counterfactual distribution}: ``Given that we have observed $\mathbf{W} = \mathbf{w}$, what would $\mathbf{Y}$ have been had we set $\mathbf{A} \gets \mathbf{a}$, instead of the value $\mathbf{A}$ has actually taken?''. In this notation the conditioning set $\mathbf{W}$ is conditioned on in the pre-interventional world. 
Counterfactuals capture properties of a ``subpopulation'' $\mathbf{u} \sim \pr_{\mathbf{U} \mid \mathbf{W} = \mathbf{w}}$ compatible with the observations.\footnote{Note that conditioning in an interventional distribution is different from a counterfactual and our notation is quite subtle here $\pr_{\mathbf{Y}^*_{\mathbf{a}}}(\mathbf{y} \mid \mathbf{W} =\mathbf{w}) \ne \pr_{\mathbf{Y}^*_{\mathbf{a}} \mid \mathbf{W} = \mathbf{w}} (\mathbf{y})$.}
Even for fine-grained $\mathbf{W}$, there may be multiple units $\mathbf{u}$ in the support of this distribution.
In contrast, ``unit level counterfactuals'' often considered in philosophy contrast $\mathbf{Y}^*_{\mathbf{a}}(\mathbf{u})$ with $\mathbf{Y}^*_{\mathbf{a}'}(\mathbf{u})$ for a single unit $\mathbf{u}$.
Such unit level counterfactuals are too fine-grained in our setting. Hence, our used definition of counterfactual invariance is:
\begin{definition}[Counterfactual invariance]\label{def:counterfactual_invariance}
Let $\mathbf{A}$, $\mathbf{W}$ be (not necessarily disjoint) sets of nodes in a given SCM.
Then, $\mathbf{Y}$ is \emph{counterfactually invariant in $\mathbf{A}$ w.r.t.\ $\mathbf{W}$} if $\pr_{\mathbf{Y}^*_{\mathbf{a}} \mid \mathbf{W} = \mathbf{w}}(\mathbf{y}) = \pr_{\mathbf{Y}^*_{\mathbf{a}'} \mid \mathbf{W} = \mathbf{w}}(\mathbf{y})$ almost surely, for all $\mathbf{a}, \mathbf{a}'$ in the domain of $\mathbf{A}$ and all $\mathbf{w}$ in the domain of $\mathbf{W}$.\footnote{With an abuse of notation, if $\mathbf{W} = \emptyset$ then the requirement of conditional counterfactual invariance becomes $\pr_{\mathbf{Y}_{\mathbf{a}}}(\mathbf{y}) = \pr_{\mathbf{Y}_{\mathbf{a}'}}(\mathbf{y})$ almost surely, for all $\mathbf{a}, \mathbf{a}'$ in the domain of $\mathbf{A}$.
The ``almost surely'' part in our definition merely refers to the type of equality of distributions and is not related to the ``almost sure'' in a.s.-CI defined by \citet{fawkes2023results}.}
\end{definition}
\paragraph{Predictors in SCMs.}
Ultimately, we aim at learning a predictor $\hat{\mathbf{Y}}$ for the outcome $\mathbf{Y}$.
Originally, the predictor $\hat{\mathbf{Y}}$ is not part of the DGP, because we get to learn $f_{\hat{\mathbf{Y}}}$ from data.
Using supervised learning, the predictor $f_{\hat{\mathbf{Y}}}$ depends both on the chosen inputs $\mathbf{X} \subset \mathbf{V}$ as well as the target $\mathbf{Y}$.
However, once $f_{\hat{\mathbf{Y}}}$ is fixed, it is a deterministic function with arguments $\mathbf{X} \subset \mathbf{V}$, so $(\mathbf{U}, \mathbf{V} \cup \{\hat{\mathbf{Y}}\}, F \cup \{f_{\hat{\mathbf{Y}}}\}, \pr_{\mathbf{U}})$ is a valid SCM and we can consider $\hat{\mathbf{Y}}$ an observed variable with incoming arrows from only $\mathbf{X}$.
Hence, the definition of counterfactual invariance can be applied to the predictor $\hat{\mathbf{Y}}$.
\paragraph{Kernel mean embeddings (KME).} 
Our method relies on kernel mean embeddings (KMEs).
We describe the main concepts pertaining KMEs and refer the reader to \citet{smola2007hilbert,learning_kernels_book,prob_stats_book,Muandet17:KME} for details.
Fix a measurable space $\mathscr{Y}$ with respect to a $\sigma$-algebra $\mathcal{F}_{\mathscr{Y}}$, and consider a probability measure $\pr$ on the space $(\mathscr{Y}, \mathcal{F}_{\mathscr{Y}})$.
Let $\mathcal{H}$ be a reproducing kernel Hilbert space (RKHS) with a bounded kernel $k_\mathbf{Y} \colon \mathscr{Y} \times \mathscr{Y} \rightarrow \mathbb{R}$, i.e., $k_\mathbf{Y}$ is such that $\sup_{\mathbf{y} \in \mathscr{Y}}k(\mathbf{y},\mathbf{y}) < \infty$.
The kernel mean embedding $\mu_{\pr}$ of $\pr$ is defined as the expected value of the function $k(\ \cdot \ ,\mathbf{y})$ with respect to $\mathbf{y}$, i.e., $\mu_{\pr} \coloneqq \expect{}{k(\ \cdot \  ,\mathbf{y})}$.
The definition of KMEs can be extended to conditional distributions \citep{DBLP:journals/jmlr/FukumizuSG13,DBLP:conf/icml/GrunewalderLGBPP12,DBLP:conf/icml/SongHSF09,DBLP:journals/spm/SongFG13}.
Consider two random variables $\mathbf{Y}$, $\mathbf{S}$, and denote with $(\Omega_\mathbf{Y}, \mathcal{F}_{\mathbf{Y}})$ and $(\Omega_\mathbf{S}, \mathcal{F}_{\mathbf{S}})$ the respective measurable spaces.
These random variables induce a probability measure $\pr_{\mathbf{Y}, \mathbf{S}}$ in the product space $\Omega_\mathbf{Y} \times \Omega_\mathbf{S}$.
Let $\mathcal{H}_\mathbf{Y}$ be a RKHS with a bounded kernel $k_\mathbf{Y} (\cdot, \cdot )$ on $\Omega_\mathbf{Y}$.
We define the KME of a conditional distribution $\pr_{\mathbf{Y}\mid \mathbf{S}}(\cdot \mid \mathbf{s})$ via $\mu_{\mathbf{Y}\mid \mathbf{S} = \mathbf{s}} \coloneqq \expect{}{k_\mathbf{Y}(\ \cdot \ ,\mathbf{y})\mid \mathbf{S} = \mathbf{s}}$.
Here, the expected value is taken over $\mathbf{y}$.
KMEs of conditional measures can be estimated from samples \citep{DBLP:conf/icml/GrunewalderLGBPP12}. \citet{pogodin2022efficient} recently proposed an efficient kernel-based regularizer for learning features of input data that allow for estimating a target while being conditionally independent of a distractor given the target.
Since CIP ultimately enforces conditional independence (see \cref{lemma:conditiona_counterfactual}), we believe it could further benefit from leveraging the efficiency and convergence properties of their technique, which we leave for future work.

\subsection{Related work and contributions}\label{sec:relatedwork}

While we focus on counterfactuals in the SCM framework \citep{pearlj,peters2016causal}, there are different incompatible frameworks to describe counterfactuals \citep{von2022backtracking,dorr2016against,woodward2021causation}, which may give rise to orthogonal notions of counterfactual invariance.

Research on algorithmic fairness has explored a plethora of causal ``invariance'' notions with the goal of achieving fair predictors \citep{loftus2018causal,carey2022causal,plecko2022causal}.
\citet{kilbertus2017avoiding} conceptually introduce a notion based on group-level interventions, which has been refined to take into account more fine-grained context by \citet{salimi2019capuchin,galhotra2022causal}, who then obtain fair predictors by viewing it as a database repair problem or a causal feature selection problem, respectively.
A counterfactual-level definition was proposed by \citet{DBLP:conf/nips/KusnerLRS17} and followed up by path-specific counterfactual notions \citep{nabi2018fair,DBLP:conf/aaai/Chiappa19}, where the protected attribute may take different values along different paths to the outcome.
Recently, \citet{dutta2021fairness} developed an information theoretic framework to decompose the overall causal influence allowing for exempted variables and properly dealing with synergies across different paths. 

Our focus is on counterfactuals because they are fundamentally more expressive than mere interventions \citep{pearlj,bareinboim2022pearl}, but do not require a fine-grained path- or variable-level judgment of ``allowed'' and ``disallowed'' paths or variables, which may be challenging to devise in practice.
Since CI already requires strong assumptions, we leave path-specific counterfactuals---even more challenging in terms of identifiability \citep{avin2005identifiability}---for future work. While our \Cref{def:counterfactual_invariance} requires equality in distribution, \citet{DBLP:conf/nips/VeitchDYE21} suggest a definition of a counterfactually invariant predictor $f_{\hat{\mathbf{Y}}}$ which requires almost sure equality of $\hat{\mathbf{Y}}^*_{\mathbf{a}}$ and $\hat{\mathbf{Y}}^*_{\mathbf{a}'}$, where we view $\hat{\mathbf{Y}}$ as an observed variable in the SCM as described above.
\citet{fawkes2023results} recently carefully formulated various precise technical definitions of what counterfactual invariance may mean in different contexts, such as ``almost sure CI'' (a.s.-CI in \Cref{def:fawkes-a.s.}), ``distributional CI'' ($\cD$-CI in \Cref{def:fawkes-D}), and ``CI of predictors'' ($\cF$-CI in \Cref{def:fawkes-F}). They provide various connections between them such as the fact that $f_{\hat{\mathbf{Y}}}$ being $\cF$-CI is equivalent to $\hat{\mathbf{Y}}$ being $\cD$-CI conditioned on $\mathbf{X}$, yielding an equivalence to the definition of counterfactual fairness \citep{DBLP:conf/nips/KusnerLRS17}.
The notion of counterfactual invariance in \Cref{def:counterfactual_invariance} is most closely related to $\cD$-CI from \citet{fawkes2023results}, but we do not enforce conditioning on the intervening variable.

Inspired by problems in natural language processing (NLP), \citet{DBLP:conf/nips/VeitchDYE21} aim at ``stress-testing'' models for spurious correlations.
It differs from our work in that they (i) focus only on two specific graphs, and (ii) provide a \emph{necessary} but not sufficient criterion for CI in terms of a conditional independence.
Their method enforces the conditional independence via maximum mean discrepancy (MMD) (in \emph{discrete settings only}).
However, enforcing a consequence of CI, does not necessarily improve CI.
Indeed, \citet[Prop.~4.4]{fawkes2023results} show that while a.s.-CI implies certain conditional independencies, no set of conditional independencies implies any bounds on the difference in counterfactuals. On the contrary, distributional notions of CI such as $\cD$-CI or our \Cref{def:counterfactual_invariance} are weaker than a.s.-CI (see \citep[Lem.~2.4]{fawkes2023results}.
Therefore, these weaker notions can indeed be written equivalently as conditional independencies in the observational distribution under additional assumptions about the data generating mechanism \citep[Lem.~A.3]{fawkes2023results}.
For example, \citet[Lem.~A.3]{fawkes2023results} shows that $\cD$-CI can be implied by conditional independence in special settings where the counterfactual distribution is identified from the observational one.
Similarly, our reduction of CI (as in \Cref{def:counterfactual_invariance}
) to conditional independence in \Cref{lemma:conditiona_counterfactual} requires a strong injectivity assumption that essentially amounts to being able to remove exogenous uncertainty.

\paragraph{Contributions.}
We provide such a sufficient graphical criterion for $\cD$-CI under an injectivity condition of a structural equation. Depending on the assumed causal graph, this can also come at the cost of requiring certain variables to be observed. As our main contribution, we propose a model-agnostic learning framework, called Counterfactually Invariant Prediction (CIP), using a kernel-based conditional dependence measure that also works for mixed categorical and continuous, multivariate variables. We evaluate CIP extensively in (semi-)synthetic settings and demonstrate its efficacy in enforcing counterfactual invariance even when the strict assumptions may be violated.

\begin{figure}[t]
  \centering
    \begin{tikzpicture}[node distance=6mm and 8mm, main/.style = {draw, circle, minimum size=0.6cm, inner sep=1pt}, >={triangle 45}]  
    \node[main] (1) {$\scriptstyle \mathbf{A}$};
    \node[main] (2) [right =of 1] {$\scriptstyle \mathbf{L}$}; 
    \node[main] (3) [right =of 2] {$\scriptstyle \mathbf{Y}$}; 
    \node[main] (4) [above =of $(2)!0.5!(3)$] {$\scriptstyle \mathbf{S}$}; 
    \node[] (0) [left =of 1]  {}; 
    \draw[->] (1) -- (2); 
    \draw[->] (2) -- (3); 
    \draw[->] (4) -- (2);
    \draw[->] (4) -- (3);
    \draw[->] (1) to [bend right] (3); 
    \node (label) [above =of 1] {\textbf{(a)}};
  \end{tikzpicture}%
  \hfill
  \begin{tikzpicture}[node distance=6mm and 8mm, main/.style = {draw, circle, minimum size=0.6cm, inner sep=1pt}, >={triangle 45}]  
    \node[main] (1) {$\scriptstyle \mathbf{A}$}; 
    \node[main] (2) [right =of 1] {$\scriptstyle \mathbf{X}^{\wedge}$}; 
    \node[main] (3) [right =of 2] {$\scriptstyle \mathbf{Y}$}; 
    
    \node[main] (4) [above =of $(2)!0.5!(3)$] {$\scriptstyle \mathbf{X}^{\bot}_{\mathbf{A}}$}; 
    \node[main] (5) [above =of $(1)!0.5!(2)$] {$\scriptstyle \mathbf{X}^{\bot}_{\mathbf{Y}}$}; 
    \draw[->] (1) -- (2); 
    \draw[->] (1) -- (5); 
    \draw[->] (2) -- (3); 
    \draw[->] (2) -- (5); 
    \draw[->] (4) -- (2); 
    \draw[->] (4) -- (3); 
    \draw[<->, dashed, bend right] (1) to (3); 
    \node (label) [above =of 1] {\textbf{(b)}};
  \end{tikzpicture}%
  \hfill
  \begin{tikzpicture}[node distance=6mm and 8mm, main/.style = {draw, circle, minimum size=0.6cm, inner sep=1pt}, >={triangle 45}]  
    \node[main] (3) {$\scriptstyle \mathbf{X}^{\bot}_{\mathbf{Y}}$}; 
    \node[main] (5) [right =of 3] {$\scriptstyle \mathbf{X}^{\wedge}$}; 
    \node[main] (4) [right =of 5] {$\scriptstyle \mathbf{X}^{\bot}_{\mathbf{A}}$}; 
    \node[main] (1) [above =of $(3)!0.5!(5)$]{$\scriptstyle \mathbf{Y}$}; 
    \node[main] (2) [above =of $(5)!0.5!(4)$] {$\scriptstyle \mathbf{A}$}; 

    \node[] (0) [right =of 4]  {}; 
    
    \draw[->] (1) -- (3); 
    \draw[->] (2) -- (4); 
    \draw[->] (1) -- (5); 
    \draw[->] (2) -- (5); 
    \draw[<->, dashed] (1) -- (2); 
    \node (label) (label) [above =of 3] {\textbf{(c)}}; 
    \draw[<->, draw=none, bend right] (3) to (4); 
  \end{tikzpicture}\\[-5mm]
  \begin{tikzpicture}[node distance=6mm and 8mm, main/.style = {draw, circle, minimum size=0.6cm, inner sep=1pt}, >={triangle 45}]  
    \node[main] (1) {$\scriptstyle \mathbf{A}$}; 
    \node[main] (2) [right =of 1] {$\scriptstyle \mathbf{L}$}; 
    \node[main] (3) [right =of 2] {$\scriptstyle \mathbf{Y}$}; 
    \node[main] (4) [above =of 2] {$\scriptstyle \mathbf{S}$}; 

    \node[] (5) [above =of 4]  {};
    \node[] (0) [left =of 1]  {}; 
    
    \node (label) [left =of 4] {\textbf{(d)}};
    \draw[->] (1) -- (2); 
    \draw[->] (2) -- (3); 
    \draw[->] (4) -- (1);
    \draw[->] (4) -- (2); 
    \draw[->] (4) -- (3); 
    \draw[->] (1) to [bend right] (3);
  \end{tikzpicture}
  \hfill
  \begin{tikzpicture}[node distance=6mm and 8mm, main/.style = {draw, circle, minimum size=0.6cm, inner sep=1pt}, >={triangle 45}]  
    \node[main] (1) {$\scriptstyle \mathbf{A}$}; 
    \node[main] (2) [right =of 1] {$\scriptstyle \mathbf{L}$}; 
    \node[main] (3) [right =of 2] {$\scriptstyle \mathbf{Y}$}; 
    \node[main] (4) [above =of 2] {$\scriptstyle \mathbf{C}$}; 
    \draw[->] (1) -- (2); 
    \draw[->] (2) -- (3); 
    \draw[->] (4) -- (2); 
    \draw[->] (4) -- (3); 
    \draw[->] (1) to [bend right] (3); 
    \node (label) [left =of 4] {\textbf{(e)}};
  \end{tikzpicture}%
  \hfill
  \begin{tikzpicture}[node distance=6mm and 8mm, main/.style = {draw, circle, minimum size=0.6cm, inner sep=1pt}, >={triangle 45}]  
    \node[main] (1) {$\scriptstyle \mathbf{A}$}; 
    \node[main] (3) [right =of 1] {$\scriptstyle \mathbf{U}$}; 
    \node[main] (5) [above =of 3] {$\scriptstyle \mathbf{C}$}; 
    \node[main] (4) [right =of 3] {$\scriptstyle \mathbf{Y}$}; 
    \node[main] (6) [above =of 4] {$\scriptstyle \mathbf{L}$}; 
    
    \node[] (0) [right =of 4]  {}; 

    \draw[->] (1) -- (3); 
    \draw[->] (5) -- (1); 
    \draw[->] (5) -- (3); 
    \draw[->] (5) -- (4); 
    \draw[->] (3) -- (4); 
    \draw[->] (6) -- (4); 
    \draw[->] (1) to [bend right] (4); 
    \node (label) [left =of 5] {\textbf{(f)}};
  \end{tikzpicture}
  \caption{%
    \textbf{(a)} Exemplary graph in which a predictor $\hat{\mathbf{Y}}$ with $\hat{\mathbf{Y}}\indep \mathbf{A} \cup \mathbf{L} \mid \mathbf{S}$ is CI in $\mathbf{A}$ w.r.t.\ $\{\mathbf{L},\mathbf{S}\}$.
    \textbf{(b)-(c)} Causal and anti-causal structure from \citet{DBLP:conf/nips/VeitchDYE21} where $\mathbf{X}^{\bot}_{\mathbf{A}}$ is not causally influenced by $\mathbf{A}$, $\mathbf{X}^{\bot}_{\mathbf{Y}}$ does not causally influence $\mathbf{Y}$, and $\mathbf{X}^\wedge$ is both influenced by $\mathbf{A}$ and influences $\mathbf{Y}$. \textbf{(d)} Assumed causal structure for the synthetic experiments, see \cref{sec:synthetic_experiments,app:synthetic} for details.
    \textbf{(e)} Assumed causal graph for the UCI Adult dataset (\cref{sec:fairnessresults}), where $\mathbf{A} = \{\text{Gender, Age}\}$. \textbf{(f)} Causal structure for our semi-synthetic image experiments (\cref{sec:imagedata}), where $\mathbf{A} = \{\text{\small{Pos.X}}\}$, $\mathbf{U} = \{\text{\small{Scale}}\}$, $\mathbf{C} = \{\text{\small{Shape}}, \text{\small{Pos.Y}}\}$,
    $\mathbf{L} = \{\text{\small{Color}}, \text{\small{Orientation}}\}$, and $\mathbf{Y} = \{\text{\small{Outcome}}\}$.
}\label{fig:graph}
\end{figure}
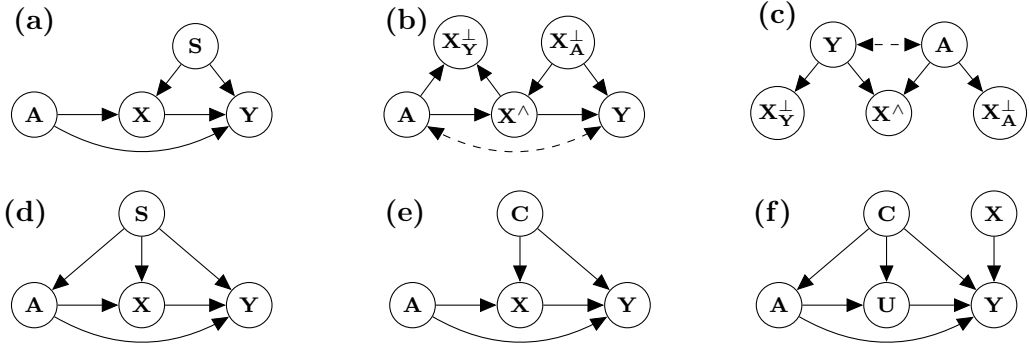
\section{Counterfactually invariant prediction (CIP)}\label{sec:ci}
\subsection{Sufficient criterion for counterfactual invariance}%

We will now establish a sufficient graphical criterion to express CI as conditional independence in the observational distribution, rendering it estimable from data. First, we need some terminology.
\paragraph{Graph terminology.}
Consider a path $\pi$ (a sequence of distinct adjacent nodes) in a DAG $\mathcal{G}$. A set of nodes $\mathbf{S}$ is said to \emph{block $\pi$}, if $\pi$ contains a triple of consecutive nodes $A, B, C$ such that one of the following hold: (i) $A \to B \to C$ or $A \gets B \gets C$ or $A \gets B \to C$ and $B \in \mathbf{S}$; (ii) $A \rightarrow B \gets C$ and neither $B$ nor any descendent of $B$ is in $\mathbf{S}$. Further, we call $\pi$ a \emph{causal path} between sets of nodes $\mathbf{A}, \mathbf{B}$, when it is a directed path from a node in $\mathbf{A}$ to a node in $\mathbf{B}$. A causal path $\pi$ is a \emph{proper causal path} if it only intersects $\mathbf{A}$ at the first node in $\pi$. Finally, we denote with $\mathcal{G}_{\mathbf{A}}$ the graph obtained by removing from $\mathcal{G}$ all incoming arrows into nodes in $\mathbf{A}$. We now define the notion of valid adjustment sets \citep[Def.~5]{DBLP:conf/uai/ShpitserVR10}, which our graphical criterion for CI relies on.
\begin{restatable}[valid adjustment set]{definition}{validadjset}
\label{def:valid_adjustment set}
Let $\mathcal{G}$ be a causal graph and let $\mathbf{X}$, $\mathbf{Y}$ be disjoint (sets of) nodes in $\mathcal{G}$. A set of nodes $\mathbf{S}$ is a valid adjustment set for $(\mathbf{X}, \mathbf{Y})$, if
(i) no element in $\mathbf{S}$ is a descendant in $\mathcal{G}_{\mathbf{X}}$ of any node $W \notin \mathbf{X}$ which lies on a proper causal path from $\mathbf{X}$ to $\mathbf{Y}$, and (ii) $\mathbf{S}$ blocks all non-causal paths from $\mathbf{X}$ to $\mathbf{Y}$ in $\mathcal{G}$.
\end{restatable}
We now state the sufficient graphical criterion that renders CI equivalent to a conditional independence.
The proof builds on known results from the literature such as the full characterization of valid adjustment sets in fully observed structural causal models and the backdoor criteria, but the final statement requires a substantial original theoretical contribution, which we detail in \cref{proof:conditiona_counterfactual}.

\begin{restatable}{theorem}{condcounterfactual}
\label{lemma:conditiona_counterfactual}
Let $\mathcal{G}$ be a causal graph, $\mathbf{A}$, $\mathbf{W}$ be two (not necessarily disjoint) sets of nodes in $\mathcal{G}$, such that $(\mathbf{A} \cup \mathbf{W}) \cap \mathbf{Y} = \emptyset$, let $\mathbf{S}$ be a valid adjustment set for $(\mathbf{A} \cup \mathbf{W}, \mathbf{Y})$.
Further, for any random variable $X \in \mathbf{W} \setminus \mathbf{A}$ denote with $g_X(\mathsf{pa}(X), U_X)$ its structural equation, and suppose that $\mathsf{pa}(X) \subseteq \mathbf{A} \cup \mathbf{W} $. Suppose that $g_X$ is injective in the variable $\mathbf{u}$.\footnote{The injectivity of $g_X$ is defined as follows. Consider two pairs $\{\mathbf{p},u\}$ and $\{\mathbf{p},u'\}$ with $\mathbf{p}$ in the support of $\mathsf{pa}(X)$ and $u,u'$ in the support of $U_X$. Suppose that $\mathbb{P}_{\mathsf{pa}(X),U_X}(\mathbf{p},u)\neq 0$ and $\mathbb{P}_{\mathsf{pa}(X),U_X}(\mathbf{p},u')\neq 0$. Then, it holds $g(\mathbf{p},u) = g(\mathbf{p},u')$ if and only if $u = u'$.} Then, in all SCMs compatible with $\mathcal{G}$, if a predictor $\hat{\mathbf{Y}}$ satisfies $\hat{\mathbf{Y}}\indep \mathbf{A}\cup \mathbf{W} \mid \mathbf{S}$, then $\hat{\mathbf{Y}}$ is counterfactually invariant in $\mathbf{A}$ with respect to $\mathbf{W}$.
\end{restatable}
\paragraph{Assumptions.}
First, we do \emph{not} assume the SCM (or any structural equations) to be known.
We do assume the causal graph to be known. This is a standard assumption widely made in the causality literature, even though it is a strong one \citep{cartwright2007hunting}.
The additional assumption of \cref{lemma:conditiona_counterfactual}, namely injectivity of $g_X$ is satisfied---but more general than---a wide variety of commonly considered models in causality such as the widely used Additive Noise Models \citep{peters2017elements}.
More broadly, \citet[Lem.~A.3]{fawkes2023results} show that to achieve CI from the observational distribution and the causal graph, additional assumptions are always required. 
\subsection{Example use-cases of counterfactually invariant prediction}
\cref{fig:graph}(a) shows an exemplary graph in which the outcome $\mathbf{Y}$ is affected by (disjoint) sets $\mathbf{A}$ (in which we want to be CI), $\mathbf{L}$ and $\mathbf{S}$ (inputs to $f_{\hat{\mathbf{Y}}}$).
We consider $\mathbf{W} = \mathbf{L} \cup \mathbf{A} \cup \mathbf{S}$.
Here we aim to achieve $\hat{\mathbf{Y}} \indep \mathbf{A} \cup \mathbf{L} \mid \mathbf{S}$ to obtain CI in $\mathbf{A}$ w.r.t.\ $\mathbf{W}$.
In our synthetic experiments, we also allow $\mathbf{S}$ to affect $\mathbf{A}$, see \cref{fig:graph}(d).
Let us further illustrate concrete potential applications of CI, which we later also study in our experiments.
\paragraph{Counterfactual fairness.}
Counterfactual fairness \citep{DBLP:conf/nips/KusnerLRS17} informally challenges a consequential decision: ``\emph{Would I have gotten the same outcome had my gender, race, or age been different with all else being equal}?''. Here $\mathbf{Y} \subset \mathbf{V}$ denotes the outcome and $\mathbf{A} \subset \mathbf{V} \setminus \mathbf{Y}$ the \emph{protected attributes} such as gender, race, or age---protected under anti-discrimination laws \citep{us_law}---by $\mathbf{A} \subseteq \mathbf{V} \setminus \mathbf{Y}$. Collecting all remaining observed covariates into $\mathbf{W} \coloneqq \mathbf{V} \setminus \mathbf{Y}$ counterfactual fairness reduces to counterfactual invariance. In experiments, we build a semi-synthetic DGP assuming the graph in \cref{fig:graph}(e) for the UCI adult dataset \citep{uci_dataset}.
\paragraph{Robustness.}
CI serves as a strong notion of robustness in settings such as image classification: ``\emph{Would the truck have been classified correctly had it been winter in this situation instead of summer}?'' For concrete demonstration, we use the dSprites dataset \citep{dsprites17} consisting of simple black and white images of different shapes (squares, ellipses, \ldots), sizes, orientations, and locations. We devise a DGP for this dataset with the graph depicted in \cref{fig:graph}(f).
\paragraph{Text classification.}
\citet{DBLP:conf/nips/VeitchDYE21} motivate the importance of counterfactual invariance in text classification tasks. Specifically, they consider the causal and anti-causal structures depicted in \citet[Fig.~1]{DBLP:conf/nips/VeitchDYE21}, which we replicate in \cref{fig:graph}(b,c). Both diagrams consist of protected attributes $\mathbf{A}$, observed covariates $\mathbf{X}$, and outcomes $\mathbf{Y}$. To apply our sufficient criterion to their settings, we must assume that $\mathbf{A}$ and $\mathbf{Y}$ are unconfounded.
We show in \cref{app:sec:baseline_veitch} that CIP still performs on par with \citet{DBLP:conf/nips/VeitchDYE21} even when this assumption is violated. \Cref{lemma:conditiona_counterfactual} provides a sufficient condition for CI (\Cref{def:counterfactual_invariance}) in terms of  the conditional independence $\hat{\mathbf{Y}} \indep \mathbf{A} \cup \mathbf{W} \mid \mathbf{S}$. We next develop an operator $\hscic{\hat{\mathbf{Y}},\mathbf{A} \cup \mathbf{W}}{\mathbf{S}}$ that is (a) efficiently estimable from data, (b) differentiable, (c) a monotonic measure of conditional dependence, and (d) is zero if and only if $\hat{\mathbf{Y}} \indep \mathbf{A} \cup \mathbf{W} \mid \mathbf{S}$. Hence, it is a practical objective to enforce CI.
\subsection{HSCIC for conditional independence}
Consider two sets of random variables $\mathbf{Y}$ and $\mathbf{A}\cup \mathbf{W}$, and denote with $(\Omega_\mathbf{Y}, \mathcal{F}_{\mathbf{Y}})$ and $(\Omega_{\mathbf{A}\cup \mathbf{W}}, \mathcal{F}_{\mathbf{A}\cup \mathbf{W}})$ the respective measurable spaces. Suppose that we are given two RKHSs $\mathcal{H}_{\mathbf{Y}}$, $\mathcal{H}_{\mathbf{A}\cup \mathbf{W}}$ over the support of $\mathbf{Y}$ and $\mathbf{A}\cup \mathbf{W}$ respectively. The tensor product space $\mathcal{H}_{\mathbf{Y}} \otimes \mathcal{H}_{\mathbf{A}\cup \mathbf{W}}$ is defined as the space of functions of the form $(f \otimes g)(\mathbf{y}, [\mathbf{a}, \mathbf{w}]) \coloneqq f(\mathbf{y}) g(\mathbf{[\mathbf{a}, \mathbf{w}]})$, for all $f \in \mathcal{H}_{\mathbf{Y}}$ and $g \in \mathcal{H}_{\mathbf{A}\cup \mathbf{W}}$. The tensor product space yields a natural RKHS structure, with kernel $k$ defined by $k(\mathbf{y} \otimes [\mathbf{a}, \mathbf{w}], \mathbf{y}' \otimes [\mathbf{a}', \mathbf{w}']) \coloneqq k_{\mathbf{Y}}(\mathbf{y}, \mathbf{y}') k_{\mathbf{A}\cup \mathbf{W}}([\mathbf{a}, \mathbf{w}], [\mathbf{a}', \mathbf{w}'])$. We refer the reader to \citet{DBLP:journals/jmlr/SzaboS17} for more details on tensor product spaces.
\begin{definition}[\hsciconly{}]
\label{def:definition_hscic}
For (sets of) random variables $\mathbf{Y}$, $\mathbf{A} \cup \mathbf{W}$, $\mathbf{S}$, the \hsciconly{} \emph{between $\mathbf{Y}$ and $\mathbf{A}\cup \mathbf{W}$ given $\mathbf{S}$} is defined as the real-valued random variable $\hscic{\mathbf{Y},\mathbf{A}\cup \mathbf{W}}{\mathbf{S}} = H_{\mathbf{Y},\mathbf{A}\cup \mathbf{W}\mid \mathbf{S}} \circ \mathbf{S}$ where $H_{\mathbf{Y},\mathbf{A}\cup \mathbf{W}\mid \mathbf{S}}$ is a real-valued deterministic function, defined as $H_{\mathbf{Y},\mathbf{A}\cup \mathbf{W}\mid \mathbf{S}}(\mathbf{s}) \coloneqq  \| \mu_{\mathbf{Y},\mathbf{A}\cup \mathbf{W}\mid \mathbf{S} = \mathbf{s}} - \mu_{\mathbf{Y}\mid \mathbf{S} = \mathbf{s}}\otimes \mu_{\mathbf{A}\cup \mathbf{W}\mid \mathbf{S} = \mathbf{s}} \|$ with $\left \| \cdot \right \|$ the norm induced by the inner product of the tensor product space $\mathcal{H}_{\mathbf{X}}\otimes \mathcal{H}_{\mathbf{A}\cup \mathbf{W}}$. 
\end{definition}
Our \Cref{def:definition_hscic} is motivated by, but differs slightly from \citet[Def.~5.3]{Park20:CME}, which relies on the Bochner conditional expected value. While it is functionally equivalent (with the same implementation, see \cref{eq:estimate_hscic}), ours has the benefit of bypassing some technical assumptions required by \citet{Park20:CME} (see \cref{appendix:ind_bochner,appendix:ind_cross_covariance} for details).
The \hsciconly{} has the following important property, proved in \cref{appendix:hscic_independence}.
\begin{restatable}[Theorem~5.4 by \citet{Park20:CME}]{theorem}{hscicind}
\label{thm:hscic_independence}
If the kernel $k$ of $\mathcal{H}_{\mathbf{X}}\otimes \mathcal{H}_{\mathbf{A}\cup \mathbf{W}}$ is characteristic\footnote{The tensor product kernel $k$ is characteristic if $\pr_{\mathbf{Y}, \mathbf{A}\cup \mathbf{W}} \mapsto \expect{\mathbf{y}, [\mathbf{a},\mathbf{w}]}{k(\ \cdot \ , \mathbf{y}\otimes [\mathbf{a},\mathbf{w}])}$ is injective.}, $\hscic{\mathbf{Y},\mathbf{A}\cup \mathbf{W}}{\mathbf{S}} = 0$ almost surely if and only if $\mathbf{Y}\indep \mathbf{A}\cup \mathbf{W} \mid \mathbf{S}$. 
\end{restatable}
Because ``most interesting'' kernels such as the Gaussian and Laplacian kernels are characteristic, and the tensor product of translation-invariant characteristic kernels is characteristic again \citep{DBLP:journals/jmlr/SzaboS17}, this natural assumption is non-restrictive in practice.
Combining \cref{lemma:conditiona_counterfactual,thm:hscic_independence}, we can now use \hsciconly{} to reliably achieve counterfactual invariance.
\begin{corollary}
\label{cor:hscic_independence}
Under the assumptions of \cref{lemma:conditiona_counterfactual}, if $\hscic{\hat{\mathbf{Y}},\mathbf{A}\cup \mathbf{W}}{\mathbf{S}} = 0$ almost surely, then $\hat{\mathbf{Y}}$ is counterfactually invariant in $\mathbf{A}$ with respect to $\mathbf{W}$.
\end{corollary}
In addition, since \hsciconly{} is defined in terms of the MMD (\Cref{def:definition_hscic} and \citet[Def.~5.3]{Park20:CME}), it inherits the weak convergence property, i.e., if $\hscic{\hat{\mathbf{Y}},\mathbf{A}\cup \mathbf{W}}{\mathbf{S}}$ converges to zero, then the counterfactual distributions (for different intervention values $\mathbf{a}$) weakly converge to the same distribution. We refer to \citet{Simon18:KDE,Simon20:Metrizing} for a precise characterization. Hence, as \hsciconly{} decreases, the predictor approaches counterfactual invariance and we need not drive \hsciconly{} all the way to zero to obtain meaningful results.

\subsection{Learning counterfactually invariant predictors (CIP)}\label{sec:learningCIP}

\Cref{cor:hscic_independence} justifies our proposed objective, namely to minimize the following loss
\begin{equation}\label{eq:total_loss}
    \mathcal{L}_{\textsc{CIP}}(\hat{\mathbf{Y}}) = \mathcal{L}(\hat{\mathbf{Y}}) + \gamma \cdot\hsciconly{}(\hat{\mathbf{Y}},\mathbf{A}\cup \mathbf{W} \mid \mathbf{S})\:, \qquad {\color{gray}\text{[CIP loss]}}
\end{equation}
where $\mathcal{L}(\hat{\mathbf{Y}})$ is a task-dependent loss function (e.g., cross-entropy for classification, or mean squared error for regression) and $\gamma \ge 0$ is a parameter that regulates the trade-off between predictive performance and counterfactual invariance. 

\paragraph{The meaning of $\gamma$ and how to choose it.}
The second term in \cref{eq:total_loss} amounts to the additional objective of CI, which is typically at odds with predictive performance within the observational distribution $\mathcal{L}$. In practice, driving \hsciconly{} to zero, i.e., viewing our task as a constrained optimization problem, typically deteriorates predictive performance too much to be useful for prediction---especially in small data settings.\footnote{In particular, \hsciconly{} does not regularize an ill-posed problem, i.e., it does not merely break ties between predictors with equal $\mathcal{L}(\hat{\mathbf{Y}})$. Hence it also need not decay to zero as the sample size increases.} As the choice of $\gamma$ amounts to choosing an operating point between predictive performance and CI, it cannot be selected in a data-driven fashion. As different settings call for different tradeoffs, we advocate for employing the following procedure: (i) Train an unconstrained predictor for a base predictive performance (e.g., 92\% accuracy or $0.21$ MSE). (ii) Fix a tolerance level $\alpha$, indicating the maximally tolerable loss in predictive performance (e.g., at most 5\% drop in accuracy or at most 10\% increase in MSE). (iii) Perform a log-spaced binary search on $\gamma$ (e.g., on $[10^{-4}, 10^4]$) to find the largest $\gamma$ such that the predictive performance of the resulting predictor achieves predictive performance within the tolerance $\alpha$---see \cref{app:choosinggamma} for an illustration. A similar search for the optimal value of $\gamma$ can be conducted, when there is a fixed requirement for a maximum tolerance of counterfactual invariance as measured by \hsciconly{}.
\paragraph{Estimating the \hsciconly{} from samples.}
The key benefit of \hsciconly{} as a conditional independence measure is that it does not require parametric assumptions on the underlying probability distributions, and it is applicable for any mixed, multi-dimensional data modalities, as long as we can define positive definite kernels on them. Given $n$ samples $\{(\hat{\mathbf{y}}_i, \mathbf{a}_i, \mathbf{w}_i, \mathbf{s}_i) \}_{i = 1}^n$, denote with $\hat{K}_{\mathbf{\hat{Y}}}$ the kernel matrix with entries $[\hat{K}_{\mathbf{\hat{Y}}}]_{i,j} \coloneqq k_{\hat{\mathbf{Y}}}(\hat{\mathbf{y}}_i, \hat{\mathbf{y}}_j)$, and let $\hat{K}_{\mathbf{A}\cup \mathbf{W}}$ be the kernel matrix for $\mathbf{A}\cup \mathbf{W}$. We estimate the $H_{\mathbf{\hat{Y}},\mathbf{A}\cup \mathbf{W}\mid \mathbf{S}} \equiv H_{\mathbf{\hat{Y}},\mathbf{A}\cup \mathbf{W}\mid \mathbf{S}}(\cdot)$ as
\begin{align}\label{eq:estimate_hscic}
    \hat{H}^2_{\mathbf{\hat{Y}},\mathbf{A}\cup \mathbf{W}\mid \mathbf{S}}
     &= \hat{w}^T_{\mathbf{\hat{Y}},\mathbf{A}\cup \mathbf{W}\mid \mathbf{S}} \left ( \hat{K}_{\mathbf{\hat{Y}}} \odot \hat{K}_{\mathbf{A}\cup \mathbf{W}}\right ) \hat{w}_{\mathbf{\hat{Y}},\mathbf{A}\cup \mathbf{W}\mid \mathbf{S}}\\
     &- 2 \left ( \hat{w}^T_{\mathbf{\hat{Y}} \mid \mathbf{S}} \hat{K}_{\mathbf{Y}} \hat{w}_{\mathbf{\hat{Y}},\mathbf{A}\cup \mathbf{W} \mid \mathbf{S}}  \right ) \left ( \hat{w}^T_{\mathbf{A}\cup \mathbf{W} \mid \mathbf{S}} \hat{K}_{\mathbf{A}\cup \mathbf{W}} \hat{w}_{\mathbf{\hat{Y}},\mathbf{A}\cup \mathbf{W} \mid \mathbf{S}}  \right )\nonumber\\
     &+ \left ( \hat{w}^T_{\mathbf{\hat{Y}} \mid \mathbf{S}} \hat{K}_{\mathbf{\hat{Y}}} \hat{w}_{\mathbf{\hat{Y}} \mid \mathbf{S}}  \right ) \left ( \hat{w}^T_{\mathbf{A}\cup \mathbf{W} \mid \mathbf{S}} \hat{K}_{\mathbf{A}\cup \mathbf{W}} \hat{w}_{\mathbf{A}\cup \mathbf{W} \mid \mathbf{S}}  \right )\;, \nonumber
\end{align}
where $\odot$ is element-wise multiplication. The functions $\hat{w}_{\mathbf{\hat{Y}}\mid \mathbf{S}}\equiv \hat{w}_{\mathbf{\hat{Y}}\mid \mathbf{S}}(\cdot)$, $\hat{w}_{\mathbf{A}\cup \mathbf{W}\mid \mathbf{S}}\equiv \hat{w}_{\mathbf{A}\cup \mathbf{W}\mid \mathbf{S}}(\cdot)$, and $\hat{w}_{\mathbf{\hat{Y}},\mathbf{A}\cup \mathbf{W}\mid \mathbf{S}}\equiv \hat{w}_{\mathbf{\hat{Y}},\mathbf{A}\cup \mathbf{W}\mid \mathbf{S}}(\cdot)$ are found via kernel ridge regression. \citet{DBLP:journals/focm/CaponnettoV07} provide the convergence rates of the estimand $\hat{H}^2_{\mathbf{\hat{Y}},\mathbf{A}\cup \mathbf{W}\mid \mathbf{S}}$ under mild conditions. In practice, computing the \hsciconly{} approximation by the formula in \cref{eq:estimate_hscic} can be computationally expensive.
We provide the runtime for some of our experimental setting showing that including the \hsciconly{} term can slow down training anywhere between 2x and several 100x in \cref{app:runtime}. To speed it up, we can use random Fourier features to approximate $\hat{K}_{\mathbf{\hat{Y}}}$ and $\hat{K}_{\mathbf{A}\cup \mathbf{W}}$ \citep{DBLP:conf/nips/RahimiR07,DBLP:conf/icml/AvronKMMVZ17}. The details on how to substantially speed up the computations with such approximations are described in \cref{appendix:random_fourier_features} and we describe the improvements in terms of computational complexity in \cref{app:runtime}. We emphasize that \cref{eq:estimate_hscic} allows us to consistently estimate the \hsciconly{} \emph{from observational i.i.d.\ samples, without prior knowledge of the counterfactual distributions}.

\subsection{Measuring counterfactual invariance.}

Besides predictive performance, e.g., mean squared error (MSE) for regression or accuracy for classification, our key metric of interest is the level of counterfactual invariance achieved by the predictor $\hat{\mathbf{Y}}$. First, we emphasize again that counterfactual distributions are generally not identified from the observational distribution (i.e., from available data) meaning that \emph{CI is generally untestable in practice} from observational data. We can thus only evaluate CI in (semi-)synthetic settings where we have access to the full SCM and thus all counterfactual distributions.

A measure for CI must capture how the distribution of $\hat{\mathbf{Y}}^*_{\mathbf{a}'}$ changes for different values of $\mathbf{a}'$ across all conditioning values $\mathbf{w}$ (which may include an observed value $\mathbf{A} = \mathbf{a}$). We propose the \textbf{V}ariance of \textbf{C}ounter\textbf{F}actuals (\VCF{}) as a metric of CI
\begin{equation}\label{eq:vcf}
  \VCF{}(\hat{\mathbf{Y}}) := \mathbb{E}_{\mathbf{w} \sim\pr_{\mathbf{W}}}\Bigl[ \var_{\mathbf{a}' \sim \pr_{\mathbf{A}}}\bigl[\mathbb{E}_{\hat{\mathbf{Y}}^*_{\mathbf{a}'} \mid \mathbf{W}=\mathbf{w}}[\hat{\mathbf{y}}]\bigr]\Bigr]\:.
\end{equation}
That is, we quantify how the average outcome varies with the interventional value $\mathbf{a}'$ at conditioning value $\mathbf{w}$ and average this variance over $\mathbf{w}$. For deterministic predictors (point estimators), which we use in all our experiments, the prediction is a fixed value for each input $\mathbb{E}_{\hat{\mathbf{Y}}^*_{\mathbf{a}'} \mid \mathbf{W}=\mathbf{w}}[\hat{\mathbf{y}}] = \hat{\mathbf{y}}$) and we can drop the inner expectation of \cref{eq:vcf}. In this case, the variance term in \cref{eq:vcf} is zero if and only if $\pr_{\hat{\mathbf{Y}}^*_{\mathbf{a}} \mid \mathbf{W} = \mathbf{w}}(\mathbf{y}) = \pr_{\hat{\mathbf{Y}}^*_{\mathbf{a}'} \mid \mathbf{W} = \mathbf{w}}(\mathbf{y})$ almost surely. Since the variance is non-negative, the outer expectation is zero if and only if the variance term is zero almost surely, yielding the following result.
%
\begin{corollary}
  For point-estimators, $\hat{\mathbf{Y}}$ is counterfactually invariant in $\mathbf{A}$ w.r.t.\ $\mathbf{W}$ if and only if $\VCF{}(\hat{\mathbf{Y}}) = 0$ almost surely.
\end{corollary}
Estimating \VCF{} in practice requires access to the DGP to generate counterfactuals. Given $d$ i.i.d.\ examples $(\mathbf{w}_i)_{i=1}^d$ from a fixed observational dataset we sample $k$ intervention values from the marginal $\pr_{\mathbf{A}}$ and compute corresponding predictions. The inner expectation is simply the deterministic predictor output, and we compute the empirical expectation over the $d$ observed $\mathbf{w}$ values and empirical variances over the $k$ sampled interventional values (for each $\mathbf{w}$). Since the required counterfactuals are by their very nature unavailable in practice, our analysis of \VCF{} is limited to (semi-)synthetic settings. Notably, the proposed procedure for choosing $\gamma$ does not require $\VCF{}$. Our experiments corroborate the weak convergence property of \hsciconly{}---small \hsciconly{} implies small \VCF{}.
Hence, \hsciconly{} may serve as a strong proxy for \VCF{} and thus CI in practice.

In practical terms, HSCIC is considered the relevant metric, which is also estimable from data. Theoretically, however, counterfactual invariance is only implied by an exact HSCIC value of zero (\cref{cor:hscic_independence}). Even if HSCIC exhibits continuity, meaning it approaches zero as distributions converge weakly, VCF may still be preferable for directly assessing counterfactual invariance because it provides a more interpretable metric. Our experimental investigations into VCF aim to establish the practical utility of HSCIC as a measure of counterfactual invariance.

\begin{figure}
  \centering
    \includegraphics[width=0.325\linewidth]{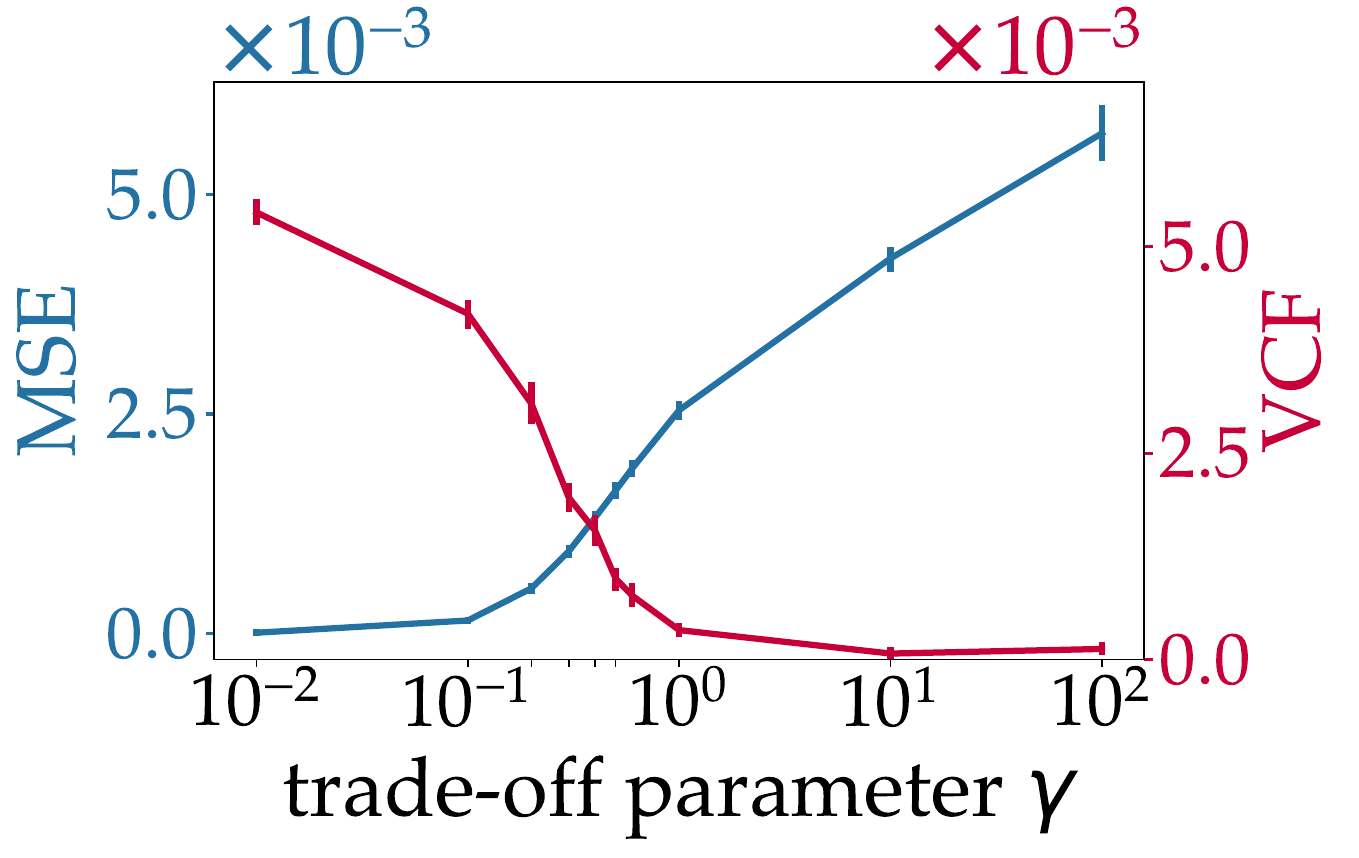}
    \includegraphics[width=0.325\linewidth]{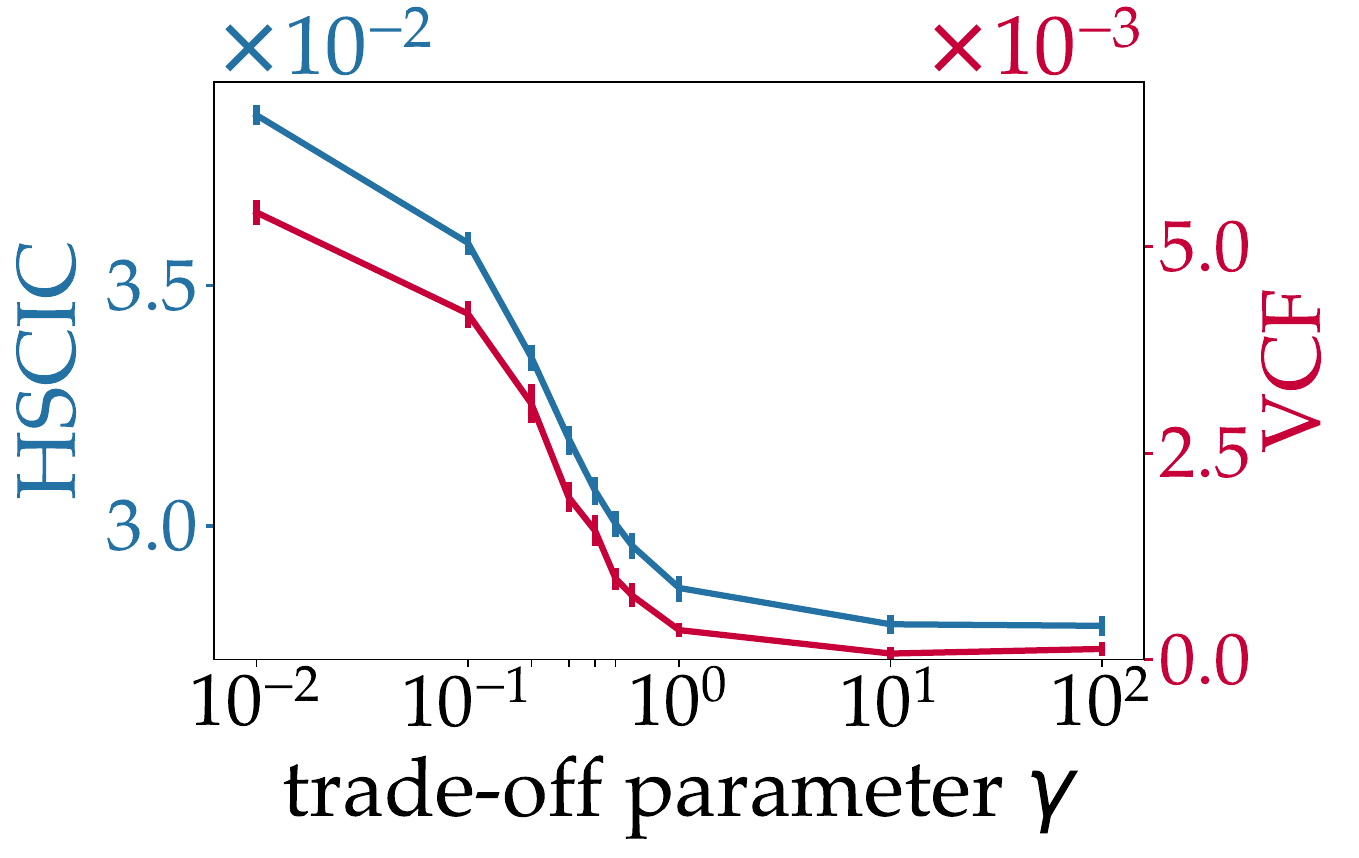}
    \includegraphics[width=0.31\linewidth]{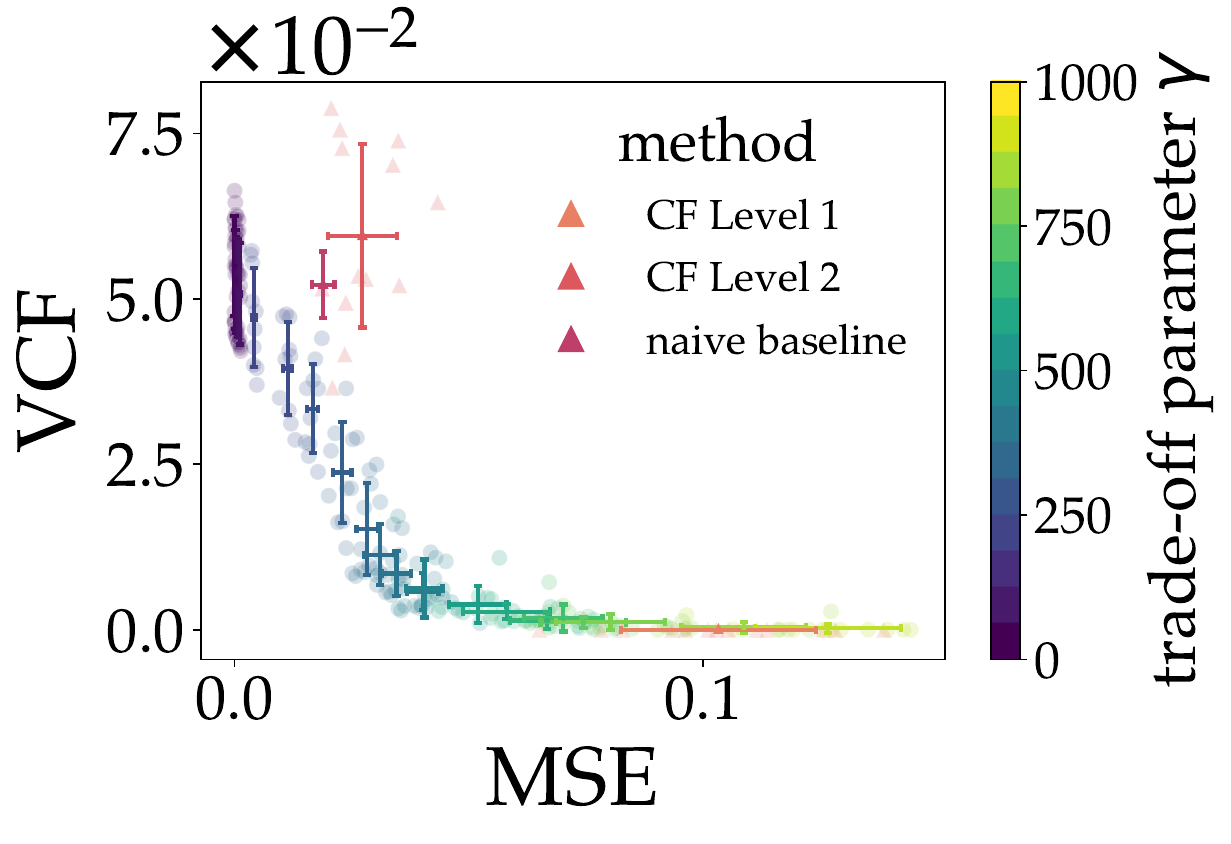}
  \caption{
  Results on synthetic data (see \cref{app:synthetic-experiment} and \cref{appendix:baselines}).
  \textbf{Left:} trade-off between MSE and counterfactual invariance (\VCF{}).
  \textbf{Middle:} strong correspondence between \hsciconly{} and \VCF{}.
  \textbf{Right:} performance of CIP against baselines CF1 and CF2 and the naive baseline. 
  As $\gamma$ increases, CIP traces out a frontier characterizing the trade-off between MSE and CI.
  CF2 and the naive baseline are Pareto-dominated by CIP, i.e., we can pick $\gamma$ to outperform CF2 in both MSE and \VCF{} simultaneously. CF1 has zero $\VCF{}$ by design, but worse predictive performance than CIP at near zero $\VCF{}$.
  Error bars are standard errors over 10 seeds.}\label{fig:simulated_exp}
\end{figure}
\section{Experiments}\label{sec:new_experiments}
The code for the experiments is available at: \url{https://github.com/ceciliacasolo/CIP}.
\paragraph{Baselines.} As many existing methods focus on cruder purely observational or interventional invariances (see \cref{sec:relatedwork}), our choice of baselines for true counterfactual invariance is highly limited.
First, we compare CIP to \citet{DBLP:conf/nips/VeitchDYE21} in their two limited settings (\cref{app:fig:graphs}(b-c)) in \cref{app:sec:baseline_veitch}, showing that our method performs on par with theirs.
Next, we compare to two methods proposed by \citet{DBLP:conf/nips/KusnerLRS17} in settings where they apply.
CF1 (their `Level 1') consists of only using non-descendants of $\mathbf{A}$ as inputs to $f_{\hat{\mathbf{Y}}}$.
CF2 (their `Level 2') assumes an additive noise model and uses the residuals of descendants of $\mathbf{A}$ after regression on $\mathbf{A}$ together with non-descendants of $\mathbf{A}$ as inputs to $f_{\hat{\mathbf{Y}}}$.
We refer to these two baselines as CF1 and CF2 respectively. We also compare CIP to the `naive baseline' which consists in training a predictor ignoring $\mathbf{A}$.
In settings where $\mathbf{A}$ is binary, we also compare to \citet{DBLP:conf/aaai/Chiappa19}, devised for path-wise counterfactual fairness.
Finally, we develop heuristics based on data augmentation as further possible baselines in \cref{app:sec:baseline_heuristic}.

\subsection{Synthetic experiments}\label{sec:synthetic_experiments}

First, we generate various synthetic datasets following the causal graph in \cref{fig:graph}(d).
They contain (i) the prediction targets $\mathbf{Y}$, (ii) variable(s) we want to be CI in $\mathbf{A}$, (iii) covariates $\mathbf{L}$ mediating effects from $\mathbf{A}$ on $\mathbf{Y}$, and (iv) confounding variables $\mathbf{S}$.
The goal is to learn a predictor $\hat{\mathbf{Y}}$ that is CI in $\mathbf{A}$ w.r.t.\ $\mathbf{W} \coloneqq \mathbf{A} \cup \mathbf{L} \cup \mathbf{S}$. 
The datasets cover different dimensions for the observed variables and their correlations and are described in detail in \cref{app:synthetic}.

\paragraph{Model choices and parameters.} For all synthetic experiments, we train fully connected neural networks (MLPs) with MSE loss $\mathcal{L}_{\textsc{mse}}( \hat{\mathbf{Y}})$ as the predictive loss $\mathcal{L}$ in \cref{eq:total_loss} for continuous outcomes $\mathbf{Y}$.
We generate $10$k samples from the observational distribution in each setting and use an $80$ to $20$ train-test split.
All metrics reported are on the test set.
We perform hyper-parameter tuning for MLP hyperparameters based on a random strategy (see \cref{app:synthetic} for details).
The $\hscic{\hat{\mathbf{Y}},\mathbf{A}\cup \mathbf{W}}{\mathbf{S}}$ term is computed as in \cref{eq:estimate_hscic} using a Gaussian kernel with amplitude $1.0$ and length scale $0.1$.
The regularization parameter $\lambda$ for the ridge regression coefficients is set to $\lambda = 0.01$.
We set $d=1000$ and $k=500$ in the estimation of \VCF{}.
%


\begin{figure}
  \centering
  \hspace{0.8cm}
  10-dimensional $\mathbf{A}$ \hspace{5.5cm} 50-dimensional $\mathbf{A}$\\[5mm]
  \includegraphics[height=3.7cm]{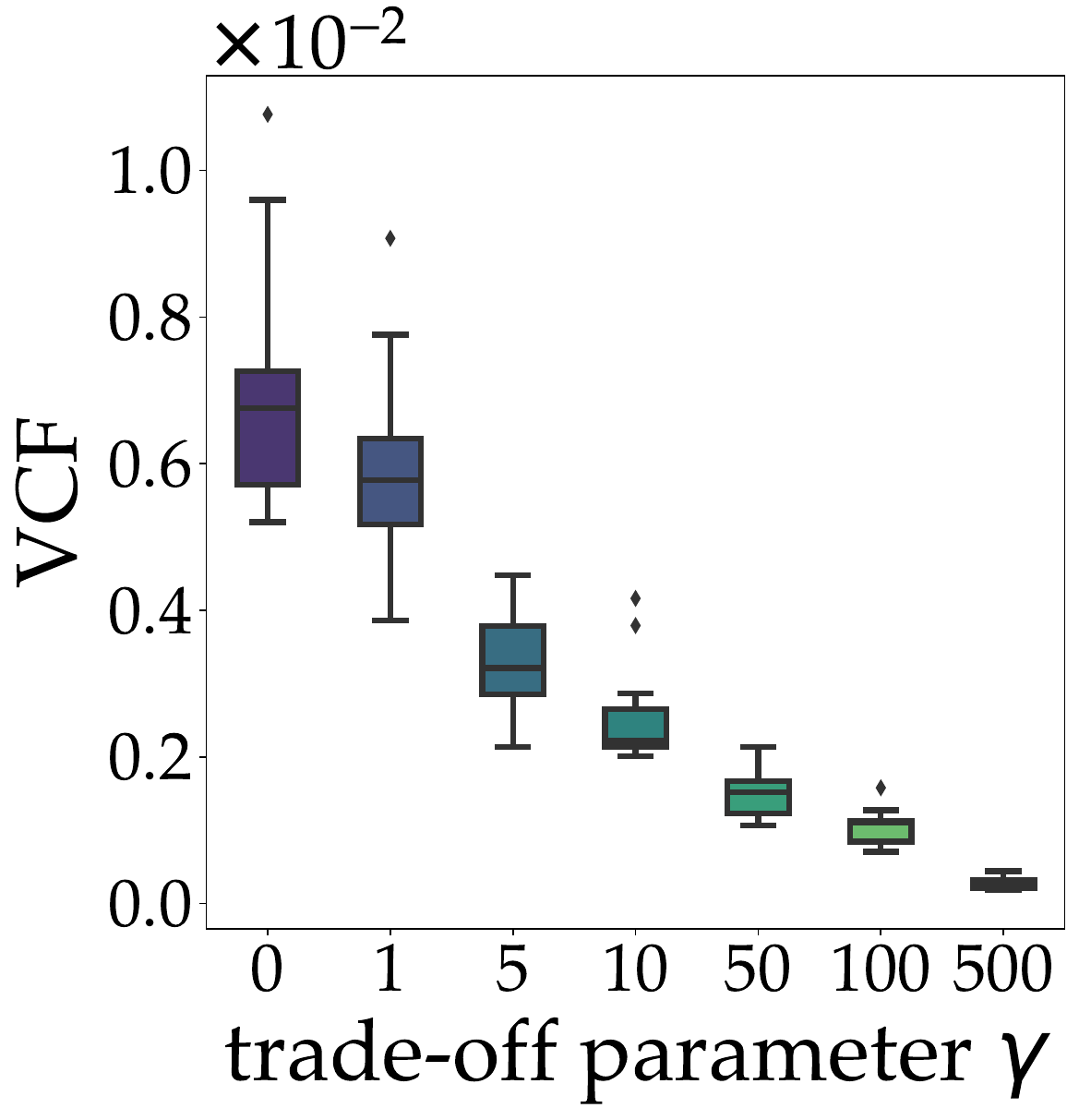}\hfill
    \includegraphics[height=3.7cm]{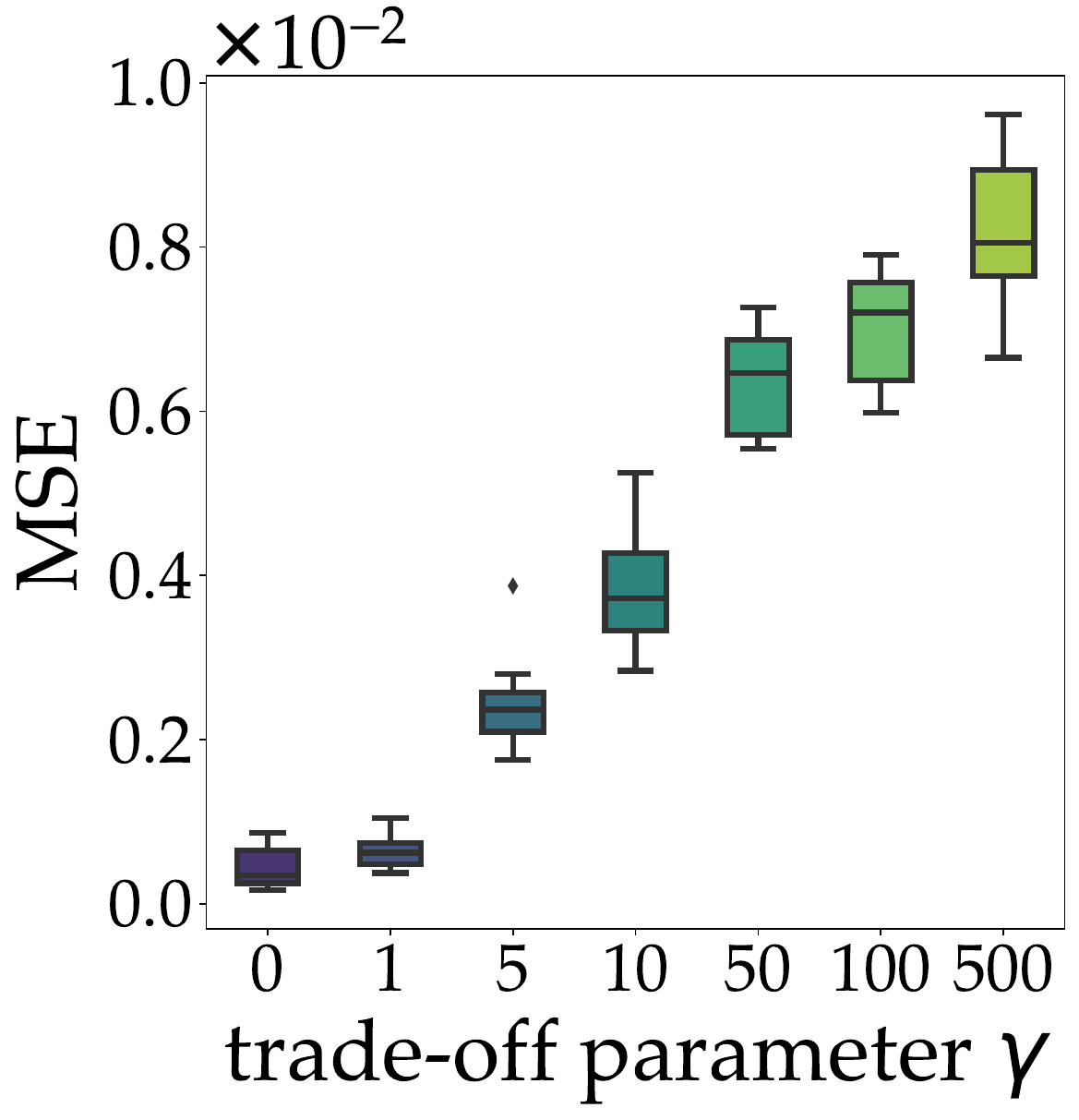}\hfill
  \includegraphics[height=3.7cm]{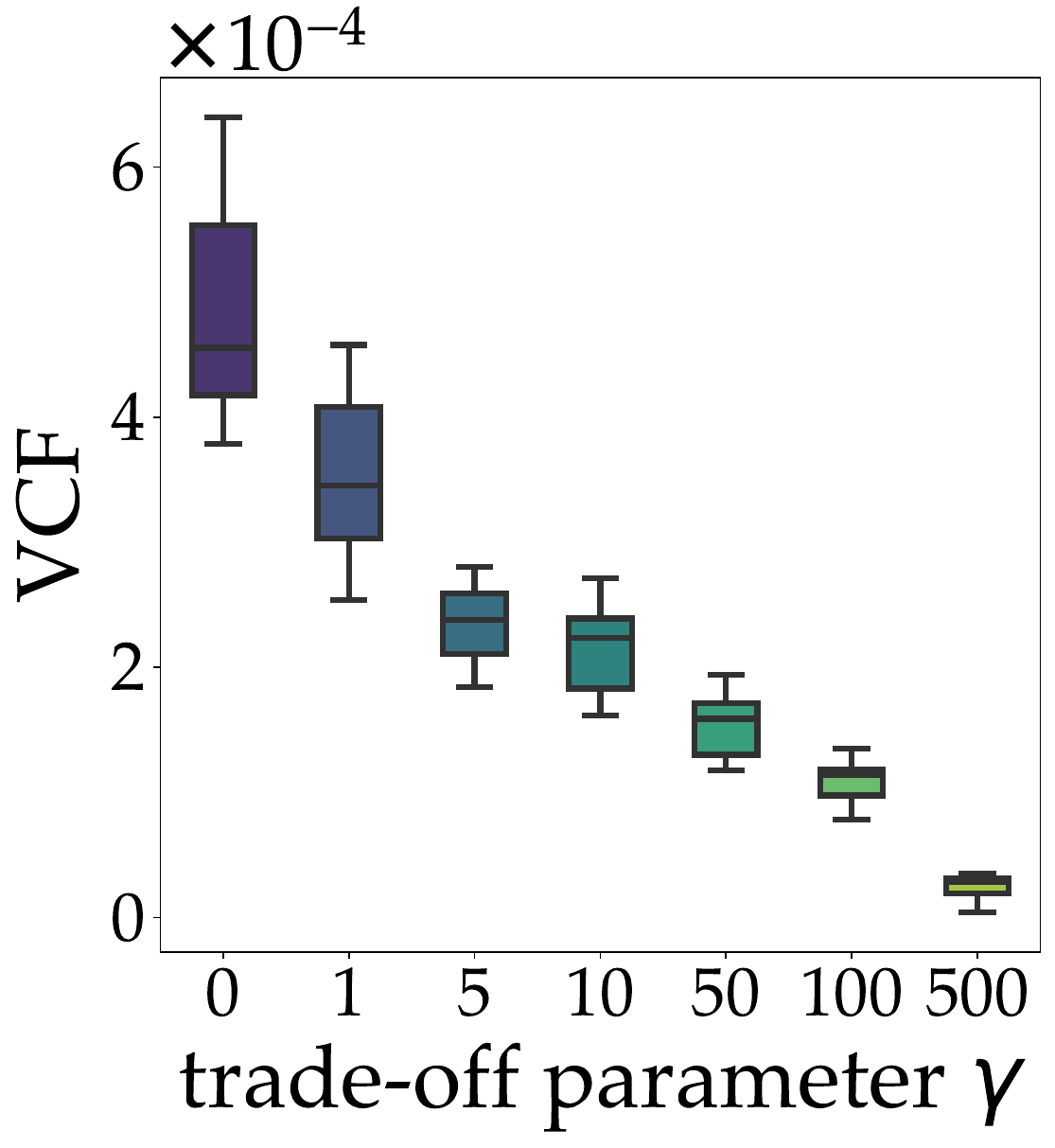}\hfill
  \includegraphics[height=3.7cm]{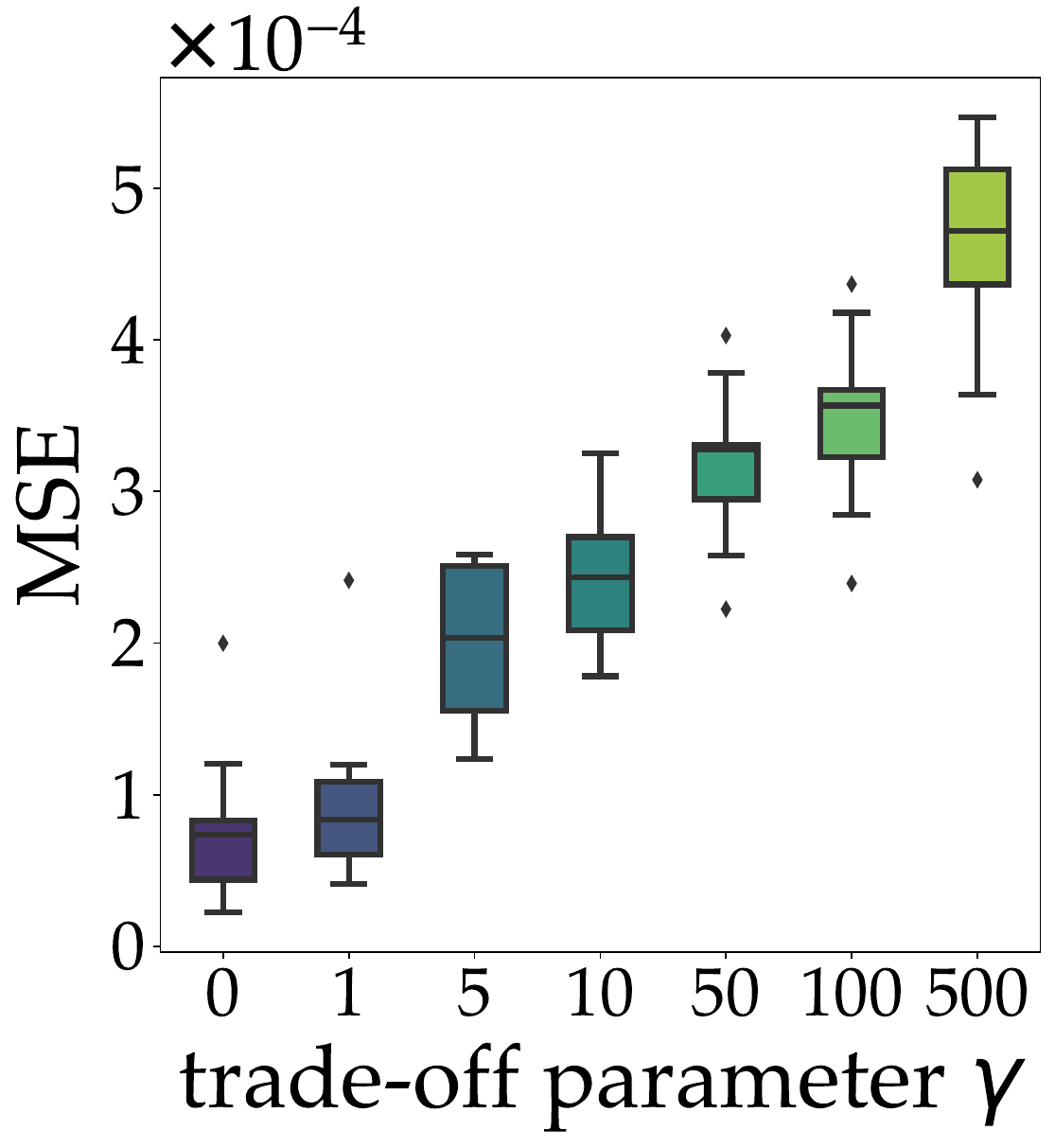}\hfill
  \caption{MSE and \VCF{} for synthetic data (\cref{appendix:multid_setting}) with 10- and 50-dimensional $\mathbf{A}$ for different $\gamma$ and 15 random seeds per box.
  CIP reliably achieves CI as $\gamma$ increases.}\label{fig:multidimensional_results_A}
\end{figure}

\paragraph{Model performance.} We first study the effect of the HSCIC on accuracy and counterfactual invariance on the simulated dataset in \cref{app:synthetic-experiment}.
\cref{fig:simulated_exp} (left) depicts the expected trade-off between MSE and \VCF{} for varying $\gamma$,
whereas \cref{fig:simulated_exp} (middle) highlights that \hsciconly{} (estimable from observational data) is a strong proxy of counterfactual invariance measured by \VCF{} (see discussion after \cref{eq:vcf}).
\Cref{fig:simulated_exp} (right) compares CIP to baselines for a simulated non-additive noise model in \cref{appendix:baselines}. 
For a suitable choice of $\gamma$, CIP outperforms the baseline CF2 and the naive baseline in both MSE and \VCF{} simultaneously.
While CF1 achieves perfect CI by construction ($\VCF{} = 0$), its MSE is higher than CIP at almost perfect CI ($\VCF{}$ near zero).
To conclude, our method flexibly and reliably trades predictive performance for counterfactual invariance via a single parameter $\gamma$ and Pareto-dominates existing methods.
In \cref{appendix:baselines} we present extensive results on further simulated settings and compare CIP to other heuristic methods in \cref{app:sec:baseline_heuristic}.

\begin{wrapfigure}{r}{0.53\textwidth}
  \vspace{-5mm}
  \centering
  \includegraphics[height=3.9cm]{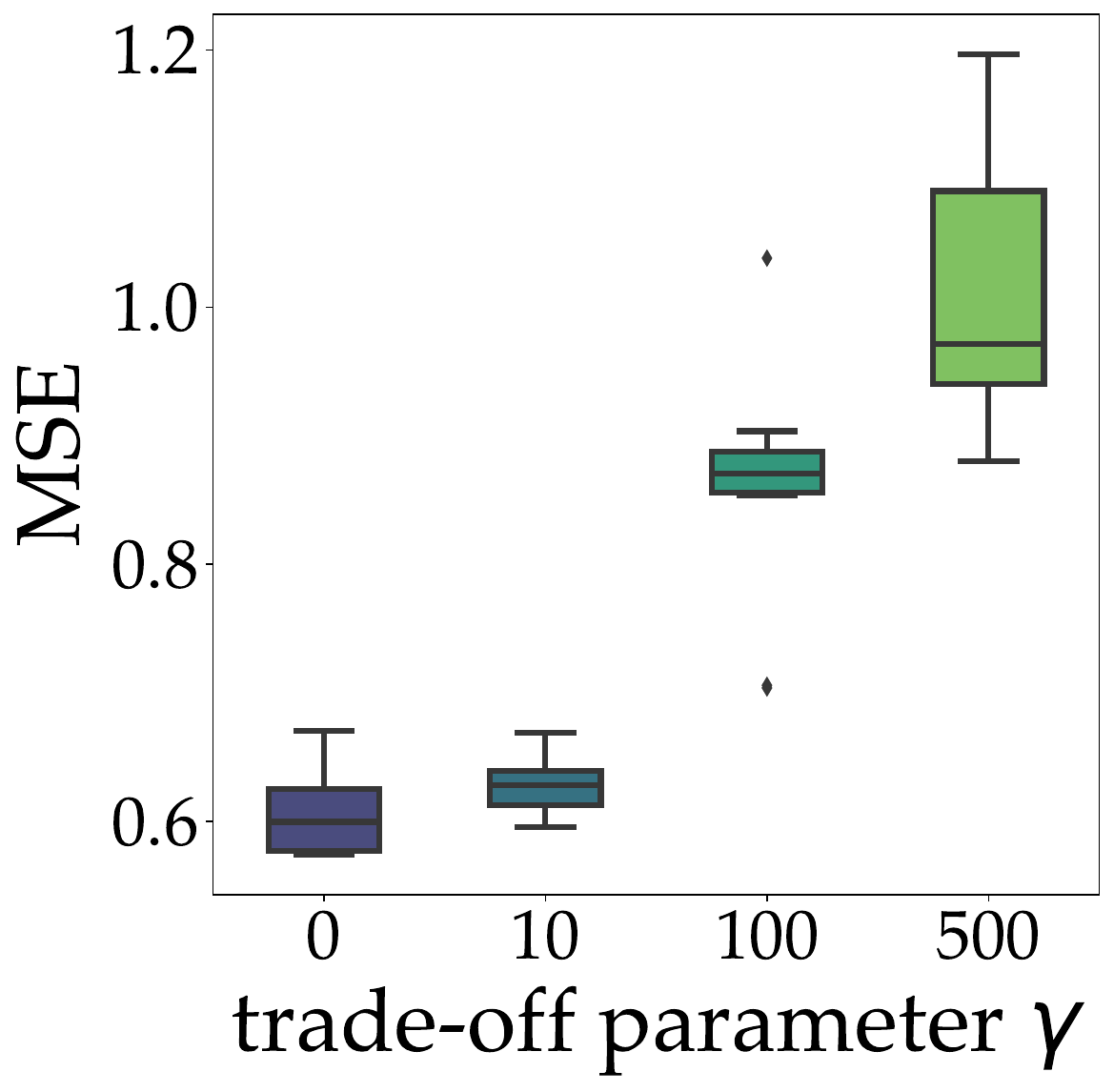}
  \includegraphics[height=3.9cm]{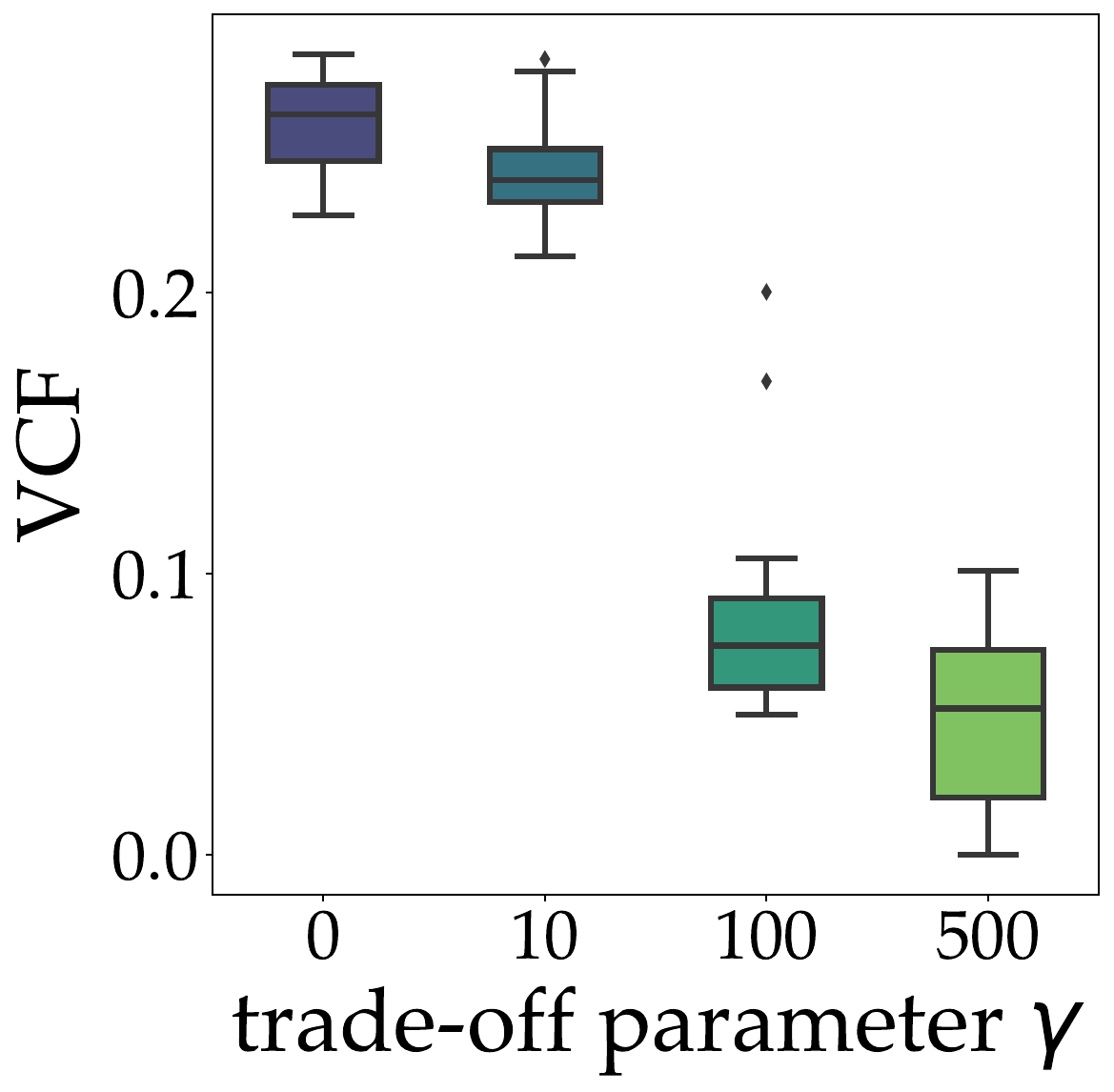}
  \caption{On the dSprites image dataset, CIP trades off MSE for \VCF{} and achieves almost full CI as $\gamma$ increases. Boxes are for 8 random seeds.}\label{fig:image_exp}
  \vspace{-6mm}
\end{wrapfigure}

\paragraph{Effect of dimensionality of $\mathbf{A}$.}
A key advantage of CIP is that it can deal with multi-dimensional $\mathbf{A}$.
We consider simulated datasets described in \cref{appendix:multid_setting}, where we gradually increase the dimension of $\mathbf{A}$.
The results in \cref{fig:multidimensional_results_A} for different trade-off parameters $\gamma$ and different dimensions of $\mathbf{A}$ demonstrate that CIP effectively enforces CI also for multi-dimensional $\mathbf{A}$.\footnote{In all boxplots, boxes represent the interquartile range, the horizontal line is the median, and whiskers show minimum and maximum values, excluding outliers (dots).}
Further results are shown in \cref{appendix:multid_setting}.
\subsection{Image experiments}\label{sec:imagedata}
We consider an image classification task on the dSprites dataset \citep{dsprites17}, with a causal model as depicted in \cref{fig:graph}(f).
The full structural equations are provided in \cref{app:sec:imageresults}.
This experiment is particularly challenging due to the mixed categorical and continuous variables in $\mathbf{C}$ (\texttt{shape}, \texttt{y-pos}) and $\mathbf{L}$ (\texttt{color}, \texttt{orientation}), with continuous $\mathbf{A}$ (\texttt{x-pos}).
We seek a predictor $\hat{\mathbf{Y}}$ that is CI in the x-position w.r.t.\ all other observed variables.
Following \cref{lemma:conditiona_counterfactual}, we achieve $\hat{\mathbf{Y}} \indep \{\texttt{x-pos}, \texttt{scale}, \texttt{color}, \texttt{orientation}\} \mid \{\texttt{shape}, \texttt{y-pos}\}$ via the \hsciconly{} operator.
To accommodate the mixed input types, we first extract features from the images via a CNN and from other inputs via an MLP.
We then use an MLP on all concatenated features for $\hat{\mathbf{Y}}$.
\cref{fig:image_exp} shows that CIP gradually enforces  CI as $\gamma$ increases and illustrates the inevitable increase of MSE.
\subsection{Fairness with continuous protected attributes}\label{sec:fairnessresults}
Finally, we apply CIP to the widely-used UCI Adult dataset \citep{uci_dataset} and compare it against a `naive baseline' which simply ignores $\mathbf{A}$, CF1, CF2, and path-specific counterfactual fairness (PSCF) \citep{DBLP:conf/aaai/Chiappa19}. Like CF1, PSCF always achieves $\VCF{}=0$ by design.
\begin{wrapfigure}{r}{0.54\textwidth}
  \centering
  \includegraphics[height=3.9cm]{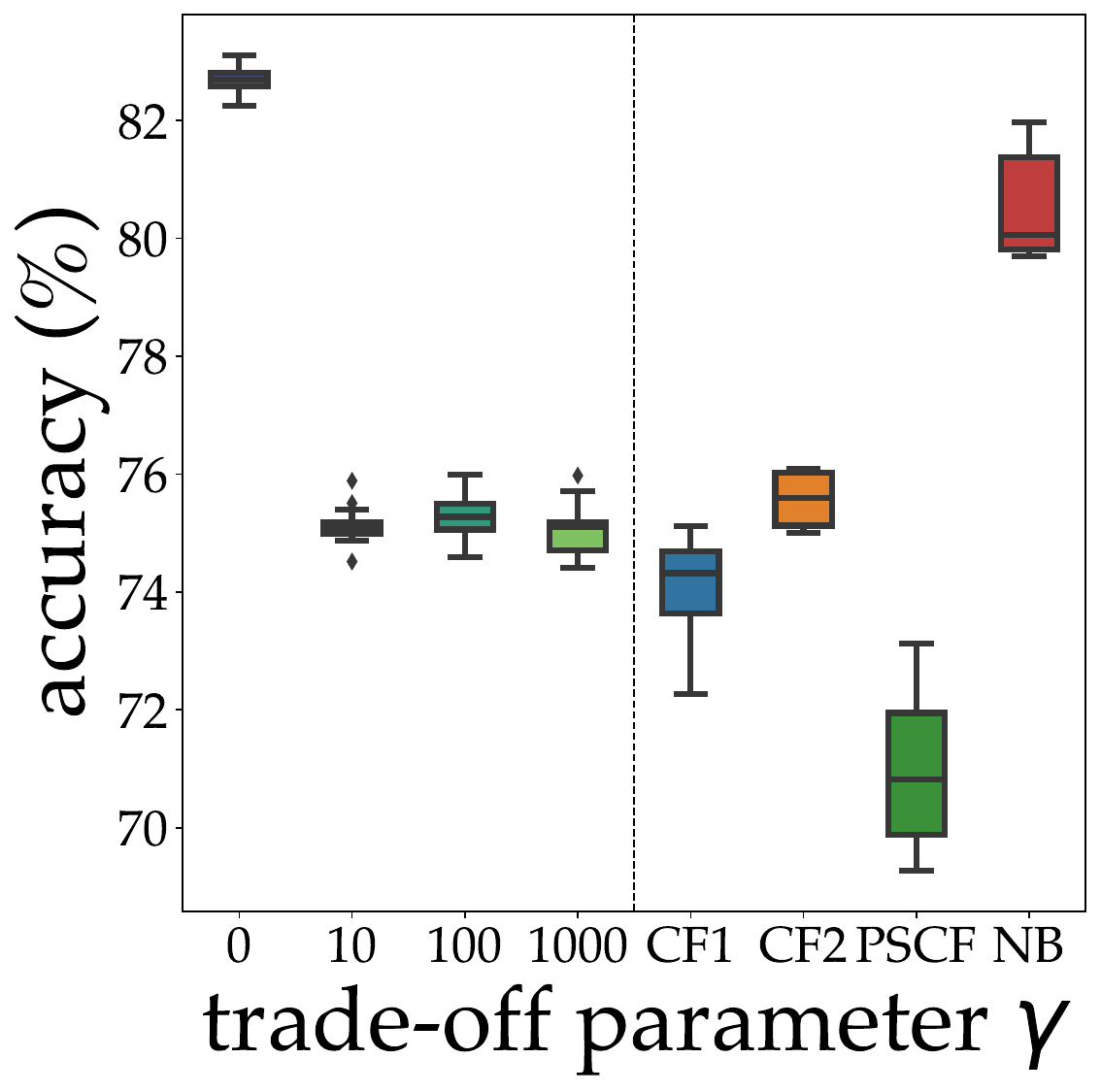}
  \includegraphics[height=3.9cm]{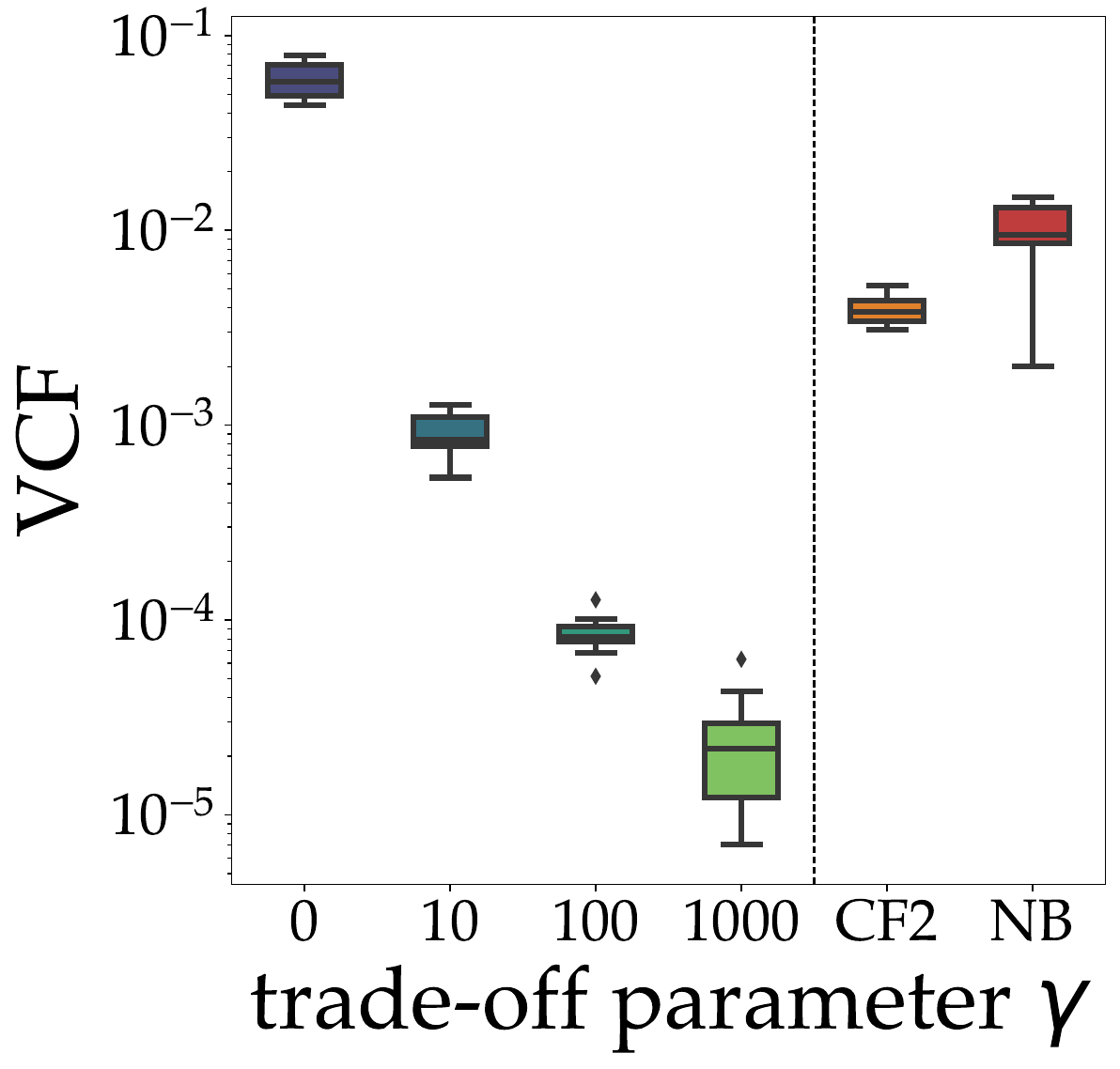}
  \caption{Accuracy and \VCF{} on the Adult dataset. CIP achieves better \VCF{} than CF2 and the naive baseline (NB), improved in accuracy compared to PSCF and is on par with CF1 in accuracy at $\VCF{} \approx 0$. }\label{fig:adult_experiment}
  \vspace{-4mm}
\end{wrapfigure}
We explicitly acknowledge the shortcomings of the UCI Adult dataset to reason about social justice \citep{ding2021retiring}.
Instead, we chose it due to previous investigations into plausible causal structures based on domain knowledge \citep{DBLP:conf/kdd/ZhangWW17}.
The task is to predict whether an individual's income is above a threshold based on demographic information, including protected attributes.
We follow \citet{nabi2018fair,DBLP:conf/aaai/Chiappa19}, where a causal structure is assumed for a subset of the variables as in \cref{fig:graph}(e) (see \cref{app:sec:adult} and \cref{fig:causal_graph_uci} for details).
We choose gender and age as the protected attributes $\mathbf{A}$, collect marital status, level of education, occupation, working hours per week, and work class into $\mathbf{L}$, and combine the remaining observed attributes in $\mathbf{C}$.
Our aim is to learn a predictor $\hat{\mathbf{Y}}$ that is CI in $\mathbf{A}$ w.r.t.\ $\mathbf{W} = \mathbf{C} \cup \mathbf{L}$.
Achieving (causal) fairness for (mixed categorical and) continuous protected attributes is under active investigation  \citep{Mary19:Fairness,chiappa2021fairness}, but directly supported by CIP.

We use an MLP with binary cross-entropy loss for $\hat{\mathbf{Y}}$.
Since this experiment is based on real data, the true counterfactual distribution cannot be known.
Following \citet{chiappa2021fairness} we estimate a possible SCM by inferring the posterior distribution over the unobserved variables using variational autoencoders \citep{KingmaW13}.
Even though CF1, PSCF always achieves VCF = 0 by design, \Cref{fig:adult_experiment} shows that CIP gradually achieves CI and even manages to keep a constant accuracy after an initial drop.
It outperforms PSCF in accuracy and reaches comparable accuracy to CF1 when reaching $\VCF{} \approx 0$. It Pareto-dominates CF2 and can achieve better \VCF{} than the naive baseline, even though NB is more accurate for $\gamma \ge 5$ than CIP.
\section{Discussion and future work}\label{sec:discussion}
We developed CIP, a method to learn counterfactually invariant predictors $\hat{\mathbf{Y}}$. First, we presented a sufficient graphical criterion to characterize counterfactual invariance and reduced it to conditional independence in the observational distribution under an injectivity assumption of a causal mechanism. We then built on kernel mean embeddings and the Hilbert-Schmidt Conditional Independence Criterion to devise an efficiently estimable, differentiable, model-agnostic objective to train CI predictors for mixed continuous/categorical, multi-dimensional variables.
We demonstrated the efficacy of CIP in extensive empirical evaluations on various regression and classification tasks.

A limitation of this work is the computational cost of estimating \hsciconly{}, which may limit the scalability of CIP to very high-dimensional settings even when using efficient random Fourier features. While the increased robustness of counterfactually invariant predictors are certainly desirable in many contexts, this presupposes the validity of our assumptions. Thus, an important direction for future work is to assess the sensitivity of CIP to misspecifications of the causal graph or insufficient knowledge of the required adjustment set.
Lastly, we envision our graphical criterion and KME-based objective to be useful also for causal representation learning to isolate causally relevant, autonomous factors underlying the data.

Another key limitation of our work, shared by all studies in this domain, is the assumption that the causal graph is known. Assessing the validity of a causal DAG in real-world settings is challenging, since one can often conjecture plausible confounding mechanisms for any missing edge. For this reason, in this work we found it challenging to provide convincing examples that corroborate the “potential validity” of our assumptions. However, our empirical results demonstrate that CIP is effective even when these assumptions are violated. Similarly, our real-world experiments show a comparable invariance and accuracy trade-off trend as the simulated experiments. We believe this to be a better indicator of validity and robustness than isolated toy examples where assumptions may be satisfied. Practitioners using our proposed framework are recommended to take these considerations into account before they apply the method to their own data, and combine causal discovery methods with domain knowledge to assess the validity of the underlying DGP.
\section*{Acknowledgements}
Francesco Quinzan acknowledges funding from ELSA: European Lighthouse on Secure and Safe AI project (grant agreement
No. 101070617 under UK guarantee). Cecilia Casolo is supported by the DAAD programme Konrad Zuse Schools of Excellence in Artificial Intelligence, sponsored by the Federal Ministry of Education and Research.
\bibliographystyle{apalike}
\bibliography{bibliography}
\newpage
\renewcommand{\thesection}{\Alph{section}}
\setcounter{section}{0}
\noindent {\LARGE\textbf{Appendix}}
\section{Discussion on related work}\label{app:related_work}
In this section, we give an overview of the definitions used in the related work \citet{fawkes2023results}.

\begin{definition}[Def.~2.1 by \citet{fawkes2022selection}]\label{def:fawkes-a.s.}
    A variable $Y$ satisfies \textbf{almost sure counterfactual invariance} (a.s.-CI) with respect to $A$ if:
\begin{equation*}
    Y_a = Y_{a'} \mbox{ a.s. for all } a, a'
\end{equation*}
\end{definition}

\begin{definition}[Def.~2.3 by \citet{fawkes2023results} and Def.~1.1 \citet{DBLP:conf/nips/VeitchDYE21}]\label{def:fawkes-F}
    A function $f: \mathcal{X} \rightarrow  \mathcal{Y}$ is \textbf{counterfactually invariant} ($\mathcal{F}$-CI) if the variable $\hat{Y}:=f(X)$ satisfies almost sure counterfactual invariance. That is:
\begin{equation*}
f(X_a)=f(X_{a'}) \mbox{ a.s.}
\end{equation*}
\end{definition}

\begin{definition}[Def.~2.2 by \citet{fawkes2023results}]\label{def:fawkes-D}
    A variable $Y$ satisfies \textbf{distributional counterfactually invariance} ($\mathcal{D}$-CI) conditional on some set of variables $W$, with respect to $A$ , if:
\begin{equation*}
P(Y_a=y | W=w, A=a)=P(Y_{a'}=y | W=w, A=a)
\end{equation*}
for all $a$, $a'$ and for almost all $y$, $w$.
\end{definition}
\Cref{def:fawkes-D} is similar to the one provided in the main paper \Cref{def:counterfactual_invariance}, however enforces conditioning on the intervening variable $A$. As it can be noticed from the definitions, distributional notions of Counterfactual Invariance (CI), such as \Cref{def:fawkes-D} and \Cref{def:counterfactual_invariance}, are less stringent than \Cref{def:fawkes-a.s.}, as indicated in the following result in \citet[Lem.~2.4]{fawkes2023results}:
\begin{lemma}[Lem.~2.4 by \citet{fawkes2023results}]
    We have that a.s.-CI implies $\cD$-CI conditional on any set of variables, but $\cD$-CI implies a.s.-CI only if the conditioning set contains the outcome $Y$.
\end{lemma}

\section{Proof of \cref{lemma:conditiona_counterfactual}}\label{proof:conditiona_counterfactual}
To provide a mostly self-contained proof, we start by stating the necessary definitions and repeat required known results from the literature. \Cref{app:a1,app:a2,app:a3} contain these preliminary definitions and results, whereas the final statement and proof provided in \cref{app:a4} is our original contribution (making use of the preliminaries stated before).
\subsection{Overview of the proof techniques}\label{app:a1}
We restate the main theorem for  completeness.
\condcounterfactual*
Our proof technique generalizes the work of \citet{DBLP:conf/uai/ShpitserP09}. To understand the proof technique, note that conditional counterfactual distributions of the form $\pr_{\mathbf{Y}^*_{\mathbf{a}}\mid \mathbf{W}}(\mathbf{y} \mid \mathbf{w})$ involve quantities \emph{from two different worlds}. The variables $\mathbf{W}$ belong to the pre-interventional world, and the interventional variable $\mathbf{Y}_{\mathbf{a}}$ belongs to the world after performing the intervention $\mathbf{A}\gets \mathbf{a}$. The variable $\mathbf{Y}^*_{\mathbf{a}\mid \mathbf{W}}$ refers to the counterfactual of $\mathbf{Y}_{\mathbf{a}}$ after conditioning in the pre-interventional world on $\mathbf{W}$. 

\subsection{Identifiability of counterfactual distributions}\label{app:a2}

In this section, we discuss the well-known backdoor criterion for the identifiability of conditional distributions, which we will then use to prove \cref{lemma:conditiona_counterfactual}. To this end, we use the notions of a blocked path and valid adjustment set, which we restate for clarity.
\begin{definition}
\label{def:blocked_path}
Consider a path $\pi$ of causal graph $\mathcal{G}$. A set of nodes $\mathbf{Z}$ blocks $\pi$, if $\pi$ contains a triple of consecutive nodes connected in one of the following ways: $N_i \rightarrow Z \rightarrow N_j $, $N_i \gets Z \rightarrow N_j $, with $N_i, N_j \notin \mathbf{Z}$, $Z \in \mathbf{Z}$, or $N_i \rightarrow M \gets N_j $ and neither $M$ nor any descendent of $M$ is in $\mathbf{Z}$.
\end{definition}
Using this definition, we define the concept of a valid adjustment set.
\validadjset*
\Cref{def:valid_adjustment set} is a useful graphical criterion for the identifiability of counterfactual distributions. In fact, the following theorem holds:
\begin{theorem}[Theorem 4 by \citet{DBLP:conf/uai/ShpitserVR10}]
\label{eq:new_lemma2} 
Let $\mathcal{G}$ be a causal graph, $\mathbf{A}$, $\mathbf{W}$ be two (not necessarily disjoint) sets of nodes in $\mathcal{G}$, such that $(\mathbf{A} \cup \mathbf{W}) \cap \mathbf{Y} = \emptyset$. Define the set $\mathbf{X} = \mathbf{W}\setminus \mathbf{A}$. Suppose that a set of nodes $\mathbf{S}$ satisfies the adjustment criterion relative to $(\mathbf{A \cup \mathbf{W}}, \mathbf{Y})$ in $\mathcal{G}$. Then, it holds $\mathbf{Y}^*_{\mathbf{a}', \mathbf{x}'} \indep \mathbf{A} , \mathbf{X}  \mid \mathbf{S}$ for any intervention $\mathbf{A}, \mathbf{X} \gets \mathbf{a}', \mathbf{x}'$.
\end{theorem}

\subsection{$d$-separation and conditional independence}\label{app:a3}

In this section, we discuss a well-known criterion for conditional independence, which we will then use to prove \cref{lemma:conditiona_counterfactual}. We use the notion of a blocked path, as in \Cref{def:d_separation} and the concept of $d$-separation as follows.
\begin{definition}[$d$-Separation]
\label{def:d_separation}
Consider a causal graph $\mathcal{G}$.  Two sets of nodes $\mathbf{X}$ and $\mathbf{Y}$ of $\mathcal{G}$ are said to be $d$-separated by a third set $\mathbf{S}$ if every path from any node of $\mathbf{X}$ to any node of $\mathbf{Y}$ is blocked by $\mathbf{S}$.
\end{definition}
We use the notation $\mathbf{X} \indep_{\mathcal{G}} \mathbf{Y}\mid \mathbf{S}$ to indicate that $\mathbf{X}$ and $\mathbf{Y}$ are $d$-separated by $\mathbf{S}$ in $\mathcal{G}$. We use \Cref{def:d_separation} as a graphical criterion for conditional independence \citep{pearlj}.
\begin{lemma}[Markov Property]
\label{lemma:d_separation}
Consider a causal graph $\mathcal{G}$, and suppose that two sets of nodes $\mathbf{X}$ and $\mathbf{Y}$ of $\mathcal{G}$ are $d$-separated by $\mathbf{S}$. Then, $\mathbf{X}$ is independent of $\mathbf{Y}$ given $\mathbf{S}$ in any model induced by the graph $\mathcal{G}$.
\end{lemma}
The Markov Property is also referred to as $d$-separation property. We use the notation $\mathbf{X} \indep_{\mathcal{G}} \mathbf{Y}\mid \mathbf{S}$ to indicate that $\mathbf{X}$ and $\mathbf{Y}$ are $d$-separated by $\mathbf{S}$ in $\mathcal{G}$. 

\subsection{Proof of \cref{lemma:conditiona_counterfactual}}\label{app:a4}

\begin{figure*}[t!]
\centering
\begin{tikzpicture}[node distance=6mm and 8mm, main/.style = {draw, circle, minimum size=0.6cm, inner sep=1pt}, >={triangle 45}] 
    \node[main] (1) {$\scriptstyle\mathbf{A}$}; 
    \node[main] (2) [right =of 1] {$\scriptstyle\mathbf{X_1}$}; 
    \node[main] (3) [right =of 2] {$\scriptstyle\mathbf{Y}$}; 
    \node[main] (4) [above =of 2] {$\scriptstyle\mathbf{X_2}$};
    \node (label) [left =of 4] {\textbf{(a)}};
    \draw[draw=red,->] (1) -- (2); 
    \draw[->] (2) -- (3); 
    \draw[->] (4) -- (1);
    \draw[->] (4) -- (2);
    \draw[->] (4) -- (3);
  \end{tikzpicture}
  \hfill
\begin{tikzpicture}[node distance=6mm and 8mm, main/.style = {draw, circle, minimum size=0.6cm, inner sep=1pt}, >={triangle 45}] 
    \node[main] (1) {$\scriptstyle\mathbf{A}$}; 
    \node[main] (2) [right =of 1] {$\scriptstyle\mathbf{X_3}$}; 
    \node[main] (3) [right =of 2] {$\scriptstyle\mathbf{Y}$}; 
    \node[main] (4) [above =of 2] {$\scriptstyle\mathbf{X_2}$};
    \node[main] (5) [above =of 1] {$\scriptstyle\mathbf{X_1}$};
    \node (label) [right =of 4] {\textbf{(b)}};
    \draw[->] (1) -- (2); 
    \draw[draw=red,->] (1) -- (5);
    \draw[->] (2) -- (3); 
    \draw[draw=red,->] (4) -- (2);
    \draw[draw=red,->] (5) -- (4);
    \draw[->] (5) -- (2);
\end{tikzpicture}
\hfill
\begin{tikzpicture}[node distance=6mm and 8mm, main/.style = {draw, circle, minimum size=0.6cm, inner sep=1pt}, square/.style = {draw, rectangle, minimum size=0.6cm, inner sep=1pt}, >={triangle 45}] 
    \node[main] (1) {$\scriptstyle\mathbf{A}$}; 
    \node[main] (2) [right =of 1] {$\scriptstyle\mathbf{X_2}$}; 
    \node[main] (3) [right =of 2] {$\scriptstyle\mathbf{Y}$}; 
    \node[main] (4) [above =of 2] {$\scriptstyle\mathbf{X_1}$};
    \node (label) [left =of 4] {\textbf{(c)}};
    \draw[->] (2) -- (1); 
    \draw[->] (2) -- (3); 
    \draw[->] (4) -- (2);
    \draw[->] (4) -- (1);
    \draw[->] (4) -- (3);
\end{tikzpicture}%
\caption{(a) In this causal graph the elements of the set $\mathbf{X}=\{\mathbf{X_1}, \mathbf{X_2}\}$ have depth $\mathsf{depth}(\mathbf{X_1}) = 1$ and $\mathsf{depth}(\mathbf{X_2}) = 0$. (b) In this causal graph the elements of the set $\mathbf{X}=\{\mathbf{X_1}, \mathbf{X_2}, \mathbf{X_3}\}$ have depth $\mathsf{depth}(\mathbf{X_1}) = 1$, $\mathsf{depth}(\mathbf{X_2}) = 2$, $\mathsf{depth}(\mathbf{X_3}) = 3$. (c) In this causal graph the elements of the set $\mathbf{X}=\{\mathbf{X_1}, \mathbf{X_2}\}$ have depth $\mathsf{depth}(\mathbf{X_1}) = \mathsf{depth}(\mathbf{X_2}) = 0$.}
\label{app:fig:graphs}
\end{figure*}
Before discussing the proof of \cref{lemma:conditiona_counterfactual}, we prove an additional auxiliary result. All the following theoretical results are novel and based on the previous sections.
\begin{lemma}
\label{eq:new_lemma_correction1} 
Let $\mathcal{G}$ be a causal graph, $\mathbf{A}$, $\mathbf{X}$ be two disjoint sets of nodes in $\mathcal{G}$. Suppose that any r.v. $X \subseteq \mathbf{X}$ is defined by a structural equation of the form $X = g(\mathsf{pa}(X), \mathbf{U_{X}})$, with $\mathsf{pa}(X) \subseteq \mathbf{X}\cup \mathbf{A}$ and $\mathbf{U_{X}}$ exogenous independent noise. Furthermore, suppose that $g$ is injective in $\mathbf{U_{X}}$. Then, for any observational values $\mathbf{a}, \mathbf{x}$ with $\mathbb{P}_{\mathbf{X},\mathbf{A}}(\mathbf{x}, \mathbf{a})\neq 0$,  there exist a value $\mathbf{u}$ in the support of $\mathbf{U_{X}}$ such that $\pr_{\mathbf{U_{X}}\mid \mathbf{A}=\mathbf{a} ,\mathbf{X}=\mathbf{x}} (\mathbf{u})=\mathbb{1}_{\mathbf{u}}$. Here, $\mathbb{1}_{\mathbf{u}}$ is the Dirac $\delta$-distribution.%
\end{lemma}
\begin{proof}
In this proof, we use the following additional notation. Consider a set of nodes $\mathbf{T}\subseteq \mathbf{A} \cup \mathbf{X}$, and let $\mathbf{a}$ be a point in the support of $\mathbf{A}$. We denote with $\mathbf{a}_{\mathbf{T}}$ the restriction of $\mathbf{a}$ to $\mathbf{A}\cap \mathbf{T}$. Similarly, for a point $\mathbf{x}$ in the support of $\mathbf{X}$, we denote with $\mathbf{x}_{\mathbf{T}}$ the restriction of $\mathbf{a}$ to $\mathbf{X}\cap \mathbf{T}$. Furthermore, given two points $\mathbf{a}$, $\mathbf{x}$ as above, we define $\mathbf{a}\cup \mathbf{x}$ as the only point in the support of $\mathbf{A}\cup \mathbf{X}$ such that $(\mathbf{a}\cup \mathbf{x})\mid_{\mathbf{A}} = \mathbf{a}$ and $(\mathbf{a}\cup \mathbf{x})\mid_{\mathbf{X}} = \mathbf{x}$ respectively.

Furthermore, we denote with $\mathbf{X} = g_{\mathbf{X}}(\mathsf{pa}(\mathbf{X}), \mathbf{U}_{\mathbf{X}})$ the structural equation for the joint random variable $\mathbf{X}$. In this equation, $\mathbf{U}_{\mathbf{X}}$ corresponds to the joint exogenous noise. Note that the function $g_{X}(\mathsf{pa}(X), \mathbf{U}_{\mathbf{X}})$ is injective in $\mathbf{U}_{\mathbf{X}}$, by the injectivity assumption on the structural equation $g_{X}(\mathsf{pa}(X), \mathbf{U_{X}})$ for each $X \subseteq \mathbf{X}$. Define the function $h(\mathbf{p})$ over the support of $\mathsf{pa}(\mathbf{X})$ as
\begin{equation*}
    h(\mathbf{a},\mathbf{x}) \coloneqq 
    \left \{
    \begin{array}{ll}
\mathbf{u} \ \text{such that} \ g_{\mathbf{X}}((\mathbf{a}\cup\mathbf{x})\mid_{\mathsf{pa}(\mathbf{X})}, \mathbf{u}) = \mathbf{x} & \text{if} \ \mathbb{P}_{\mathsf{pa}(\mathbf{X}), \mathbf{U}_{\mathbf{X}}}((\mathbf{a}\cup\mathbf{x})\mid_{\mathsf{pa}(\mathbf{X})}, \mathbf{u})\neq 0 \\
0 & \text{otherwise}
    \end{array}
    \right .
\end{equation*}
Note that by the injectivity of $g_{\mathbf{X}}$, the function $h$ as above is well-defined. Since $\mathsf{pa}(\mathbf{X}) \subseteq \mathbf{X}\cup \mathbf{A}$, then it holds $\pr_{\mathbf{U}_\mathbf{X}\mid \mathbf{A}=\mathbf{a} ,\mathbf{X}=\mathbf{x}} (\mathbf{u})
=\mathbb{1}_{h(\mathbf{a},\mathbf{x})}$. The claim follows, by defining $\mathbf{u} \coloneqq h(\mathbf{a},\mathbf{x})$.
\end{proof}
We use lemma \ref{eq:new_lemma_correction1} to prove the following result.
\begin{lemma}
\label{eq:new_lemma_correction2} 
Under the assumptions of Lemma \ref{eq:new_lemma_correction1}, fix a set $\mathbf{T}\subseteq \mathbf{A}\cup\mathbf{X}$. Then, for any intervention $\mathbf{A} \gets \mathbf{a}'$ and observational values $\mathbf{a}, \mathbf{x}$ with $\mathbb{P}_{\mathbf{X},\mathbf{A}}(\mathbf{x}, \mathbf{a})\neq 0$,  there exist a point $\mathbf{t}$ in the support of $\mathbf{T}$ such that $\pr_{\mathbf{T}^*_{\mathbf{a}'}\mid \mathbf{A}=\mathbf{a},\mathbf{X}=\mathbf{x}} (\mathbf{t}) = \mathbb{1}_{\mathbf{t}}$.
\end{lemma}
\begin{proof}
In this proof, we use the notation introduced in Lemma \ref{eq:new_lemma_correction1}. Consider a set of nodes $\mathbf{T}\subseteq \mathbf{A} \cup \mathbf{X}$, and let $\mathbf{a}$ be a point in the support of $\mathbf{A}$. We denote with $\mathbf{a}_{\mathbf{T}}$ the restriction of $\mathbf{a}$ to $\mathbf{A}\cap \mathbf{T}$. We prove the claim with a induction argument. To this end, we define $\mathsf{depth}(\mathbf{T})$ as the length of the longest directed path $\pi$ in $\mathcal{G}$ from any node in $\mathbf{A}\cup \mathbf{U}_{\mathbf{T}}$ to any node in $\mathbf{T}$ (see Figure \ref{app:fig:graphs}).\footnote{Note that since $\mathsf{pa}(X) \in \mathbf{X} \cup \mathbf{A}$ for all $X\in \mathbf{T}$, then any such path $\pi$ only consists of nodes that are in $\mathbf{X} \cup \mathbf{A}\cup \mathbf{U}_{\mathbf{T}}$.}

(Base case). We first assume that  with $\mathsf{depth}(\mathbf{T}) = 0$. We further assume w.l.o.g. that $\mathbf{T}\cap \mathbf{A} = \emptyset$, since the variables $\mathbf{T}\cap \mathbf{A}$ are fixed by the intervention $\mathbf{A}\gets \mathbf{a}'$. Denote with $\mathbf{T} = g_{\mathbf{T}}(\mathsf{pa}(\mathbf{T}), \mathbf{U}_{\mathbf{T}})$ the structural equation for the joint random variable $\mathbf{T}$. Since $\mathsf{depth}(\mathbf{T}) = 0$, it holds $\mathsf{pa}(\mathbf{T}) \subseteq \mathbf{A}$.\footnote{Here, $\mathsf{pa}(\mathbf{T})$ can also be the empty set. In this case, $\mathbf{T} = \mathbf{U}_{\mathbf{T}} $ is an exogenous random variable and the proof follows.} Then, the random variable ${\mathbf{T}}^*_{\mathbf{a}'}$ is defined by the formula ${\mathbf{T}}^*_{\mathbf{a}'} = g_{\mathbf{T}}(\mathbf{a}'\mid_{\mathbf{T}}, \mathbf{U}_{\mathbf{T}})$. By Lemma \ref{eq:new_lemma_correction1}, there exist a value $\mathbf{u}'$ in the support of $\mathbf{U}_{\mathbf{T}}$ such that $\pr_{\mathbf{U}_{\mathbf{T}}\mid \mathbf{A}=\mathbf{a} ,\mathbf{X}=\mathbf{x}} (\mathbf{u})=\mathbb{1}_{\mathbf{u}'}$. It follows that $\pr_{{\mathbf{T}}^*_{\mathbf{a}'}\mid \mathbf{A} = \mathbf{a},\mathbf{X} = \mathbf{x}}(\mathbf{t}) = \mathbb{1}_{\mathbf{t}'}$ with  $\mathbf{t} \coloneqq g_{\mathbf{T}}(\mathbf{a}'\mid_{\mathbf{T}}, \mathbf{u}')$.

(Inductive step). We assume by induction that the claim holds for all ${\mathbf{T}}$ such that $\mathsf{depth}({\mathbf{T}}) \leq k$, and we prove the claim for a set ${\mathbf{T}}$ of depth $\mathsf{depth}({\mathbf{T}}) = k+1$. Since $\mathsf{depth}(\mathsf{pa}({\mathbf{T}})) < \mathsf{depth}({\mathbf{T}}) = k + 1$, by the inductive hypothesis, it holds
\begin{equation}
\label{eq:inductive}
\pr_{\mathsf{pa}({\mathbf{T}})^*_{a'}\mid \mathbf{A}=\mathbf{a} ,\mathbf{X}=\mathbf{x}} (\mathbf{x}) = \mathbb{1}_{{\mathbf{t}}'},
\end{equation}
for a point ${\mathbf{t}}'$ in the support of $\mathsf{pa}({\mathbf{T}})$. Again, consider the structural equation $\mathbf{T} = g_{\mathbf{T}}(\mathsf{pa}(\mathbf{T}), \mathbf{U}_{\mathbf{T}})$ for the joint random variable $\mathbf{T}$. In this case, the random variable ${\mathbf{T}}^*_{a'}$ is defined by the formula ${\mathbf{T}}^*_{\mathbf{a}'} = g_{\mathbf{T}}(\mathsf{pa}(\mathbf{T})^*_{a'}, \mathbf{U}_{\mathbf{T}})$. By Lemma \ref{eq:new_lemma_correction1}, there exist a value $\mathbf{u}'$ in the support of $\mathbf{U}_{\mathbf{T}}$ such that $\pr_{\mathbf{U}_{\mathbf{T}}\mid \mathbf{A}=\mathbf{a} ,\mathbf{X}=\mathbf{x}} (\mathbf{u})=\mathbb{1}_{\mathbf{u}'}$. Combining this observation with \cref{eq:inductive}, we get $\pr_{\mathbf{T}^*_{a'}\mid \mathbf{A}=\mathbf{a} ,\mathbf{X}=\mathbf{x}} (\mathbf{t}) = \mathbb{1}_{g_{\mathbf{T}}(\mathbf{t}', \mathbf{u}')}$. The claim follows by defining $\mathbf{t}\coloneqq g_{\mathbf{T}}(\mathbf{t}', \mathbf{u}')$.
\end{proof}
We use the results above to prove a third auxiliary lemma.
\begin{lemma}
\label{eq:new_lemma_correction3} 
Under the assumptions of Lemma \ref{eq:new_lemma_correction1} and Lemma \ref{eq:new_lemma_correction2}, for any intervention $\mathbf{A} \gets \mathbf{a}'$ and observational values $\mathbf{a}, \mathbf{x}$,  there exist an intervention $\mathbf{X} \gets \mathbf{x}'$ such that $\pr_{\mathbf{Y}_{\mathbf{a}', \mathbf{x}'}\mid \mathbf{A}=\mathbf{a},\mathbf{X}=\mathbf{x}} (\mathbf{y}) = \pr_{\mathbf{Y}_{\mathbf{a}'}\mid \mathbf{A}=\mathbf{a},\mathbf{X}=\mathbf{x}} (\mathbf{y})$.
\end{lemma}
\begin{proof} First note that by lemma \ref{eq:new_lemma_correction2} there exists a point $\mathbf{x}'$ such that $\pr_{\mathbf{X}^*_{\mathbf{a}'}\mid \mathbf{A}=\mathbf{a},\mathbf{X}=\mathbf{x}} (\mathbf{y}) = \mathbb{1}_{\mathbf{x}'}$. We choose this point to define the intervention $\mathbf{X} \gets \mathbf{x}'$ as in the statement of this lemma. Denote with $\mathcal{G}_{\mathbf{a}'}$ the causal graph after performing the intervention $\mathbf{A}\gets \mathbf{a}'$. Furthermore, denote with $\mathcal{G}_{\mathbf{a}',\mathbf{x}'}$ the causal graph after performing the interventions $\mathbf{A}\gets \mathbf{a}'$ and $\mathbf{X}\gets \mathbf{x}'$. We first prove that for any node $\mathbf{V}$ of $\mathcal{G}_{\mathbf{a}'}$ and $\mathcal{G}_{\mathbf{a}',\mathbf{x}'}$ it holds
\begin{equation}
\label{eq:new1_trial_asd}
\pr_{\mathbf{V}\mid \mathbf{A}=\mathbf{a} ,\mathbf{X}=\mathbf{x}} (\mathbf{v})
= \pr_{\mathbf{V}\mid \mathbf{A}=\mathbf{a} , \mathbf{X}=\mathbf{x} ,\mathbf{X}^*_{\mathbf{a}'}=\mathbf{x}'} (\mathbf{v}).
\end{equation}
In fact, it holds
\begin{align*}
\pr_{\mathbf{V}\mid \mathbf{A}=\mathbf{a} ,\mathbf{X}=\mathbf{x}} (\mathbf{v})
& = \int \pr_{\mathbf{V}\mid \mathbf{A}=\mathbf{a} , \mathbf{X}=\mathbf{x} ,\mathbf{X}^*_{\mathbf{a}'}=\mathbf{x}'} (\mathbf{v}) d \pr_{\mathbf{X}^*_{\mathbf{a}'} \mid \mathbf{A}=\mathbf{a}, \mathbf{X}=\mathbf{x}} (\mathbf{x}') & \text{(by marginalization)}\\
& = \int \pr_{\mathbf{V}\mid \mathbf{A}=\mathbf{a} ,\mathbf{X}=\mathbf{x},\mathbf{X}^*_{\mathbf{a}'}=\mathbf{x}'} (\mathbf{v}) d \mathbb{1}_{\mathbf{x}'} & \text{(by lemma \ref{eq:new_lemma_correction2} with $\mathbf{T} = \mathbf{X}$)}\\
& = \pr_{\mathbf{V}\mid \mathbf{A} = \mathbf{a},\mathbf{X}= \mathbf{x},\mathbf{X}^*_{\mathbf{a}'}=\mathbf{x}'} (\mathbf{v}). & \text{(by marginalization)} 
\end{align*}
Hence, \cref{eq:new1_trial_asd} holds. It follows that 
\begin{align*}
\pr_{\mathbf{Y}^*_{\mathbf{a}', \mathbf{x}'}\mid \mathbf{A}=\mathbf{a} ,\mathbf{X}=\mathbf{x}} (\mathbf{y})
& = \pr_{\mathbf{Y}^*_{\mathbf{a}', \mathbf{x}'}\mid \mathbf{A}=\mathbf{a} , \mathbf{X}=\mathbf{x} ,\mathbf{X}^*_{\mathbf{a}'}=\mathbf{x}'} (\mathbf{y}) & \text{(\cref{eq:new1_trial_asd} with $\mathbf{V} = \mathbf{Y}^*_{\mathbf{a}', \mathbf{x}'}$)} \\
& = \pr_{\mathbf{Y}^*_{\mathbf{a}'}\mid \mathbf{A}=\mathbf{a} ,\mathbf{X}=\mathbf{x},\mathbf{X}^*_{\mathbf{a}'}=\mathbf{x}'} (\mathbf{y}) & \text{(Axiom of Consistency \citep{pearlj})} \\
& = \pr_{\mathbf{Y}^*_{\mathbf{a}'}\mid \mathbf{A} = \mathbf{a},\mathbf{X}= \mathbf{x}} (\mathbf{y}). & \text{(\cref{eq:new1_trial_asd} with $\mathbf{V} = \mathbf{Y}^*_{\mathbf{a}'}$)}
\end{align*}
Hence, the claim holds.
\end{proof}

We now prove our main result.

\begin{proof}[Proof of \cref{lemma:conditiona_counterfactual}]
Following the notation of \cref{eq:new_lemma2}, define the set $\mathbf{X} = \mathbf{W}\setminus \mathbf{A}$. 
Note that, using this notation, the assumption $\mathbf{Y}\indep \mathbf{A}, \mathbf{W} \mid \mathbf{S}$ can be written as $\mathbf{Y} \indep \mathbf{A},\mathbf{X} \mid \mathbf{S}$. 
Suppose that it holds
\begin{equation}
\label{eq:thm_1}
\pr_{\mathbf{Y}^*_{\mathbf{a}', \mathbf{x}'}\mid \mathbf{A} = \mathbf{a},\mathbf{X} =  \mathbf{x}} (\mathbf{y}) = \int \pr_{\mathbf{Y} \mid  \mathbf{A} =  \mathbf{a}',\mathbf{X} = \mathbf{x}', \mathbf{S} = \mathbf{s}}(\mathbf{y})d\pr_{\mathbf{S}\mid  \mathbf{A} = \mathbf{a},\mathbf{X} =  \mathbf{x}} ( \mathbf{s}) 
\end{equation}
for any intervention $\mathbf{A},\mathbf{X} \gets \mathbf{a}',\mathbf{x}'$, and for any possible value $\mathbf{w}$ attained by $\mathbf{W}$. Assuming that \cref{eq:thm_1} holds, we have that
\begin{align}
\pr_{\mathbf{Y}^*_{\mathbf{a}', \mathbf{x}'}\mid \mathbf{A} =  \mathbf{a},\mathbf{X} =  \mathbf{x}} (\mathbf{y}) 
& = \int \pr_{\mathbf{Y} \mid  \mathbf{A} = \mathbf{a}',\mathbf{X}=\mathbf{x}', \mathbf{S}=\mathbf{s}}(\mathbf{y})d\pr_{\mathbf{S}\mid \mathbf{A} =\mathbf{a},\mathbf{X}=\mathbf{x}} ( \mathbf{s}) & (\mbox{assuming \cref{eq:thm_1}}) \nonumber \\
    & = \int \pr_{\mathbf{Y} \mid  \mathbf{A}=\mathbf{a},\mathbf{X}=\mathbf{x}'', \mathbf{S}=\mathbf{s}}(\mathbf{y})d \pr_{\mathbf{S}\mid  \mathbf{A}=\mathbf{a},\mathbf{X}=\mathbf{x}} ( \mathbf{s}) & (\mathbf{Y} \indep \mathbf{A},\mathbf{X} \mid \mathbf{S})\nonumber \\
    & = \pr_{\mathbf{Y}^*_{\mathbf{a}, \mathbf{x}''}\mid \mathbf{A}=\mathbf{a},\mathbf{X}=\mathbf{x}} (\mathbf{y}), & (\mbox{assuming \cref{eq:thm_1}}) \label{eq:new}
\end{align}
for any value $\mathbf{x}''$ in the support of $\mathbf{X}$. To conclude, define the set $\mathbf{T} = \mathbf{A} \setminus \mathbf{W}$. Note that $ \mathbf{W} \cup \mathbf{T} = \mathbf{A} \cup \mathbf{X}$. It follows that
\begin{align}
    \pr_{\mathbf{Y}^*_{\mathbf{a}'}\mid \mathbf{W}= \mathbf{w}} (\mathbf{y}) & = \int \pr_{\mathbf{Y}^*_{\mathbf{a}'}\mid \mathbf{A} = \mathbf{a},\mathbf{X}=\mathbf{x}} (\mathbf{y}) d \pr_{\mathbf{T}\mid \mathbf{W}=\mathbf{w}}(\mathbf{t}) & (\mbox{by conditioning})\nonumber\\ 
    & = \int \pr_{\mathbf{Y}^*_{\mathbf{a}',\mathbf{x}'}\mid \mathbf{A} = \mathbf{a},\mathbf{X}=\mathbf{x}} (\mathbf{y}) d \pr_{\mathbf{T}\mid \mathbf{W}=\mathbf{w}}(\mathbf{t}) & (\mbox{by Lemma \ref{eq:new_lemma_correction3}})\nonumber\\ 
    & = \int \pr_{\mathbf{Y}^*_{\mathbf{a}, \mathbf{x}''}\mid \mathbf{A} =  \mathbf{a},\mathbf{X} =  \mathbf{x}} (\mathbf{y})d \pr_{\mathbf{T}\mid \mathbf{W}= \mathbf{w}}( \mathbf{t} )& (\mbox{by \cref{eq:new}})\nonumber\\  
    & = \int \pr_{\mathbf{Y}^*_{\mathbf{a}}\mid \mathbf{A} = \mathbf{a},\mathbf{X}=\mathbf{x}} (\mathbf{y}) d \pr_{\mathbf{T}\mid \mathbf{W}=\mathbf{w}}(\mathbf{t}) & (\mbox{by Lemma \ref{eq:new_lemma_correction3}})\nonumber\\ 
    & = \pr_{\mathbf{Y}^*_{\mathbf{a}}\mid \mathbf{W}=\mathbf{w}} (\mathbf{y}). & (\mbox{by unconditioning})\label{eq:new_correction}
\end{align}

Note that the claim follows follows from \cref{eq:new_correction}. The proof of \cref{lemma:conditiona_counterfactual} thus boils down to proving \cref{eq:thm_1}. 
\begin{align}
    &\pr_{\mathbf{Y}^*_{\mathbf{a}', \mathbf{x}'}\mid \mathbf{A} = \mathbf{a},\mathbf{X} = \mathbf{x}} (\mathbf{y}) & \nonumber \\
    &\ = \int  \pr_{\mathbf{Y}^*_{\mathbf{a}', \mathbf{x}'}\mid \mathbf{A} = \mathbf{a},\mathbf{X} = \mathbf{x}, \mathbf{S} = \mathbf{s}} (\mathbf{y})  d \pr_{\mathbf{S}\mid  \mathbf{A}=\mathbf{a},\mathbf{X}=\mathbf{x}} ( \mathbf{s}) & (\mbox{by conditioning})\nonumber \\
    &\  = \int \pr_{\mathbf{Y}^*_{\mathbf{a}', \mathbf{x}'}\mid \mathbf{A} = \mathbf{a}',\mathbf{X} = \mathbf{x}', \mathbf{S} = \mathbf{s}} (\mathbf{y})  d \pr_{\mathbf{S}\mid  \mathbf{A}=\mathbf{a},\mathbf{X}=\mathbf{x}} ( \mathbf{s}) & (\mbox{by Theorem \ref{eq:new_lemma2}})\nonumber \\
    &\  = \int \pr_{\mathbf{Y}^* \mid \mathbf{A} = \mathbf{a}',\mathbf{X} = \mathbf{x}', \mathbf{S} = \mathbf{s}} (\mathbf{y})  d \pr_{\mathbf{S}\mid  \mathbf{A}=\mathbf{a},\mathbf{X}=\mathbf{x}} ( \mathbf{s}) & (\mbox{Axiom of Consistency \citep{pearlj}})\nonumber \\
\end{align}
and \cref{eq:thm_1} follows.
\end{proof}

\section{Proof of \cref{thm:hscic_independence}}
\label{appendix:hscic_independence}
We prove that the HSCIC can be used to promote conditional independence, using a similar technique as \citet{Park20:CME}. The following theorem holds.

\hscicind*
\begin{proof}
By definition, we can write $\hscic{\mathbf{Y},\mathbf{A} \cup \mathbf{W}}{\mathbf{S}} = H_{\mathbf{Y},\mathbf{A} \cup \mathbf{W} \mid \mathbf{S}} \circ \mathbf{S}$, where $H_{\mathbf{Y},\mathbf{A} \cup \mathbf{W}\mid \mathbf{S}}$ is a real-valued deterministic function. Hence, the HSCIC is a real-valued random variable, defined over the same domain $\Omega_{\mathbf{S}}$ of the random variable $\mathbf{X}$. 

We first prove that if $\hscic{\mathbf{Y}, \mathbf{A}\cup \mathbf{W}}{\mathbf{S}} = 0$ almost surely, then it holds $\mathbf{Y}\indep \mathbf{A}\cup \mathbf{W}\mid \mathbf{S}$. To this end, consider an event $\Omega'  \subseteq \Omega_{\mathbf{X}}$ that occurs almost surely, and such that it holds $(H_{\mathbf{Y},\mathbf{A}\cup \mathbf{W}\mid \mathbf{X}} \circ \mathbf{X})(\omega) = 0$ for all $\omega \in \Omega'$. Fix a sample $\omega \in \Omega'$, and consider the corresponding value $\mathbf{s}_\omega = \mathbf{S}(\omega)$, in the support of $\mathbf{S}$. It holds
\begin{align*}
    \int & k(\mathbf{y} \otimes [\mathbf{a},\mathbf{w}], \ \cdot \ )  d\pr_{\mathbf{Y},\mathbf{A}\cup \mathbf{W}\mid \mathbf{S} = \mathbf{s}_\omega} = \mu_{\mathbf{Y},\mathbf{A}\cup \mathbf{W}\mid \mathbf{S} = \mathbf{s}_\omega} & (\mbox{by definition}) \\
    & = \mu_{\mathbf{Y}\mid \mathbf{S} = \mathbf{s}_\omega} \otimes \mu_{\mathbf{A}\cup \mathbf{W}\mid \mathbf{S} = \mathbf{s}_\omega} & (\mbox{since $\omega \in \Omega'$})\\
    & = \int k_\mathbf{Y}(\mathbf{y}, \ \cdot \ ) d\pr_{\mathbf{Y} \mid \mathbf{S} = \mathbf{s}_\omega} \otimes \int k_{\mathbf{A}\cup \mathbf{W}}([\mathbf{a},\mathbf{w}], \ \cdot \ ) d\pr_{\mathbf{A}\cup \mathbf{W}\mid \mathbf{S} = \mathbf{s}_\omega} & (\mbox{by definition })\\
    & = \int k_\mathbf{Y}(\mathbf{y}, \ \cdot \ ) \otimes k_{\mathbf{A}\cup \mathbf{W}}([\mathbf{a},\mathbf{w}], \ \cdot \ ) d\pr_{\mathbf{Y} \mid \mathbf{S} = \mathbf{s}_\omega} \pr_{\mathbf{A}\cup \mathbf{W}\cup \mathbf{W} \mid \mathbf{S} = \mathbf{s}_\omega}, & (\mbox{by Fubini's Theorem})
\end{align*}
with $k_\mathbf{Y}$ and $k_{\mathbf{A}\cup \mathbf{W}}$ the kernels of $\mathcal{H}_\mathbf{Y}$ and $\mathcal{H}_{\mathbf{A}\cup \mathbf{W}}$ respectively. Since the kernel $k$ of the tensor product space $\mathcal{H}_\mathbf{Y}\otimes \mathcal{H}_{\mathbf{A}\cup \mathbf{W}}$ is characteristic, then the kernels $k_\mathbf{Y}$ and $k_{\mathbf{A}\cup \mathbf{W}}$ are also characteristic. Hence, it holds $\pr_{\mathbf{Y},\mathbf{A}\mid \mathbf{S} = \mathbf{s}_\omega} = \pr_{\mathbf{Y}\mid \mathbf{S} = \mathbf{s}_\omega} \pr_{\mathbf{A}\mid \mathbf{S} = \mathbf{s}_\omega}$ for all $\omega \in \Omega'$. Since the event $\Omega'$ occurs almost surely, then $\pr_{\mathbf{Y},\mathbf{A}\mid \mathbf{S} = \mathbf{s}_\omega} = \pr_{\mathbf{Y}\mid \mathbf{S} = \mathbf{s}_\omega} \pr_{\mathbf{A}\mid \mathbf{S} = \mathbf{s}_\omega}$ almost surely, that is $\mathbf{Y} \indep \mathbf{A} \cup \mathbf{W} \mid \mathbf{S}$.

Assume now that $\mathbf{Y} \indep \mathbf{A} \cup \mathbf{W} \mid \mathbf{S}$. By definition there exists an event $\Omega'' \subseteq \Omega_\mathbf{S}$ such that $\pr_{\mathbf{Y},\mathbf{A}\cup \mathbf{W}\mid \mathbf{S} = \mathbf{s}_\omega} = \pr_{\mathbf{Y}\mid \mathbf{S} = \mathbf{s}_\omega} \pr_{\mathbf{A}\cup \mathbf{W}\mid \mathbf{S} = \mathbf{s}_\omega}$ for all samples $\omega \in \Omega''$, with $\mathbf{s}_\omega = \mathbf{S}(\omega)$. It holds
\begin{align*}
    & \mu_{\mathbf{Y},\mathbf{A}\cup \mathbf{W}\mid \mathbf{S} = \mathbf{s}_\omega} \\
    &\ \ \ = \int k(\mathbf{y} \otimes [\mathbf{a},\mathbf{w}], \ \cdot \ )  d\pr_{\mathbf{Y},\mathbf{A}\cup \mathbf{W}\mid \mathbf{S} = \mathbf{s}_\omega} & (\mbox{by definition}) \\
    &\ \ \  = \int k(\mathbf{y} \otimes [\mathbf{a},\mathbf{w}], \ \cdot \ )  d\pr_{\mathbf{Y} \mid \mathbf{S} = \mathbf{s}_\omega} \pr_{\mathbf{A}\cup \mathbf{W} \mid \mathbf{S} = \mathbf{s}_\omega} & (\mbox{since $\omega \in \Omega'$})\\
    &\ \ \  = \int k_\mathbf{Y}(\mathbf{y}, \ \cdot \ ) k_{\mathbf{A}\cup \mathbf{W}}([\mathbf{a},\mathbf{w}], \ \cdot \ )  d\pr_{\mathbf{Y} \mid \mathbf{S} = \mathbf{s}_\omega} \pr_{\mathbf{A}\cup \mathbf{W}  \mid \mathbf{S} = \mathbf{s}_\omega} & (\mbox{by definition of $k$})\\
    &\ \ \  = \int k_\mathbf{Y}(\mathbf{y}, \ \cdot \ ) d\pr_{\mathbf{Y} \mid \mathbf{S} = \mathbf{s}_\omega} \otimes \int k_{\mathbf{A}\cup \mathbf{W}}([\mathbf{a},\mathbf{w}], \ \cdot \ ) d\pr_{\mathbf{A}\cup \mathbf{W}\mid \mathbf{S} = \mathbf{s}_\omega} & (\mbox{by Fubini's Theorem})\\
    &\ \ \  = \mu_{\mathbf{Y}\mid \mathbf{S} = \mathbf{s}_\omega} \otimes \mu_{\mathbf{A}\cup \mathbf{W}\mid \mathbf{S} = \mathbf{s}_\omega}. & (\mbox{by definition})
\end{align*}
The claim follows.
\end{proof}

\section{Conditional kernel mean embeddings and the HSCIC}
\label{appendix:ind_bochner}
The notion of conditional kernel mean embeddings has already been studied in the literature. We show that, under stronger assumptions, our definition is equivalent to the definition by \citet{Park20:CME}. In this section, without loss of generality we will assume that $\mathbf{W}=\emptyset$ and we will refer to the conditioning set as $\mathbf{Z}$.

\subsection{Conditional kernel mean embeddings and conditional independence} 
We show that, under stronger assumptions, the HSCIC can be defined using the Bochner conditional expected value. The Bochner conditional expected value is defined as follows. 
\begin{definition}
\label{cond:expectation}
Fix two random variables $\mathbf{Y}$, $\mathbf{Z}$ taking value in a Banach space $\mathcal{H}$, and denote with $(\Omega, \mathcal{F}, \pr)$ their joint probability space. Then, the Bochner conditional expectation of $\mathbf{Y}$ given $\mathbf{Z}$ is any $\mathcal{H}$-valued random variable $\mathbf{X}$ such that 
\begin{equation*}
\int_{E} \mathbf{Y} d \pr = \int_{E} \mathbf{X} d \pr    
\end{equation*}
for all $E \in \sigma(\mathbf{Z}) \subseteq \mathcal{F}$, with $\sigma(\mathbf{Z})$ the $\sigma$-algebra generated by $\mathbf{Z}$. We denote with $\expect{}{\mathbf{Y} \mid \mathbf{Z}}$ the Bochner expected value. Any random variable $\mathbf{X}$ as above is a version of $\expect{}{\mathbf{Y} \mid \mathbf{Z}}$.
\end{definition}

The existence and almost sure uniqueness of the conditional expectation are shown in \citet{vec_int}. Given a RKHS $\mathcal{H}$ with kernel $k$ over the support of $\mathbf{Y}$, \citet{Park20:CME} define the corresponding conditional kernel mean embedding as
\begin{equation*}
    \mu_{\mathbf{Y}\mid \mathbf{Z}} \coloneqq \expect{}{k(\cdot, \mathbf{y}) \mid \mathbf{Z}}.
\end{equation*}
Note that, according to this definition, $\mu_{\mathbf{Y}\mid \mathbf{Z}}$ is an $\mathcal{H}$-valued random variable, not a single point of $\mathcal{H}$. \citet{Park20:CME} use this notion to define the HSCIC as follows.
\begin{definition}[The HSCIC according to \citet{Park20:CME}]
\label{def:definition_hscic_krik}
Consider (sets of) random variables $\mathbf{Y}$, $\mathbf{A}$, $\mathbf{Z}$, and consider two RKHS $\mathcal{H}_{\mathbf{Y}}$, $\mathcal{H}_{\mathbf{A}}$ over the support of $\mathbf{Y}$ and $\mathbf{A}$ respectively. The HSCIC between $\mathbf{Y}$ and $\mathbf{A}$ given $\mathbf{Z}$ is defined as the real-valued random variable
\begin{equation*}
    \omega \mapsto \left \| \mu_{\mathbf{Y},\mathbf{A}\mid \mathbf{Z}}(\omega) - \mu_{\mathbf{Y}\mid \mathbf{Z}}(\omega) \otimes \mu_{\mathbf{A}\mid \mathbf{Z}}(\omega) \right \|,
\end{equation*}
for all samples $\omega$ in the domain $\Omega_\mathbf{Z}$ of $\mathbf{Z}$. Here, $\left \| \cdot \right \|$ the metric induced by the inner product of the tensor product space $\mathcal{H}_{\mathbf{Y}}\otimes \mathcal{H}_{\mathbf{Z}}$. 
\end{definition}
We show that, under more restrictive assumptions, \Cref{def:definition_hscic_krik} can be used to promote conditional independence. To this end, we use the notion of a regular version. 
\begin{definition}[Regular Version, following Definition 2.4 by \citet{prob_stoc}]
\label{def:regular_version}
Consider two random variables $\mathbf{Y}$, $\mathbf{Z}$, and consider the induced measurable spaces $(\Omega_{\mathbf{Y}}, \mathcal{F}_{\mathbf{Y}})$ and $(\Omega_{\mathbf{Z}}, \mathcal{F}_{\mathbf{Z}})$. A regular version $Q$ for $\pr_{\mathbf{Y} \mid \mathbf{Z}}$ is a mapping $Q \colon \Omega_{\mathbf{Z}} \times \mathcal{F}_{\mathbf{Y}} \rightarrow [0, + \infty] \colon (\omega, \mathbf{y}) \mapsto Q_\omega (\mathbf{y})$ such that:
(i) the map $\omega \mapsto Q_\omega (\mathbf{x})$ is $\mathcal{F}_{\mathbf{A}}$-measurable for all $\mathbf{y}$; (ii) the map $\mathbf{y} \mapsto Q_\omega (\mathbf{y})$ is a measure on $(\Omega_{\mathbf{Y}}, \mathcal{F}_{\mathbf{Y}})$ for all $\omega$; (iii) the function $Q_{\omega}(\mathbf{y})$ is a version for $\expect{}{\mathbb{1}_{\{\mathbf{Y} = \mathbf{y}\}} \mid \mathbf{Z}}$.
\end{definition}
The following theorem shows that the random variable as in \Cref{def:definition_hscic_krik} can be used to promote conditional independence. 

\begin{theorem}[Theorem 5.4 by \citet{Park20:CME}]
\label{thm:hscic_independence_1}
With the notation introduced above, suppose that the kernel $k$ of the tensor product space $\mathcal{H}_{\mathbf{X}}\otimes \mathcal{H}_{\mathbf{A}}$ is characteristic. Furthermore, suppose that $\pr_{\mathbf{Y}, \mathbf{A}\mid \mathbf{X}}$ admits a regular version. Then, $\left \| \mu_{\mathbf{Y},\mathbf{A}\mid \mathbf{Z}}(\omega) - \mu_{\mathbf{Y}\mid \mathbf{Z}}(\omega) \otimes \mu_{\mathbf{A}\mid \mathbf{Z}}(\omega) \right \| = 0$ almost surely if and only if $\mathbf{Y}\indep \mathbf{A} \mid \mathbf{Z}$. 
\end{theorem}
Note that the assumption of the existence of a regular version is essential in \cref{thm:hscic_independence_1}. In this work, HSCIC is not used for conditional independence testing but as a conditional independence measure.

\subsection{Equivalence with our approach}
The following theorem shows that under the existence of a regular version, conditional kernel mean embeddings can be defined using the Bochner conditional expected value. To this end, we use the following theorem.

\begin{theorem}[Following Proposition 2.5 by \citet{prob_stoc}]
\label{thm:regular_version}
Following the notation introduced in \Cref{def:regular_version}, suppose that $\pr_{\mathbf{Y}\mid \mathbf{Z}}( \cdot \mid \mathbf{Z})$ admits a regular version $Q_\omega (\mathbf{y})$. Consider a kernel $k$ over the support of $\mathbf{Y}$. Then, the mapping 
\begin{equation*}
    \omega \mapsto \int k(\cdot, \mathbf{y}) dQ_\omega (\mathbf{y})
\end{equation*}
is a version of $\expect{}{k(\cdot, \mathbf{y}) \mid \mathbf{Z}}$.
\end{theorem}
As a consequence of \cref{thm:regular_version}, we prove the following result.
\begin{lemma}
\label{lemma:eq_of_definitions}
Fix two random variables $\mathbf{Y}$, $\mathbf{Z}$. Suppose that $\pr_{\mathbf{Y}\mid \mathbf{Z}}$ admits a regular version. Denote with $\Omega_\mathbf{Z}$ the domain of $\mathbf{Z}$. Then, there exists a subset $\Omega \subseteq \Omega_\mathbf{Z}$ that occurs almost surely, such that $\mu_{\mathbf{Y} \mid \mathbf{Z}}(\omega) = \mu_{\mathbf{Y}\mid \mathbf{Z} = \mathbf{Z}(\omega)}$ for all $\omega \in \Omega$. Here, $\mu_{\mathbf{Y}\mid \mathbf{Z} = \mathbf{Z}(\omega)}$ is the embedding of conditional measures as in \cref{sec:preliminaries}. 
\end{lemma}
\begin{proof}
Let $Q_\omega(\mathbf{y})$ be a regular version of $\pr_{\mathbf{Y}\mid \mathbf{Z}}$. Without loss of generality we may assume that it holds $\pr_{\mathbf{Y}\mid \mathbf{Z}}(\mathbf{y}\mid \{\mathbf{Z} = \mathbf{Z}(\omega)\}) = Q_{\omega}(\mathbf{y})$. By \cref{thm:regular_version} there exists an event $\Omega \subseteq \Omega_\mathbf{Z}$ that occurs almost surely such that
\begin{equation}
\label{eq:new_cond_embeddings}
    \mu_{\mathbf{Y}\mid \mathbf{Z}}(\omega) = \mathbb{E}_{}[k(\mathbf{y}, \ \cdot \ ) \mid \mathbf{Z}](\omega) = \int k(\mathbf{y}, \ \cdot \ ) d Q_{\omega} (\mathbf{y}),
\end{equation}
for all $\omega \in \Omega$.
Then, for all $\omega \in \Omega$ it holds
\begin{align*}
    \mu_{\mathbf{Y}\mid \mathbf{Z}}(\omega) &  = \int k(\mathbf{x}, \ \cdot \ ) d Q_{\omega} (\mathbf{x}) & (\mbox{it follows from \cref{eq:new_cond_embeddings}})\\
    &  = \int k(\mathbf{x}, \ \cdot \ ) d \pr_{\mathbf{X}\mid \mathbf{A}}(\mathbf{x}\mid \{\mathbf{A} = \mathbf{A}(\omega)\}) & (\mbox{$Q_{\omega}(\mathbf{y}) = \pr_{\mathbf{Y}\mid \mathbf{Z}}(\mathbf{y}\mid \{\mathbf{Z} = \mathbf{Z}(\omega)\})$})\\
    & = \mu_{\mathbf{X}\mid \{\mathbf{A} = \mathbf{A}(\omega)\}}, & (\mbox{by definition as in \cref{sec:preliminaries}})
\end{align*}
as claimed.
\end{proof}
As a consequence of \cref{lemma:eq_of_definitions}, we can prove that the definition of the HSCIC by \citet{Park20:CME} is equivalent to ours. The following corollary holds.  
\begin{corollary}
\label{cor:definition_hscic_1}
Consider (sets of) random variables $\mathbf{Y}$, $\mathbf{A}$, $\mathbf{Z}$, and consider two RKHS $\mathcal{H}_{\mathbf{Y}}$, $\mathcal{H}_{\mathbf{A}}$ over the support of $\mathbf{Y}$ and $\mathbf{A}$ respectively. Suppose that $\pr_{\mathbf{Y}, \mathbf{A}\mid \mathbf{Z}}( \cdot \mid \mathbf{Z})$ admits a regular version. Then, there exists a set $\Omega \subseteq \Omega_\mathbf{A}$ that occurs almost surely, such that 
\begin{equation*}
\left \| \mu_{\mathbf{X},\mathbf{A}\mid \mathbf{Z}}(\omega) - \mu_{\mathbf{X}\mid \mathbf{Z}}(\omega) \otimes \mu_{\mathbf{A}\mid \mathbf{Z}}(\omega) \right \| = (H_{\mathbf{Y},\mathbf{A}\mid \mathbf{Z}} \circ \mathbf{Z})(\omega).
\end{equation*}
Here, $H_{\mathbf{Y},\mathbf{A}\mid \mathbf{Z}}$ is a real-valued deterministic function, defined as \[
H_{\mathbf{Y},\mathbf{A}\mid \mathbf{Z}}(\mathbf{z}) \coloneqq \left \| \mu_{\mathbf{Y},\mathbf{A}\mid \mathbf{Z} = \mathbf{z}} - \mu_{\mathbf{Y}\mid \mathbf{Z} = \mathbf{z}}\otimes \mu_{\mathbf{A}\mid \mathbf{Z} = \mathbf{z}} \right \|,
\]
and $\left \| \cdot \right \|$ is the metric induced by the inner product of the tensor product space $\mathcal{H}_{\mathbf{X}}\otimes \mathcal{H}_{\mathbf{A}}$. 
\end{corollary}
We remark that the assumption of the existence of a regular version is essential in \cref{cor:definition_hscic_1}.

\section{The cross-covariance operator} 
\label{appendix:ind_cross_covariance}
In this section, we show that under additional assumptions, our definition of conditional KMEs is equivalent to the definition based on the cross-covariance operator, under more restrictive assumptions. 
The definition of KMEs based on the cross-covariance operator requires the use of the following well-known result. 
\begin{lemma}
\label{lemma:isometric_isomorphism}
Fix two RKHS $\mathcal{H}_\mathbf{X}$ and $\mathcal{H}_\mathbf{Z}$, and let $\{\varphi_i\}_{i = 1}^\infty$ and $\{\psi_j\}_{j = 1}^\infty$ be orthonormal bases of $\mathcal{H}_\mathbf{X}$ and $\mathcal{H}_\mathbf{Z}$ respectively. Denote with $\mathsf{HS}(\mathcal{H}_\mathbf{X}, \mathcal{H}_\mathbf{Z})$ the set of Hilbert-Schmidt operators between $\mathcal{H}_\mathbf{X}$ and $\mathcal{H}_\mathbf{Z}$. There is an isometric isomorphism between the tensor product space $\mathcal{H}_\mathbf{X} \otimes \mathcal{H}_\mathbf{Z}$ and $\mathsf{HS}(\mathcal{H}_\mathbf{X}, \mathcal{H}_\mathbf{Z})$, given by the map
\begin{equation*}
    T\colon \sum_{i = 1}^{\infty}\sum_{j = 1}^{\infty} c_{i,j} \varphi_i \otimes \psi_j \mapsto \sum_{i = 1}^{\infty}\sum_{j = 1}^{\infty}c_{i,j} \langle \ \cdot \ , \varphi_i \rangle_{\mathcal{H}_{\mathbf{X}}} \psi_j.
\end{equation*}
\end{lemma}
For proof of this result see i.e., \citet{Park20:CME}. This lemma allows us to define the cross-covariance operator between two random variables, using the operator $T$.
\begin{definition}[Cross-Covariance Oprator]
Consider two random variables $\mathbf{X}$, $\mathbf{Z}$. Consider corresponding mean embeddings $\mu_{\mathbf{X},\mathbf{Z}}$, $\mu_{\mathbf{X}}$ and $\mu_{\mathbf{Z}}$, as defined in \cref{sec:ci}. The cross-covariance operator is defined as $\Sigma_{\mathbf{X},\mathbf{Z}} \coloneqq T (\mu_{\mathbf{X},\mathbf{Z}} - \mu_{\mathbf{X}} \otimes \mu_{\mathbf{Z}})$. Here, $T$ is the isometric isomorphism as in \cref{lemma:isometric_isomorphism}.
\end{definition}
It is well-known that the cross-covariance operator can be decomposed into the covariance of the marginals and the correlation. That is, there exists a unique bounded operator $\Lambda_{\mathbf{Y}, \mathbf{Z}} $ such that
\begin{equation*}
    \Sigma_{\mathbf{Y}, \mathbf{Z}}  = \Sigma_{\mathbf{Y}, \mathbf{Y}}^{1/2}\circ \Lambda_{\mathbf{Y}, \mathbf{Z}} \circ \Sigma_{\mathbf{Z}, \mathbf{Z}}^{1/2}
\end{equation*}
Using this notation, we define the \emph{normalized conditional cross-covariance operator}. Given three random variables $\mathbf{Y}$, $\mathbf{A}$, $\mathbf{Z}$ and corresponding kernel mean embeddings, this operator is defined as
\begin{equation}
\label{operator_cross_covarinace}
    \Lambda_{\mathbf{Y},\mathbf{A}\mid \mathbf{Z}} \coloneqq \Lambda_{\mathbf{Y},\mathbf{A}} - \Lambda_{\mathbf{Y},\mathbf{Z}} \circ \Lambda_{\mathbf{Z},\mathbf{A}}.
\end{equation}
This operator was introduced by \citet{DBLP:conf/nips/FukumizuGSS07}. The normalized conditional cross-covariance can be used to promote statistical independence, as shown in the following theorem. \begin{theorem}[Theorem 3 by \citet{DBLP:conf/nips/FukumizuGSS07}]
\label{thm:new_thm_embenddings}
Following the notation introduced above, define the random variable $\ddot{\mathbf{A}} \coloneqq (\mathbf{A}, \mathbf{Z})$. Let $\pr_\mathbf{Z}$ be the distribution of the random variable $\mathbf{Z}$, and denote with $L^2(\pr_\mathbf{Z})$ the space of the square integrable functions with probability $\pr_\mathbf{Z}$. Suppose that the tensor product kernel $k_\mathbf{Y}\otimes k_\mathbf{A} \otimes k_\mathbf{Z}$ is characteristic. Furthermore, suppose that $\mathcal{H}_\mathbf{Z} + \mathbb{R}$ is dense in $L^2(\pr_\mathbf{Z})$. Then, it holds
\begin{equation*}
    \Lambda_{\mathbf{Y},\ddot{\mathbf{A}}\mid \mathbf{Z}} = 0 \quad \mbox{if and only if} \quad \mathbf{Y} \indep \mathbf{A}\mid \mathbf{Z}.
\end{equation*}
Here, $\Lambda_{\mathbf{Y},\ddot{\mathbf{A}}\mid \mathbf{Z}}$ is an operator defined as in \cref{operator_cross_covarinace}.
\end{theorem}
By \cref{thm:new_thm_embenddings}, the operator $\Lambda_{\mathbf{Y},\ddot{\mathbf{A}}\mid \mathbf{Z}}$ can also be used to promote conditional independence. However, CIP is more straightforward since it requires less assumptions. In fact, \cref{thm:new_thm_embenddings} requires to embed the variable $\mathbf{Z}$ in an RKHS. In contrast, CIP only requires the embedding of the variables $\mathbf{Y}$ and $\mathbf{A}$.

\section{Random fourier features}
\label{appendix:random_fourier_features}
Random Fourier features is an approach to scaling up kernel methods for shift-invariant kernels \citep{DBLP:conf/nips/RahimiR07}. Recall that a shift-invariant kernel is a kernel of the form $k(\mathbf{z}, \mathbf{z}') = h_k(\mathbf{z} - \mathbf{z}')$, with $h_k$ a positive definite function. 

Fourier features are defined via the following well-known theorem.
\begin{theorem}[Bochner's Theorem]
 \label{thm:bochner_random_fourier}
For every shift-invariant kernel of the form $k(\mathbf{z}, \mathbf{z}') = h_k(\mathbf{z} - \mathbf{z}')$ with $h_k(\mathbf{0}) = 1$, there exists a probability probability density function $\pr_k(\mathbf \eta )$ such that
\begin{equation*}
    k(\mathbf{z}, \mathbf{z}') = \int e^{-2 \pi i \mathbf \eta^T(\mathbf{z}-\mathbf{z}')} d\pr_k.
\end{equation*}
\end{theorem}
Since both the kernel $k $ and the probability distribution $\pr_k$ are real-valued functions, the integrand in \cref{thm:bochner_random_fourier} ca be replaced by the function $\cos \mathbf \eta^T (\mathbf{z}- \mathbf{z}')$, and we obtain the following formula
\begin{equation}
\label{eq:fourier_features}
    k(\mathbf{z}, \mathbf{z}') = \int \cos \mathbf \eta^T (\mathbf{z}- \mathbf{z}') d\pr_k = \expect{}{\cos \mathbf \eta^T (\mathbf{z}- \mathbf{z}')},
\end{equation}
where the expected value is taken with respect to the distribution$\pr_k(\mathbf \eta )$.
This equation allows to approximate the kernel $k(\mathbf{z}, \mathbf{z}')$, via the empirical mean of points $\mathbf \eta_1, \dots , \mathbf \eta_l$ sampled independently according to $\pr_k$. In fact, it is possible to prove exponentially fast convergence of an empirical estimate for $\expect{}{\cos \mathbf \eta^T (\mathbf{z}- \mathbf{z}')}$, as shown in the following theorem.
\begin{theorem}[Uniform Convergence of Fourier Features, Claim 1 by \citet{DBLP:conf/nips/RahimiR07}]
\label{thm:fourier_features}
Following the notation introduced above, fix any compact subset $\Omega$ in the domain of $k$, and consider points $\mathbf \eta_1, \dots, \mathbf \eta_l$ sampled independent according to the distribution $\pr_k$. Define the function
\begin{equation*}
    \hat{k}(\mathbf{z},\mathbf{z}') \coloneqq \frac{1}{l} \sum_{j = 1}^l \cos \mathbf \eta_j^T (\mathbf{z}-\mathbf{z}'),
\end{equation*}
for all $(\mathbf{z}, \mathbf{z}') \in \Omega$. Then, it holds
\begin{equation*}
    \pr \left ( \sup_{\mathbf{z}, \mathbf{z}' } \left | \hat{k}(\mathbf{z}, \mathbf{z}') - k(\mathbf{z},\mathbf{z}') \right | \geq \varepsilon \right ) \leq 2^8 \sigma_k  \frac{\mathsf{diam}(\Omega)}{\varepsilon} \exp \left \{ - \frac{\varepsilon^2l}{4(d + 1)}\right \}.
\end{equation*}
Here $\sigma^2_k$ is the second moment of the Fourier transform of the kernel $k$, and $d$ is the dimension of the arrays $\mathbf{z}$ and $\mathbf{z}'$.
\end{theorem}
By \cref{thm:fourier_features}, the estimated kernel $\hat{k}$ is a good approximation of the true kernel $k$ on the set $\Omega$.

Similarly, we can approximate the Kernel matrix using Random Fourier features. Following the notation introduced above, define the function
\begin{equation}
\label{eq:fourier_function}
    \zeta_{k,l}(\mathbf{z}) \coloneqq \frac{1}{\sqrt{l}}\left [\cos \mathbf \eta_1^T \mathbf{z}, \dots , \cos \mathbf \eta_l^T \mathbf{z} \right ]
\end{equation}
with $\mathbf \eta_1, \dots , \mathbf \eta_l$ sampled independent according to the distribution $\pr_k$.

We can approximate the Kernel matrix using the functions defined as in \cref{eq:fourier_function}. Consider $n$ samples $\mathbf{z}_1, \dots, \mathbf{z}_n$, and denote with $Z$ the $n\times l$ matrix whose $i$-th row is given by $\zeta_{k,l}(\mathbf{z}_i)$. Similarly, denote with $Z^*$ the $l\times n$ matrix whose $i$-th column is given by $\zeta_{k,l}^*(\mathbf{z}_i)$. Then, we can approximate the kernel matrix as $\hat{K}_{\mathbf{Z}} \approx Z Z^*$. 

We can also use this approximation to compute the kernel ridge regression parameters as in \cref{sec:ci} using the formula $\hat{w}_{\mathbf{Y}\mid \mathbf{Z}}(\cdot) \approx ( Z Z^* - n \lambda I )^{-1} \left [
\begin{array}{c}
k_\mathbf{Z}(\cdot,\mathbf{z}_1), \cdots , 
k_\mathbf{Z}(\cdot,\mathbf{z}_n)
\end{array}
\right ]^T$. \citet{DBLP:conf/icml/AvronKMMVZ17} argue that the approximate kernel ridge regression, as defined above, is an accurate estimate of the true distribution. Their argument is based on proving that the matrix $Z Z^* - n \lambda I$ is a \emph{good approximation} of $\hat{K}_{\mathbf{Z}} - n \lambda I$. The notion of good approximation is clarified by the following definition.
\begin{definition}
Fix two Hermitian matrices $A$ and $B$ of the same size.
We say that a matrix $A$ is a $\gamma$-spectral approximation of another matrix $B$, if it holds $(1 - \gamma)B \preceq A \preceq (1 +\gamma)B$. Here, the $\preceq$ symbol means that $A - (1 - \gamma)B$ is positive semi-definite, and that $(1 + \gamma)B - A$ is positive semi-definite.
\end{definition}
\citet{DBLP:conf/icml/AvronKMMVZ17} prove that $Z Z^* - n \lambda I$ is a $\gamma$- approximation of $\hat{K}_{\mathbf{Z}} - n \varepsilon I$, if the number of samples $\mathbf \eta_1, \dots , \mathbf \eta_l$ is sufficiently large.

\begin{theorem}[Theorem 7 by \citet{DBLP:conf/icml/AvronKMMVZ17}]
Fix a constant $\gamma \leq 1/2$. Consider $n$ samples $\mathbf{z}_1, \dots, \mathbf{z}_n$, and denote with $\hat{K}_{\mathbf{Z}}$ the corresponding kernel matrix. Suppose that it holds $\|\hat{K}_{\mathbf{Z}}\|_2 \geq n \lambda$ for a constant $\lambda > 0 $. Fix $\mathbf \eta_1, \dots , \mathbf \eta_l$ samples with  
\begin{equation*}
l \geq  \frac{8}{3\gamma^{2}\lambda} \ln \frac{16\  \mathsf{tr}_\lambda(\hat{K}_{\mathbf{Z}})}{\gamma}   
\end{equation*}
Then, the matrix $Z Z^* - n \lambda I$ is a $\gamma$- approximation of $\hat{K}_{\mathbf{Z}} - n \lambda I$ with probability at least $1 - \gamma$, for all $\gamma \in (0, 1)$. Here, $\mathsf{tr}_\lambda(\hat{K}_{\mathbf{Z}})$ is defined as the trace of the matrix $ \hat{K}_{\mathbf{Z}}(\hat{K}_{\mathbf{Z}} + n\lambda I)^{-1}$.
\end{theorem}

We conclude this section by illustrating the use of random Fourier features to approximate a simple Gaussian kernel. Suppose that we are given a kernel of the form
\begin{equation*}
    k(\mathbf{z}, \mathbf{z}') \coloneqq \exp \left \{ -\frac{1}{2}\sigma \| \mathbf{z}- \mathbf{z}'\|_2^2\right \}.
\end{equation*}
Then, $k(\mathbf{z}, \mathbf{z}')$ can be estimated as in \cref{thm:fourier_features}, with $\mathbf \eta_1, \dots , \mathbf \eta_l \sim \mathcal{N}(0, \Sigma)$, with $\Sigma \coloneqq \sigma^{-1} I$, with $I$ the identity matrix. The functions $\zeta_{k,l}(\mathbf{z})$ can be defined accordingly. 
\section{Additional experiments and settings}\label{app:synthetic}

This section contains detailed information on the experiments and additional results.

\subsection{Dataset for model performance with the use of the HSCIC}\label{app:synthetic-experiment}

The data-generating mechanism corresponding to the results in \cref{fig:simulated_exp} is the following:
\begin{spreadlines}{1.5em}
\begin{align*}
    & \mathbf{Z}  \sim \mathcal{N}\left (0, 1 \right ) \qquad  \qquad
    \mathbf{A}  = \mathbf{Z}^2+ \varepsilon_{\mathbf{A}}\\
    & \mathbf{L} = \exp \left \{-\frac{1}{2}\mathbf{A}^2 \right \} \sin \left ( 2 \mathbf{A} \right ) + 2\mathbf{Z} \frac{1}{5} \varepsilon_{\mathbf{L}}\\ 
    & \mathbf{Y} =  \frac{1}{2} \exp \left \{-\mathbf{L} \mathbf{Z} \right \} \cdot \sin \left ( 2 \mathbf{L}\mathbf{Z} \right ) + 5\mathbf{A}  + \frac{1}{5} \varepsilon_{\mathbf{Y}},
\end{align*}
\end{spreadlines}
where $\varepsilon_{\mathbf{A}} \overset{}{\sim} \mathcal{N}\left (0, 1 \right )$ and $ \varepsilon_{\mathbf{Y}}, \varepsilon_{\mathbf{L}} \overset{i.i.d.}{\sim} \mathcal{N}\left (0, 0.1 \right ) $.\\
In the first experiment, \cref{fig:simulated_exp} shows the results of feed-forward neural networks consisting of $8$ hidden layers with $20$ nodes each, connected with a rectified linear activation function (ReLU) and a linear final layer. Mini-batch size of $256$ and the Adam optimizer with a learning rate of $10^{-3}$ for $1000$ epochs were used.

\subsection{Datasets and results for comparison with baselines}\label{appendix:baselines}
The comparison of our method CIP with the CF1 and CF2 is done on different simulated datasets. These will be referred to as Scenario 1 and Scenario 2. The data generating mechanism corresponding to the results in \cref{fig:simulated_exp} (right) is the following:
\begin{spreadlines}{1.5em}
\begin{align*}
    & \mathbf{Z}  \sim \mathcal{N}\left (0, 1 \right ) \qquad  \qquad 
    \mathbf{A}  = \exp \left\{ \frac{1}{2}\mathbf{Z}^2\right\}\cdot \sin \left ( 2 \mathbf{Z} \right )+ \varepsilon_{\mathbf{A}}\\
    & \mathbf{L}  = (\mathbf{A}+0.1\mathbf{Z})\cdot \varepsilon_{\mathbf{L}}\\
    & \mathbf{Y}  = \mathbf{A}+\mathbf{L}+0.1\cdot \sin \left ( \mathbf{Z} \right )
\end{align*}
\end{spreadlines}
where $\varepsilon_{\mathbf{A}}, \varepsilon_{\mathbf{L}} \overset{i.i.d.}{\sim} \mathcal{N}\left (0, 1 \right )$ and $ \varepsilon_{\mathbf{Y}}  \overset{i.i.d.}{\sim} \mathcal{N}\left (0, 0.1 \right ) $. This is referred to as Scenario 1. The data generating mechanism for Scenario 2 is the following:
\begin{spreadlines}{1.5em}
\begin{align*}
    & \mathbf{Z}  \sim \mathcal{N}\left (0, 1 \right ) \qquad  \qquad 
    \mathbf{A}  =\exp \left\{ \frac{1}{2}\mathbf{Z}^2\right\}\cdot \sin \left ( 2 \mathbf{Z} \right )+ \varepsilon_{\mathbf{A}}\\
    & \mathbf{L}  = \exp \left\{-\frac{1}{2}\mathbf{A}^2\right\} \cdot\varepsilon_{\mathbf{L}} +2\mathbf{Z}\\  
    & \mathbf{Y}  =  \frac{1}{2}\sin \left ( \mathbf{Z} \mathbf{L} \right ) \cdot\exp \left\{-\mathbf{Z}\mathbf{L}\right\}  + \frac{1}{5} \varepsilon_{\mathbf{Y}},
\end{align*}
\end{spreadlines}
where $\varepsilon_{\mathbf{A}}, \varepsilon_{\mathbf{L}} \overset{i.i.d.}{\sim} \mathcal{N}\left (0, 1 \right )$ and $ \varepsilon_{\mathbf{Y}}  \overset{i.i.d.}{\sim} \mathcal{N}\left (0, 0.1 \right )$. 
\cref{fig:simulated_exp} (right) and \cref{table:baseline_comparison} present the average and standard deviation resulting from $9$ random seeds runs. For CIP, the same hyperparameters as in the previous setting are used. The MLPs implemented in CF1 and CF2 used for the prediction of $\mathbf{\hat{Y}}$ and the one used for the prediction of the $\mathbf{L}$ residuals in CF2 are all designed with similar architecture and training method. The MLP models consist of $8$ hidden layers with $20$ nodes each, connected with a rectified linear activation function (ReLU) and a linear final layer. During training, mini-batch size of $64$ and the Adam optimizer with a learning rate of $10^{-3}$ for $200$ epochs were used. 
\begin{table}[t]
  \centering
  \caption{\textbf{Performance of the \hsciconly{} against baselines} CF1 and CF2 on two synthetic datasets.  Notably, in both scenarios it is possible to select $\gamma$ values for which CIP outperforms CF2 in \textbf{MSE} and \textbf{VCF} simultaneously. }
  \label{table:baseline_comparison}
  {\small 
  \begin{tabularx}{1.0\textwidth}{lrXXXXXX}
     \toprule
     & \multicolumn{3}{c}{\textbf{Scenario 1}} 
     & \multicolumn{3}{c}{\textbf{Scenario 2}} \\
     \cmidrule(lr){2-4} \cmidrule(l){5-7} 
     \rowcolor{white}& \textbf{MSE} {\color{gray}$\times 10^6$} & \textbf{HSCIC} {\color{gray}$\times 10^3$}&$\textbf{VCF}${\color{gray}$\times 10^3$} & \textbf{MSE}{\color{gray}$\times 10^3$}& \textbf{HSCIC} {\color{gray}$\times 10^2$} & $\textbf{VCF}${\color{gray}$\times 10^2$} \\
   \midrule
   \rowcolor{gray!10!white} $\gamma=0.001$ & $12 \pm 9$ & $45.38 \pm 0.41$ & $54.93 \pm 7.50$ & $0.0006 $ & $35.64 \pm 0.32$ & $5.60 \pm 0.03$ \\
   $\gamma=0.01$ & $16 \pm 12$ & $45.35 \pm 0.41$ & $54.57 \pm 7.18$ & $0.0019 $ & $35.44 \pm 0.33$ & $5.50 \pm 0.03$ \\
   \rowcolor{gray!10!white} $\gamma=0.1$ & $32 \pm 20$ & $45.11 \pm 0.43$ & $54.16 \pm 7.58$ & $0.11 \pm 0.006$ & $33.47 \pm 0.36$ & $4.46 \pm 0.04$ \\
   $\gamma=0.2$ & $81 \pm 14$ & $44.78 \pm 0.47$ & $53.59 \pm 7.90$ & $0.42 \pm 0.02$ & $31.38 \pm 0.38$ & $3.52 \pm 0.04$ \\
   \rowcolor{gray!10!white} $\gamma=0.3$ & $192 \pm 33$ & $43.92 \pm 0.52$ & $52.92 \pm 7.54$ & $0.82 \pm 0.04$ & $29.75 \pm 0.34$ & $2.50 \pm 0.04$ \\
   $\gamma=0.4$ & $384 \pm 58$ & $43.88 \pm 0.57$ & $52.06 \pm 7.25$ & $1.21 \pm 0.05$ & $28.63 \pm 0.33$ & $1.79 \pm 0.03$ \\
   \rowcolor{gray!10!white} $\gamma=0.5$ & $685 \pm 133$ & $43.26 \pm 0.65$ & $51.64 \pm 7.40$ & $1.56 \pm 0.08$ & $27.81 \pm 0.26$ & $1.1 \pm 0.01$ \\
   $\gamma=0.6$ & $1117 \pm 165$ & $42.47 \pm 0.73$ & $50.96 \pm 7.36$ & $1.84 \pm 0.11$ & $26.87 \pm 0.22$ & $0.79 \pm 0.01$ \\
   \rowcolor{gray!10!white} $\gamma=0.7$ & $1655 \pm 223$ & $42.11 \pm 0.80$ & $50.31 \pm 7.44$ & $2.11 \pm 0.14$ & $26.08 \pm 0.20$ & $0.49 \pm 0.01$ \\
   $\gamma=0.8$ & $2225 \pm 296$ & $41.87 \pm 0.84$ & $49.76 \pm 7.25$ & $2.37 \pm 0.15$ & $25.27 \pm 0.18$ & $0.31 \pm 0.01$ \\
   \rowcolor{gray!10!white} $\gamma=0.9$ & $2832 \pm 372$ & $41.52 \pm 0.92$ & $49.17 \pm 7.41$ & $2.58 \pm 0.17$ & $24.64 \pm 0.16$ & $0.21 \pm 0.01$ \\
   $\gamma=1.0$ & $3472 \pm 422$ & $38.37 \pm 0.97$ & $48.71 \pm 7.55$ & $2.77 \pm 0.19$ & $24.21 \pm 0.15$ & $0.14 \pm 0.01$ \\
      \midrule
   \rowcolor{gray!10!white}  CF1 &         $10321 \pm 72$ & $41.37 \pm 0.58$ & $0 \pm 0.00$ & $4.59 \pm 0.4478$ & $25.01 \pm 0.25$ & $0 \pm 0.00$ \\
    CF2 &         $2728 \pm 272$ & $41.37 \pm 0.92$ & $59.50 \pm 10.35$ & $3.97 \pm 0.3479$ & $27.03 \pm 0.35$ & $2.62 \pm 0.81$ \\

   \bottomrule
  \end{tabularx}
  }
\end{table}

\subsection{Datasets and results for multi-dimensional variables experiments}
\label{appendix:multid_setting}
\begin{figure*}[t]
  \centering
  \includegraphics[height=3.5cm]{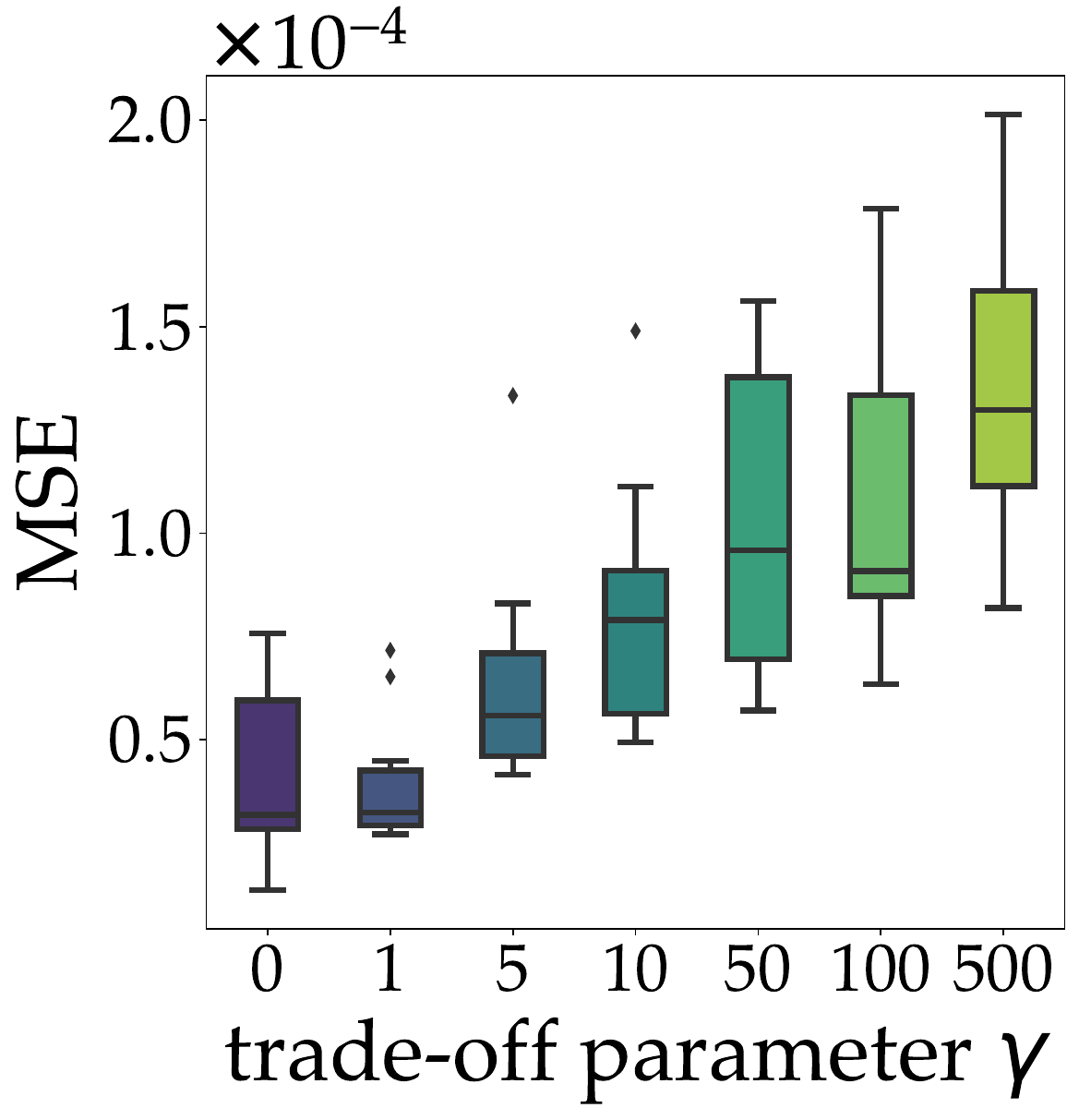}\hfill
    \includegraphics[height=3.5cm]{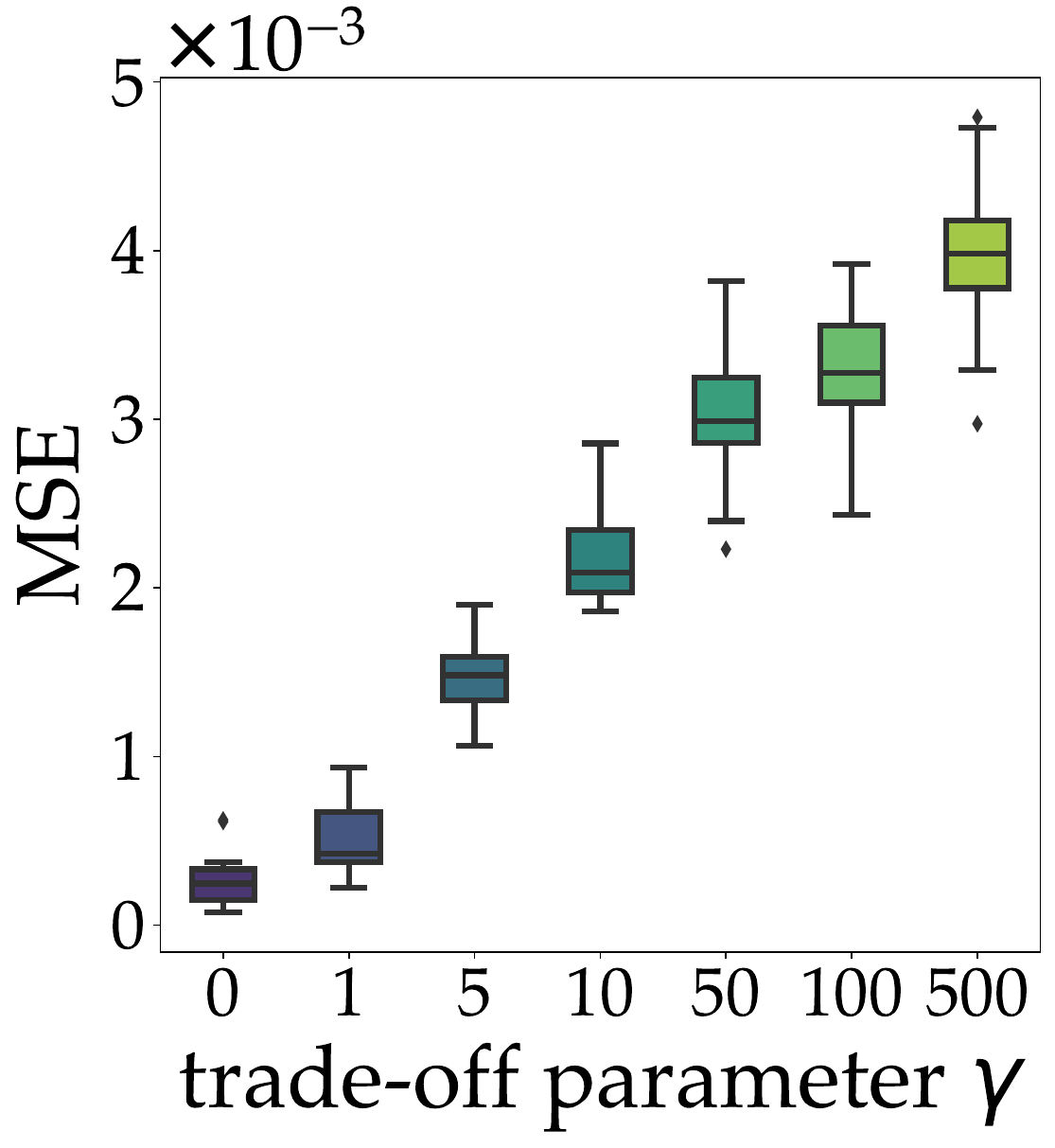}\hfill
  \includegraphics[height=3.5cm]{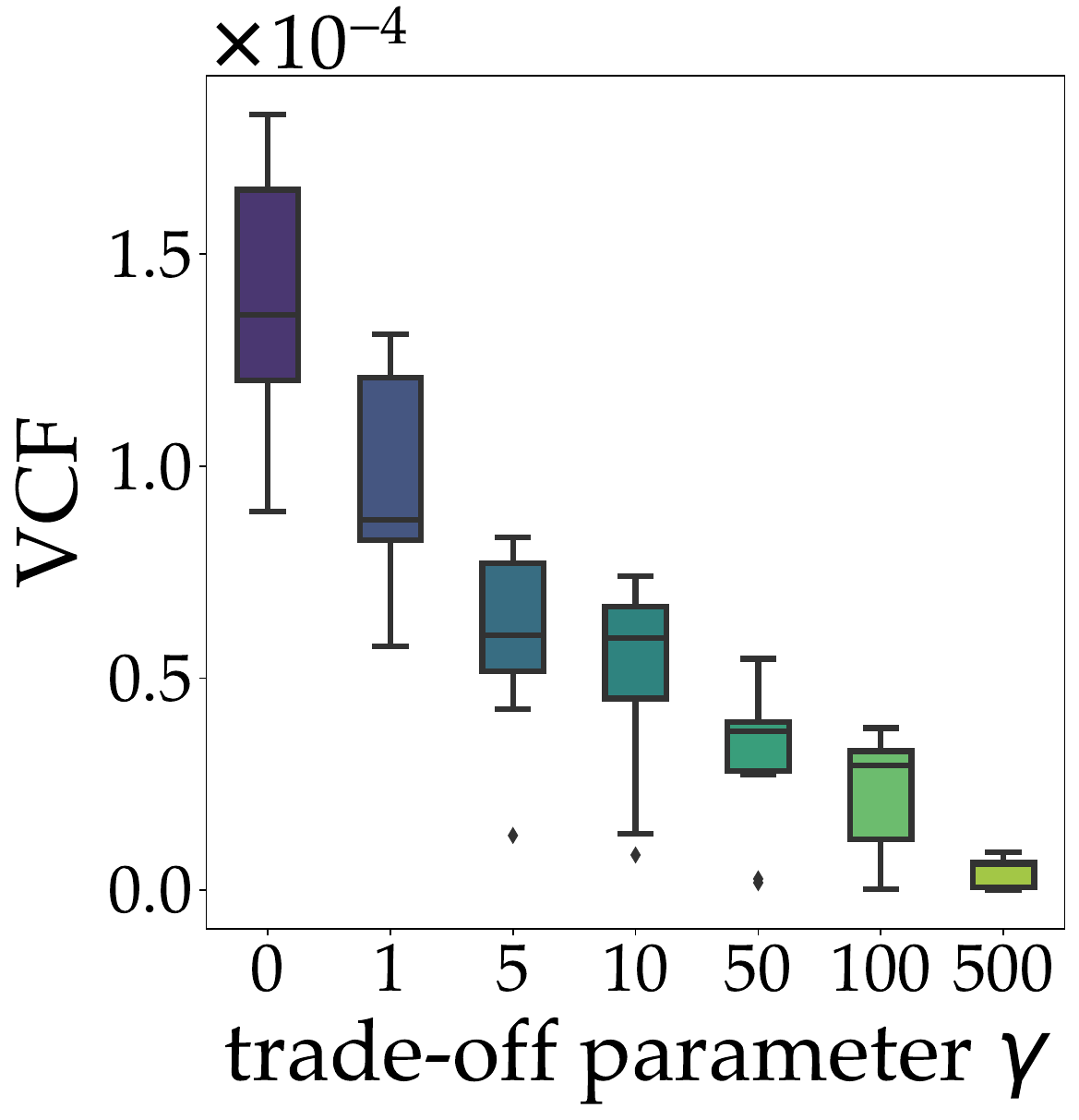}\hfill
  \includegraphics[height=3.5cm]{boxplot_MSE_multidimensional_100_squared.pdf}\hfill
  \vspace{-2mm}
  \caption{MSE , \hsciconly{}, \VCF{} for increasing dimension of $\mathbf{A}$ on synthetic data from \cref{appendix:multid_setting} with $\text{dimA}=20$ (left) and $\text{dimA}=100$ (right). All other variables are one-dimensional. 
  \label{appendix:fig:multidimensional_results_A}}
\end{figure*}

The data-generating mechanisms for the multi-dimensional settings of \cref{fig:multidimensional_results_A} are now shown. Given $\text{dimA}=D_1 \geq 2$, the datasets were generated from: 
\begin{spreadlines}{1.5em}
\begin{align*}
    \mathbf{Z}  & \sim \mathcal{N}\left (0, 1 \right ) \qquad \qquad
    \mathbf{A}_i  = \mathbf{Z}^2+ \varepsilon_{\mathbf{A}}^i \hspace{0.5cm} \text{for } i \in \{1,{D_1}\}\\
    \mathbf{L}  & =  \exp \left \{-\frac{1}{2}\mathbf{A}_1 \right \} + \sum_{i=1}^{D_1}\mathbf{A}_i\cdot \sin( \mathbf{Z})+ 0.1\cdot \varepsilon_{\mathbf{L}} \\
    \mathbf{Y}  & =  \exp \left \{-\frac{1}{2}\mathbf{A}_2 \right \} \cdot  \sum_{i=1}^{D_1}\mathbf{A}_i +  \mathbf{L}  \mathbf{Z}+ 0.1\cdot \varepsilon_{\mathbf{Y}},
\end{align*}
\end{spreadlines}
where $\varepsilon_{\mathbf{L}}, \varepsilon_{\mathbf{Y}} \overset{i.i.d}{\sim} \mathcal{N}\left (0, 0.1 \right )$ and $ \varepsilon_{\mathbf{A}}^1, ..., \varepsilon_{\mathbf{A}}^{D_1}  \overset{i.i.d}{\sim} \mathcal{N}\left (0, 1 \right )$.
In this experiment, the mini-batch size chosen is $512$ and the same hyperparameters are used as in the previous settings. The neural network architecture is trained for $800$ epochs.  \cref{appendix:fig:multidimensional_results_A} present the results corresponding to $10$ random seeds with different values of the trade-off parameter $\gamma$ corresponding to different values of $\text{dimA}$ among $\{15,100\}$. In all of the box plots, it is evident that there exists a trade-off between the accuracy and counterfactual invariance of the predictor. As the value of $\gamma$ increases, there is a consistent trend of augmenting counterfactual invariance (as evidenced by the decrease in the \VCF{} metric). Similarly to the previous boxplots visualizations, the boxes represent the interquartile range (IQR), the horizontal line is the median, and whiskers show the minimum and maximum values, excluding the outliers (determined as a function of the inter-quartile range). Outliers are represented in the plot as dots. 

\subsection{Image dataset}\label{app:sec:imageresults}

The simulation procedure for the results shown in \cref{sec:imagedata} is the following.
\begin{align*}
    \texttt{shape} &\sim \mathbb{P}(\texttt{shape})\\
    \texttt{y-pos}  &\sim \mathbb{P}(\texttt{y-pos})\\
    \texttt{color} &\sim \mathbb{P}(\texttt{color})\\
    \texttt{orientation} &\sim \mathbb{P}(\texttt{orientation})\\
    \texttt{x-pos}  &=  \text{round}(x), \hspace{0.2cm}   \text{where } x \sim \mathcal{N}(\texttt{shape}+\texttt{y-pos},1)\\
    \texttt{scale} &= \text{round}\Big( \left(\frac{\texttt{x-pos}}{24} + \frac{\texttt{y-pos}}{24}\right)\cdot \texttt{shape} +\epsilon_S \Big)\\
    \textbf{Y} &= e^{\texttt{shape}}\cdot\texttt{x-pos}+\texttt{scale}^2\cdot\sin(\texttt{y-pos}) +\epsilon_Y,
\end{align*}
where $\epsilon_S \sim \mathcal{N}(0,1)$ and $\epsilon_Y \sim \mathcal{N}(0, 0.01)$. The data has been generated via a matching procedure on the original dSprites dataset.

In \cref{table:table_cnn_arch}, the hyperparameters of the layers of the convolutional neural network are presented. Each of the convolutional groups also has a ReLU activation function and a dropout layer. Two MLP architectures have been used. The former takes as input the observed tabular features. It is composed by two hidden layers of $16$ and $8$ nodes respectively, connected with ReLU activation functions and dropout layers. The latter takes as input the concatenated outcomes of the CNN and the other MLP. It consists of three hidden layers of $8$, $8$ and $16$ nodes, respectively.  
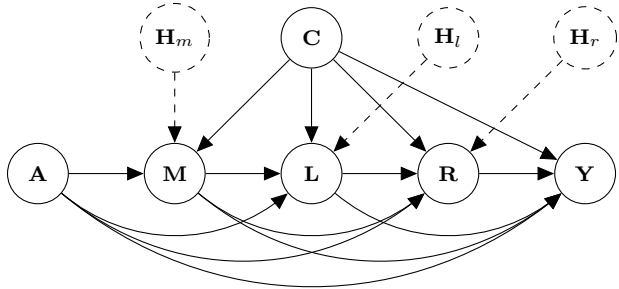
\begin{figure*}[b]
\centering
  \begin{tikzpicture}[node distance=18mm and 8mm, main/.style = {draw, circle, minimum size=0.8cm}, >={triangle 45}]  
    \node[main] (1) {$\scriptstyle\mathbf{A}$}; 
    \node[main] (2) [right of=1] {$\scriptstyle\mathbf{M}$}; 
    \node[main] (3) [right of=2] {$\scriptstyle\mathbf{D}$}; 
    \node[main] (4) [right of=3] {$\scriptstyle\mathbf{R}$}; 
    \node[main] (5) [right of=4] {$\scriptstyle\mathbf{Y}$}; 
    \node[main, dashed] (6) [above of=2] {$\scriptstyle\mathbf{H}_m$}; 
    \node[main, dashed] (7) [above of=4] {$\scriptstyle\mathbf{H}_d$}; 
    \node[main, dashed] (8) [above of=5] {$\scriptstyle\mathbf{H}_r$}; 
    \node[main] (9) [above of=3] {$\scriptstyle\mathbf{C}$}; 
    
    \draw[->] (1) -- (2); 
    \draw[->] (2) -- (3); 
    \draw[->] (3) -- (4); 
    \draw[->] (4) -- (5); 
    \draw[->, dashed] (6) -- (2); 
    \draw[->, dashed] (8) -- (4); 
    \draw[->, dashed] (7) -- (3); 
    \draw[->] (9) -- (2); 
    \draw[->] (9) -- (3); 
    \draw[->] (9) -- (4); 
    \draw[->] (9) -- (5); 
    \draw[->] (1) to [out=320, in=220, looseness=1] (3); 
    \draw[->] (1) to [out=320, in=220, looseness=1] (4); 
    \draw[->] (2) to [out=320, in=220, looseness=1] (4); 
    \draw[->] (2) to [out=320, in=220, looseness=1] (5); 
    \draw[->] (3) to [out=320, in=220, looseness=1] (5); 
    \draw[->] (1) to [out=320, in=220, looseness=1] (5); 
  \end{tikzpicture}%
\caption{Assumed causal graph for the Adult dataset, as in \citet{chiappa2021fairness}. The variables $\mathbf{H}_m$, $\mathbf{H}_d$,  $\mathbf{H}_r$ are unobserved, and jointly trained with the predictor $\hat{\mathbf{Y}}$.
  \label{fig:causal_graph_uci}}
\end{figure*}

\begin{table}[t]
  \centering
  \caption{Architecture of the convolutional neural network used for the image dataset, as described in \cref{app:sec:imageresults}.}
  \label{table:table_cnn_arch}
  \small
  \begin{tabularx}{0.8\columnwidth}{Ycccc}
  \toprule
  \rowcolor{white} \textbf{layer}  & \textbf{\# filters} & \textbf{kernel size} & \textbf{stride size} & \textbf{padding size} \\
   \midrule
   \rowcolor{gray!10!white} convolution & $16$ & $5$ & $2$ & $2$\\
  \rowcolor{white}  max pooling & $1$ & $3$ & $2$ & $0$\\
   \rowcolor{gray!10!white}  convolution & $64$ & $5$ & $1$ & $2$\\
   \rowcolor{white} max pooling & $1$ & $1$ & $2$ & $0$\\
   \rowcolor{gray!10!white} convolution & $64$ & $5$ & $1$ & $2$\\
   \rowcolor{white} max pooling & $1$ & $2$ & $1$ & $0$\\
   \rowcolor{gray!10!white}  convolution & $16$ & $5$ & $1$ & $3$\\
   \rowcolor{white} max pooling & $1$ & $2$ & $2$ & $0$\\
   \bottomrule
   \end{tabularx}%
\end{table}

\subsection{Fairness with continuous protected attributes}\label{app:sec:adult}

The pre-processing of the UCI Adult dataset was based upon the work of \citet{chiappa2021fairness}. Referring to the causal graph in \cref{fig:causal_graph_uci}, a variational autoencoder \citep{KingmaW13} was trained for each of the unobserved variables $\mathbf{H_m}$, $\mathbf{H_l}$ and $\mathbf{H_r}$. The prior distribution of these latent variables is assumed to be standard Gaussian. The posterior distributions $\mathbb{P}(\mathbf{H_m}|V)$, $\mathbb{P}(\mathbf{H_r}|V)$, $\mathbb{P}(\mathbf{H_d}|V)$ are modeled as $10$-dimensional Gaussian distributions, whose means and variances are the outputs of the encoder.

The encoder architecture consists of a hidden layer of $20$ hidden nodes with hyperbolic tangent activation functions, followed by a linear layer. The decoders have two linear layers with a hyperbolic tangent activation function. The training loss of the variational autoencoder consists of a reconstruction term (Mean-Squared Error for continuous variables and Cross-Entropy Loss for binary ones) and the Kullback–Leibler divergence between the posterior and the prior distribution of the latent variables. For training, we used the Adam optimizer with learning rate of $10^{-2}$, $100$ epochs, mini-batch size $128$.

The predictor $\mathbf{\hat{Y}}$ is the output of a feed-forward neural network consisting of a hidden layer with a hyperbolic tangent activation function and a linear final layer. In the training we used the Adam optimizer with learning rate $10^{-3}$, mini-batch size $128$, and trained for $100$ epochs. The choice of the network architecture is based on the work of \citet{chiappa2021fairness}.

The estimation of counterfactual outcomes is based on a Monte Carlo approach. Given a data point,
$500$ values of the unobserved variables are sampled from the estimated posterior distribution. Given an interventional value for $A$, a counterfactual outcome is estimated for each of the sampled unobserved values. The final counterfactual outcome is estimated as the average of these counterfactual predictions. In this experimental setting, we have $k=100$ and $d=1000$.

In the causal graph presented in \cref{fig:causal_graph_uci}, $\mathbf{A}$ includes the variables age and gender, $\mathbf{C}$ includes nationality and race, $\mathbf{M}$ marital status, $\mathbf{D}$ level of education, $\mathbf{R}$ the set of the working class, occupation, and hours per week and $\mathbf{Y}$ the income class. Compared to \citet{chiappa2021fairness}, we include the race variable in the dataset as part of the baseline features $\mathbf{C}$.
The loss function is the same as \cref{eq:total_loss} but Binary Cross-Entropy loss ($\mathcal{L}_{\mathbf{BCE}}$) is used instead of Mean-Squared Error loss:
\begin{equation}\label{eq:total_loss_uci}
    \mathcal{L}_{\textsc{CIP}}(\hat{\mathbf{Y}}) = \mathcal{L}_{\mathbf{BCE}}(\hat{\mathbf{Y}}) + \gamma \cdot\hsciconly{}\left(\hat{\mathbf{Y}},\left\{\text{age},\text{gender}, \text{marital status}, \text{education}, \text{work} \right\}\middle| \mathbf{S} \right)\:,
\end{equation}
where the set $\mathbf{S}=\{\text{Race},\text{Nationality}\}$ blocks all the non-causal paths from $\mathbf{W} \cup \mathbf{A}$ to $\mathbf{Y}$. In this example we have $\mathbf{W}=\{\mathbf{C} \cup \mathbf{M} \cup \mathbf{D} \cup \mathbf{R}\}$. The results in \cref{fig:adult_experiment} (right) refer to one run with conditioning set $\mathbf{S}=\{\text{Race},\text{Nationality}\}$. The results correspond to 4 random seeds. 
\subsection{Illustrating the choice of $\gamma$}\label{app:choosinggamma}
In \cref{sec:learningCIP}, we propose to choose $\gamma$ to obtain a maximal level of CI within a given tolerance on predictive performance.
Here, we illustrate results from running the proposed procedure that dynamically selects $\gamma$, adjusted to different predefined accuracy thresholds in a classification setting.
Specifically, the algorithm chooses the largest $\gamma$ value that yields an accuracy equal to or better than the threshold. 
As described the algorithm operates on $\gamma$ values on a logarithmic scale, thereby ensuring a fine-grained search over a wide range of potential trade-off points.
\Cref{table:choice_gamma} shows the found trade-offs for tolerated accuracies of 90\%, 70\%, and 1\% in the same setting as \cref{app:synthetic-experiment}. 

\begin{table}[t]
  \centering
  \caption{Results of MSE and \VCF{} (all times $10^2$ for readability) on synthetic data of CIP with trade-off parameters depending on the chosen accuracy threshold.}
  \label{table:choice_gamma}
  \small
  \begin{tabularx}{0.5\columnwidth}{Xrr}
  \toprule
  \rowcolor{white}& \textbf{VCF} {\color{gray}$\times 10^2$} & \textbf{HSCIC} {\color{gray}$\times 10^2$} \\ 
   \midrule
   \rowcolor{gray!10!white} $90\%$ accuracy & $3.14 \pm 0.92 $& $4.51 \pm 0.72 $\\
    \rowcolor{white} $70\%$ accuracy & $3.01 \pm 0.80 $& $4.44 \pm 0.65 $ \\
    \rowcolor{gray!10!white}$1\%$ accuracy   & $2.91 \pm 0.92 $& $4.39 \pm 0.42 $\\
   \bottomrule
   \end{tabularx}%
\end{table}

\subsection{Computational complexity and runtimes}\label{app:runtime}

In a dataset with $n=1000$ data points from the setting discussed in \cref{fig:simulated_exp}, the average training time for one epoch without the regularization term is 0.003s and 1.112s with the regularization term. In these results, Adam optimizer with batch-size of 512 was used. By using smaller batch sizes, e.g. $n=128$, the extra computational cost can further decrease. From a theoretical perspective, the estimation of the HSCIC requires kernel ridge regression (see \cref{eq:estimate_hscic} in our submission). In the high-dimensional image example, with a mini batch-size of 512, the average running time for an epoch with the regularization term is 64.03s and 34.01s without. Kernel ridge regression generally has a runtime that scales as $O(n^3)$ and memory requirements scaling like $O(n^2)$, where $n$ is the size of the dataset. However, these bounds can be significantly improved by using, i.e., random Fourier Features (see, i.e., \citet{DBLP:conf/nips/RahimiR07,DBLP:conf/icml/AvronKMMVZ17}) as detailed in \cref{appendix:random_fourier_features}. In short, by using random Fourier features, the resulting approximate kernel ridge regression estimator can be computed in a runtime of $O(ns^2)$ with $O(ns)$ memory. Here, $s$ is a parameter determining the accuracy of the approximation. In practice, $s$ can be set to be much smaller than the problem size, resulting in a dramatic speed-up. Other methods for efficient kernel computation include the popular Nystrom approximation \citep{JMLR:v6:drineas05a, NIPS2014_Hsieh}, and Memory-Efficient Kernel Approximation (MEKA) \citep{JMLR:v18:Si}.
In this work, runtimes were still reasonable for all experimental settings, which is why we did not have to resort to these faster approximations.

\section{Comparison with additional baselines}
\label{appendix:sec:additional_baselines}
In this section, we compare CIP with additional baselines. These include \citet{DBLP:conf/nips/VeitchDYE21} and different heuristic methods.

\subsection{Baseline experiments \citep{DBLP:conf/nips/VeitchDYE21}}\label{app:sec:baseline_veitch}

We provide an experimental comparison against the method by \citet{DBLP:conf/nips/VeitchDYE21}. To this end, we consider the following data-generating mechanism for the causal structure (see \cref{fig:graph}(b)):
\begin{spreadlines}{1.5em}
\begin{align*}
    & \mathbf{Z}  \sim \mathcal{N}\left (0, 1 \right ) \qquad  \qquad
    \mathbf{A}  = \sin \left( 0.1 \mathbf{Z} \right)+ \varepsilon_{\mathbf{A}}\\
    & \mathbf{X} = \exp \left \{-\frac{1}{2}\mathbf{A} \right \} \sin\left(\mathbf{A}\right) + \frac{1}{10} \varepsilon_{\mathbf{X}}\\ 
    & \mathbf{Y} =  \frac{1}{10} \exp \left \{-\mathbf{X} \right \} \cdot \sin \left ( 2 \mathbf{X}\mathbf{Z} \right ) + \mathbf{A}\mathbf{A}  + \frac{1}{10} \varepsilon_{\mathbf{Y}},
\end{align*}
\end{spreadlines}
where $\varepsilon_{\mathbf{X}}, \varepsilon_{\mathbf{A}} \overset{i.i.d}{\sim} \mathcal{N}\left (0, 1 \right )$ and $\varepsilon_{\mathbf{Y}} \overset{i.i.d}{\sim} \mathcal{N}\left (0, 0.1 \right )$. The data-generating mechanism of the anti-causal structure is the following (see \cref{fig:graph}(c)):
\begin{spreadlines}{1.5em}
\begin{align*}
    & \mathbf{Z}  \sim \mathcal{N}\left (0, 1 \right )  \qquad  \qquad \mathbf{A}  =  \frac{1}{5} \sin \left( \mathbf{Z} \right) + \varepsilon_{\mathbf{A}}\\
    & \mathbf{Y} = \frac{1}{10} \sin \left( \mathbf{Z} \right)  + \varepsilon_{\mathbf{Y}}\\
    & \mathbf{X} = \mathbf{A} + \mathbf{Y}  + \frac{1}{10} \varepsilon_{\mathbf{X}}
\end{align*}
\end{spreadlines}
where $\varepsilon_{\mathbf{Y}}, \varepsilon_{\mathbf{A}} \overset{i.i.d}{\sim} \mathcal{N}\left (0, 0.1 \right )$ and $\varepsilon_{\mathbf{X}} \overset{i.i.d}{\sim} \mathcal{N}\left (0, 1 \right )$. We compare our method (CIP) against the method by \citet{DBLP:conf/nips/VeitchDYE21} using different values for the trade-off parameter $\gamma$. In \cref{fig:graph}(b-c) the causal and anti-causal graphical settings proposed by \citet{DBLP:conf/nips/VeitchDYE21} are presented. In both of these settings there is an unobserved confounder $\mathbf{Z}$ between $\mathbf{A}$ and $\mathbf{Y}$. The graphical assumptions outlined in \cref{lemma:conditiona_counterfactual} of the CIP are not met in the graphical structures under examination, as the confounding path is not effectively blocked by an observed variable ($\mathbf{Z}$ is unobserved). In light of this, it is assumed in our implementation that there is no unobserved confounder. In the graphical structure \cref{fig:graph}(b), CIP enforces $\textsc{HSIC}(\hat{\mathbf{Y}},\mathbf{A} \cup \mathbf{X})$ to become small, gradually enforcing $\hat{\mathbf{Y}} \indep \mathbf{A}\cup \mathbf{X}$. $\textsc{HSIC}$ is the Hilbert-Schmidt Independence Criterion, which is commonly used to promote independence (see, i.e., \citet{DBLP:conf/alt/GrettonBSS05,DBLP:conf/nips/FukumizuGSS07}). \citet{DBLP:conf/nips/VeitchDYE21} enforces as independence criterion $\textsc{HSIC}(\hat{\mathbf{Y}},\mathbf{A})$, which is implied by the independence enforced in CIP.  In the anti-causal graphical setting presented in \cref{fig:graph}(c), the objective term used in CIP is $\hscic{\hat{\mathbf{Y}},\mathbf{A}}{\mathbf{X}}$, while in the method of \citet{DBLP:conf/nips/VeitchDYE21} is $\hscic{\hat{\mathbf{Y}},\mathbf{A}}{\mathbf{Y}}$. In \cref{table:baseline_victor}, the results of accuracy and \VCF{} are presented.

\begin{table}
  \centering
  \caption{\textbf{Results of the MSE, \VCF{} of CIP and the baseline \citep{DBLP:conf/nips/VeitchDYE21}} applied to the causal and anti-causal structure in \cref{fig:graph}(b-c). Although the graphical assumptions are not satisfied, CIP shows an overall decrease of \VCF{} in both of the graphical structures, performing on par with the baseline \citet{DBLP:conf/nips/VeitchDYE21} in terms of accuracy and counterfactual invariance.\\}
  \label{table:baseline_victor}
  \begin{tabularx}{0.8\columnwidth}{Xrrrr}
    
    & \multicolumn{2}{c}{\textbf{CIP}} & \multicolumn{2}{c}{\textbf{\citet{DBLP:conf/nips/VeitchDYE21}}} \\
    \cmidrule(lr){2-3} \cmidrule(lr){4-5} 
    & \textbf{MSE} {\color{gray}$\times 10^2$}&$\textbf{VCF}$ & \textbf{MSE}{\color{gray}$\times 10^2$}& $\textbf{VCF}$ \\ 
    \midrule
    \rowcolor{gray!10!white} $\gamma=0.5$  & $4.58 \pm 0.31$  & $0.19 \pm 0.02$ & $4.50 \pm 0.40$  &  $0.19 \pm 0.02$ \\
    \rowcolor{white} $\gamma=1.0$   & $5.60 \pm 0.36$  & $0.18 \pm 0.01$ &  $5.45 \pm 0.41$  & $0.18 \pm 0.02$ \\
    \bottomrule
  \end{tabularx}\\[0.3cm]
  \begin{tabularx}{0.8\columnwidth}{Xrrrr}
    
    & \multicolumn{2}{c}{\textbf{CIP}} & \multicolumn{2}{c}{\textbf{\citet{DBLP:conf/nips/VeitchDYE21}}} \\
    \cmidrule(lr){2-3} \cmidrule(lr){4-5} 
    & \textbf{MSE} {\color{gray}$\times 10^2$}&$\textbf{VCF}$ & \textbf{MSE}{\color{gray}$\times 10^2$} & $\textbf{VCF}$ \\
    \midrule
    \rowcolor{gray!10!white} $\gamma=0.5$  & $1.16 \pm 0.01$ & $1.69 \pm 0.16$ & $1.01 \pm 0.01$ &   $1.71 \pm 0.26$ \\
    \rowcolor{white} $\gamma=1.0$  & $1.37 \pm 0.02$ & $1.48 \pm 0.19$ &  $0.99 \pm 0.01$ & $1.88 \pm 0.28$ \\
    \bottomrule
  \end{tabularx}%
\end{table}

In the experiments, the predictor $\hat{\mathbf{Y}}$ is a feed-forward neural network consisting of $8$ hidden layers with $20$ nodes each, connected with a rectified linear activation function (ReLU) and a linear final layer. Mini-batch size of $256$ and the Adam optimizer with a learning rate of $10^{-4}$ for $500$ epochs were used.

\begin{table}[t]
  \centering
  \caption{Results of MSE and \VCF{} (all times $10^2$ for readability) on synthetic data of CIP with trade-off parameters $\gamma=0.5$ and $\gamma=1$ with the heuristic methods \textit{data augmentation} and \textit{causal-based data augmentation} and \textit{naive prediction}.}
  \label{table:heuristic_comparison}
  \small
  \begin{tabularx}{0.7\columnwidth}{Xrr}
  \toprule
      
  & \textbf{VCF} {\color{gray}$\times 10^3$} & \textbf{MSE} {\color{gray}$\times 10^3$} \\
   \midrule
  \rowcolor{gray!10!white}  \text{data augmentation}   & $3.12 \pm 0.16$ & $0.03 \pm 0.01 $  \\
   \rowcolor{white} \text{causal-based data augmentation} & $3.04 \pm 0.16$ & $0.13 \pm 0.12$ \\
   \rowcolor{gray!10!white} \text{CIP} ($\gamma=0.5$)  & $1.05 \pm 0.13$ & $1.64 \pm 0.22 $  \\
  \rowcolor{white}  \text{CIP} ($\gamma=1.0$)   & $0.35\pm 0.19$ & $2.50 \pm 0.72  $  \\
  \rowcolor{gray!10!white} \text{naive prediction (ignore $A$)}   &$9.01 \pm 0.02$ &$3.01 \pm 0.91$ \\
   \bottomrule
   \end{tabularx}%
\end{table}

\subsection{Comparison baselines heuristic methods}\label{app:sec:baseline_heuristic}

We provide an experimental comparison of the proposed method (CIP) with some heuristic methods, specifically data-augmentation-based methods. We consider the same data-generating procedure and causal structure as presented in \cref{app:synthetic-experiment}. The heuristic methods considered are \textit{data augmentation} and \textit{causal-based data augmentation}. In the former, data augmentation is performed by generating $N=50$ samples for every data-point by sampling new values of $\mathbf{A}$ as $a_1, ..., a_N  \overset{i.i.d}{\sim} \pr_{\mathbf{A}}$ and leaving $\mathbf{Z}, \mathbf{L}, \mathbf{Y}$  unchanged. Differently, in the latter \textit{causal-based data augmentation} method, we also take into account the causal structure given by the known DAG. Indeed, when manipulating the variable $\mathbf{A}$, its descendants (in this example $\mathbf{L}$) will also change. In this experiment, a predictor for $\mathbf{L}$ as $\hat{\mathbf{L}}=f_{\theta}(\mathbf{A},\mathbf{Z})$ is trained on $80\%$ of the original dataset. In the data augmentation mechanism, for every data-point $\{a,x,z,y\}$, $N=50$ samples are generated  by sampling new values of $\mathbf{A}$ as $a_1, ..., a_N  \overset{i.i.d}{\sim} \pr_{\mathbf{A}}$, estimating the values of  $\mathbf{L}$ as $x_1=f_{\theta}(a_1,z), ..., x_N=f_{\theta}(a_N,z)$, while leaving the values of $\mathbf{Z}$ and $\mathbf{Y}$ unchanged. Heuristic methods such as data-augmentation methods do not theoretically guarantee to provide counterfactually invariant predictors. The results of an empirical comparison are shown in \cref{table:heuristic_comparison} with the average and standard deviations after 5 random seeds. It can be shown that these theoretical insights are supported by experimental results, as the \VCF{} metric measure counterfactual invariance is lower in both of the two settings of the CIP ($\gamma=\tfrac{1}{2}$ and $\gamma=1$).

A dataset of $n=3000$ is used, along with $k=500$ and $d=500$. The architecture for predicting $\mathbf{L}$ and $\mathbf{Y}$ are feed-forward neural networks consisting of $8$ hidden layers with $20$ nodes each, connected with a rectified linear activation function (ReLU) and linear final layer. Mini-batch size of $256$ and the Adam optimizer with a learning rate of $10^{-3}$ for $100$ epochs were used. 
\end{document}